\documentclass[a4paper, 10pt, conference]{ieeeconf}      
\usepackage{FG2024}
\usepackage{multirow}
\usepackage{amsmath}
\usepackage{graphicx}
\usepackage{soul}
\FGfinalcopy 

\IEEEoverridecommandlockouts                              
\def\FGPaperID{208} 

\title{\LARGE \bf
Bridging the Gap: Protocol Towards Fair and Consistent Affect Analysis
}

\author{\parbox{16cm}{\centering
    {\large Guanyu Hu$^{1,2^\star}$, Eleni Papadopoulou$^{3^\star}$, Dimitrios Kollias$^{2^\star}$, Paraskevi Tzouveli$^3$, Jie Wei$^1$ and Xinyu Yang$^1$}\\
    {\normalsize
    $^1$ The School of Computer Science and Technology, Xi'an Jiaotong University, China\\
    $^2$ School of Electronic Engineering and Computer Science, Queen Mary University of London, U.K\\ $^3$ National Technical University of Athens, Greece }}
    \thanks{The work of Guanyu Hu was supported by the Innovation Leadership Talent Development Program of Xi'an Jiaotong University}
    \thanks{$\star$ equal contribution}
}

\usepackage{hyperref}

\usepackage{fancyhdr}
\begin{document}

\ifFGfinal
\thispagestyle{empty}
\pagestyle{empty}
\else
\author{Anonymous FG2024 submission\\ Paper ID \FGPaperID \\}
\pagestyle{plain}
\fi
\maketitle

\thispagestyle{fancy}

\begin{abstract}
The increasing integration of machine learning algorithms in daily life underscores the critical need for fairness and equity in their deployment. As these technologies play a pivotal role in decision-making, addressing biases across diverse subpopulation groups, including age, gender, and race, becomes paramount. Automatic affect analysis, at the intersection of physiology, psychology, and machine learning, has seen significant development. However, existing databases and methodologies lack uniformity, leading to biased evaluations.
This work addresses these issues by analyzing six affective databases, annotating demographic attributes, and proposing a common protocol for database partitioning. Emphasis is placed on fairness in evaluations. Extensive experiments with baseline and state-of-the-art methods demonstrate the impact of these changes, revealing the inadequacy of prior assessments. The findings underscore the importance of considering demographic attributes in affect analysis research and provide a foundation for more equitable methodologies.
Our annotations, code and pre-trained models are available at: https://github.com/dkollias/Fair-Consistent-Affect-Analysis

%
%
%
%
%

\end{abstract}

\section{Introduction}




In recent years, the rapid integration of machine and deep learning algorithms into various aspects of our daily lives has prompted a growing awareness of the critical need for fairness and equity in their deployment. As these technologies play an increasingly pivotal role in decision-making processes, there is a pressing concern to ensure that their outcomes do not inadvertently reinforce or perpetuate existing societal biases. One of the pivotal dimensions in this pursuit of fairness is the examination of algorithmic behavior across diverse subpopulation groups, including but not limited to age, gender, and race. The ethical imperative of addressing potential biases and discriminatory impacts on different demographic cohorts is paramount, demanding a concerted effort to develop machine learning models that are not only accurate and efficient but also considerate of the diverse and nuanced characteristics within our society.

Automatic affect analysis has a long history of studies in the intersection of physiology, psychology, and machine/deep learning. The analysis involves automatic: a)  recognition of categorical affect, represented via the so-called six basic expressions (e.g., anger, disgust, happiness), plus the neutral state \cite{ekman2002facial}; b) detection of activation of facial action units (AUs) (i.e., specific movements of facial muscles) \cite{ekman2002facial}; c) estimation of dimensional affect, represented via the emotion descriptors of valence (characterises an emotional state on a scale from positive to negative) and arousal (characterises an emotional state on a scale from active to passive) \cite{russell1978evidence}. 

Affect analysis methods are being developed using existing databases with the goal of maximizing their performance on the test sets and surpassing the performance of other methods. In most of the cases, this is the only criterion and there is no checking whether these comparisons are fair and/or whether the methods are unbiased and perform equally well across subjects of various demographic attributes such as race, gender and age (especially considering that the utilized databases do not contain an even distribution of subjects among these groups).
The methods, unless explicitly modified, are severely impacted by such bias since they are given more opportunities (training samples) for optimizing their objectives towards the majority demographic group in the database. This leads to
lower performance for the minority groups, i.e., subjects represented with less number of samples, which is not desired.

In the past three decades, many affective databases have been constructed with a clear tendency from small-scale to large-scale and from lab-controlled to real-world unconstrained conditions (termed ``in-the-wild").
Two of the most typically utilized in-the-wild databases for Facial Expression Recognition (FER)  are RAF-DB \cite{li2017reliable} and AffectNet \cite{mollahosseini2017affectnet}. AffectNet is further annotated in terms of valence and arousal (VA).
%
%
The most widely used databases for AU recognition are DISFA \cite{mavadati2013disfa}, GFT \cite{girard2017sayette} and the in-the-wild EmotioNet \cite{emotionet2016} and
RAF-AU \cite{yan2020raf}. All these databases have some problems that we present next.

\textit{Some databases (DISFA) do not have a pre-defined split into training, validation and test sets.} Therefore many research works perform \textit{k}-fold cross validation either  with different \textit{k} \cite{mitenkova2019valence}\cite{kossaifi2020factorized}, or with the same  \textit{k} but not the same samples in each fold. Additionally, some of these works, in the \textit{k}-fold cross-validation, perform a subject-dependent split and others a subject independent one (which means that the same subject can only appear in the same split) \cite{parameshwara2023examining}\cite{parameshwara2023efficient}. All these make the derived results of the methods non-comparable, as these methods have been developed with different training and validation samples (there is also no common test samples to compare their methods).

\textit{Other databases are split into only two sets}: RAF-DB and RAF-AU have a training and test set; AffectNet has a training and validation set (the test set has not been published); RAF-AU does not provide the exact training and test partitions. For RAF-AU, research works define their own partitions without making them available so that other works can use the same partitions while developing their method \cite{an2024learning}.
For the other databases, research works either use the one set as training set and the other one as both validation and test set \cite{parameshwara2023efficient}\cite{zheng2023poster}\cite{li2019self}\cite{chaudhari2022vitfer}, or
%
split the existing training set into a new training and validation set (without specifying the samples that belong to each set) and then use these to develop their methods, evaluate their performance on the existing test set and compare with other methods \cite{varsha}\cite{handrich2020simultaneous}.
However, the performance comparison between these methods is unfair, as the method in the first approach is developed using more training data compared to the method in the second approach. 

\textit{Some databases consist of manual and automatic annotations} (EmotioNet's training set consists of only automatic labels, whereas its validation and test sets consist of only manual labels; AffectNet also contains a part that is automatically annotated). Some research works utilize automatic annotations and develop their methods. Other works do not use them, because automatic annotations are very noisy. Instead, they opt to split the validation set into new training and validation sets, and leverage these for modeling; however, they partition in different percentages (e.g. \cite{wang2020tal} split in 90-10\%, whereas \cite{werner2020facial} split in 80-20\%) and they also do not report the exact splits. All these are problematic.

\textit{The databases' test set sizes are quite small, especially compared to the training set sizes}. AffectNet's training set consists of around 290K images, whereas its test set of only 4K images; RAF-DB's training set consists of around 12K images and its test set of only 3K; Most works following the setting \cite{shao2024joint} for GFT, in which 78 subjects with around 110K frames in training set and 18 subjects with only 25K frames exist in testing set.

The evaluation of a method's performance on the test set is a critical aspect of assessing its efficacy and generalizability, and thus developed methods use this set to compare their performance to other methods. 
%
%
Thus having a small test set may result in wrong and inaccurate conclusions from these comparisons; if a method outperforms another in that small set, it does not mean that it would still outperform the other in a larger test set. We show this in the experimental section. After we re-partitioned the databases, we train various state-of-the-art methods and compared their performance; we show that methods that outperformed others in the old partitions, are not still outperforming them on the new partitions.

\textit{Some databases have inconsistent forms of annotations}.  AffectNet's VA annotations are in continuous [-1, 1], whereas AFEW-VA's annotations are integers in [-10, 10]. DISFA is annotated in terms of AU intensities (from a scale of 0-5). To convert the AU intensities into AU activation/non-activation, some research works associate an intensity higher than 0  to activation and an intensity of 0 to non-activation \cite{li2019self}\cite{liu2023pose}; other works associate an intensity higher than 1 to activation and an intensity of 0 or 1 to non-activation \cite{zhang2024multimodal}\cite{cui2023biomechanics}. DISFA contains videos from the left and right view of 27 subjects, but many research works only consider one view.
What is more, for AU annotated databases, a different combination of AUs is being used in each database and in some of them not all the released AU annotations are used.
DISFA contains annotations of 12 AUs but research works use only 8; GFT contains annotations of 32 AUs but works use only 10. All AUs should be used for consistency with the other databases, so that they contain more common AUs, which will enable and facilitate cross-database experiments.

\textit{The evaluation metrics of the databases are not common and/or are not the most appropriate ones}. In FER, research works that utilize RAF-DB mainly report the total accuracy and less often the average accuracy (across all basic expressions), whereas works using AffectNet report the total accuracy and F1 score. Total accuracy, especially in imbalanced test sets, is not an ideal measure as it can be misleading; 
thus F1 score and average accuracy are more appropriate measures. In AU detection, research works that utilize  DISFA and GFT, utilize the F1 score, works that use RAF-AU use AUC-ROC curve and F1 score, whereas works that use EmotioNet use the mean between the average accuracy and the F1 score. For VA estimation, research works that utilize  AffectNet utilize the Concordance Correlation Coefficient (CCC),  Pearson Correlation Coefficient (PCC) and Mean Squared Error (MSE).

\textit{The databases have inconsistent label distributions in each set partition, which along with the heavy label imbalance and the small sizes constitute a bad combination}. A crucial assumption in machine learning is that the data distributions and their labels are the same for the training, validation and test sets. However, this is not the case in many databases (e.g., AffectNet, EmotioNet); this phenomenon, known as label shift, severely affects methods' performance, since minimizing the objective on the validation set no longer results in good performance on the test set.


\textit{The databases do not necessarily contain an even distribution of subjects in terms of demographic attributes such as race, gender and age and/or such attributes have not been taken into consideration when the databases' partitions were performed}.  For instance, GFT includes only subjects aged 21-28, but its test set does not include subjects aged 25-27; GFT does not contain any Asian subjects in the test set; the large-scale AffectNet had only 27 test images of people aged 70 or older. 
%
%
%
%
%
\textit{Finally, most databases do not contain labels regarding these attributes, making it difficult to assess bias and methods' performance in each attribute.}








\noindent
All things considered, the contributions of this work are:
\begin{itemize}
    \item annotating six affective databases in terms of demographic attributes (age, gender and race);
    
    \item partitioning these affective databases according to a common protocol that we define; in that protocol we pay particular attention to the demographic attributes and to fairness of evaluations;
    
    \item conducting extensive experimental study of various baseline and state-of-the-art methods in each database using the new protocol; various performance metrics are utilized, including ones measuring fairness and bias
\end{itemize}

\section{Materials \& Methods}
\noindent Here, we describe the annotation of the 7 affective databases for demographic attributes. We also define and describe a common protocol for partitioning these databases, taking into account the problems presented in the introduction section and the demographic annotations. Finally, we present the new partition of the databases and provide relevant statistics.

\subsection{Annotation}

We annotated all affective databases (AffectNet, RAF-DB, DISFA, EmotioNet, GFT and RAF-AU) for demographic attributes of age, gender, and race. Let us note that although, in practice, race and ethnicity terms are used interchangeably, they are not the same. Race and ethnicity are different categorizations of humans. Race is defined based on physical traits and ethnicity is based on cultural similarities \cite{schaefer2008encyclopedia}. Thus we decided to annotate the databases in terms of race (facial appearance is easier to judge and annotate). We adopted a commonly accepted race classification from the U.S. Census Bureau; we defined the following 5 race groups (in alphabetical order): Asian, Black (or African American), Indian (or Alaska Native), Native Hawaiian (or Other Pacific Islander), and White (or Caucasian).
During the annotation process, we did not find any cases of Native Hawaiian  or Other Pacific Islander. 

We further annotated all databases in terms of gender, with the options being: Male, Female, and Other/Uncertain. The “Uncertain” category mainly includes cases that are challenging to determine the gender, e.g., infants below the age of 3 or adolescents aged 4-19 with significant obstructions in the central part of the image, such as hats or hands.

Finally, we annotated all databases for age. We divide the age into nine categories, i.e. 0–2, 3–9, 10–19, 20–29, 30–39, 40–49, 50–59, 60–69, 70 and over. We made this categorization: i) to be consistent with age ranges in other databases (e.g. FairFace\cite{fairface}, RAF-DB); ii) to provide more fine-grained categories (rather than ones like ‘young’, ‘middle-aged’, ‘old’); and iii) to make the annotation easier and the process less noisy (it is easier to tell if a person is between 40 and 49 years old, rather than if he is 42 or 49 years old).

We developed our own annotation software that enabled the annotation of each image and video independently in terms of race, gender, and age group for each subject. Two labelers annotated around 20,000 images from each image-database and all videos from each video-database. If both agreed on
their judgments, we took the values as ground truth. 
The annotation process involved a final step. We trained a model with our ground truths and tested it on the rest of the images of each image-database so that we create annotations for these images as well. 

\subsection{Protocol}
In this subsection, we define the protocol for creating new partitions for each of the 7 affective databases. Each database has been split in the same way, with the same rationale (for consistency), which involves: i)  partitioning into 3 sets (training, validation and test); ii) each set having an adequate amount of data and subjects (to a possible extent given the whole dataset size); iii) each set having similar distributions in terms of affect labels (basic expressions, action units, or valence-arousal), age groups, race groups and genders; iv) each set being subject independent; v) utilization of only manual annotations of affect labels; vi) usage of specific performance evaluation metrics; vii) usage of a consistent annotation form. 

To give some more details on the above points: for point (vi), the F1 score is selected for evaluating AU detection and basic expression recognition (average accuracy is also good in the later case); CCC is selected for evaluating  VA estimation (average CCC of valence and arousal). For point (vii), the VA label values of AFEW-VA are scaled to the range [-1,1]. In DISFA, all 12 AUs are utilized; AU labels with an intensity higher than 0 correspond to AU activation, while an intensity of 0 corresponds to AU non-activation; both subjects' views are included in the new partition. All 14 AUs in GFT are used.

\begin{figure}[tb]
\centering
  \includegraphics[width=0.5\textwidth]{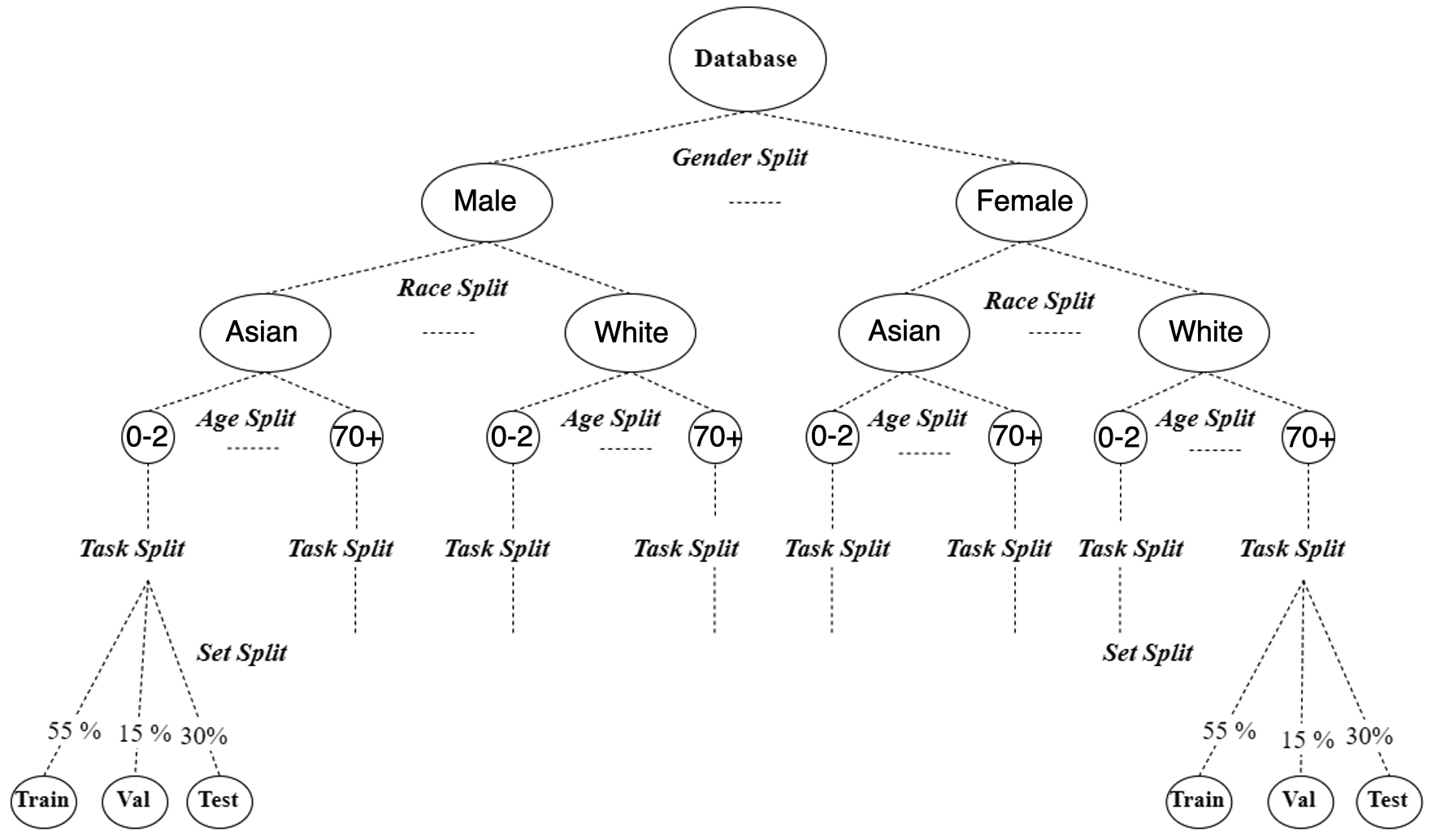}
  \caption{The proposed protocol and subsequent partition of a database}
  \label{protocol1}
\end{figure}

\begin{figure}[h]
\centering
  \includegraphics[width=0.5\textwidth]{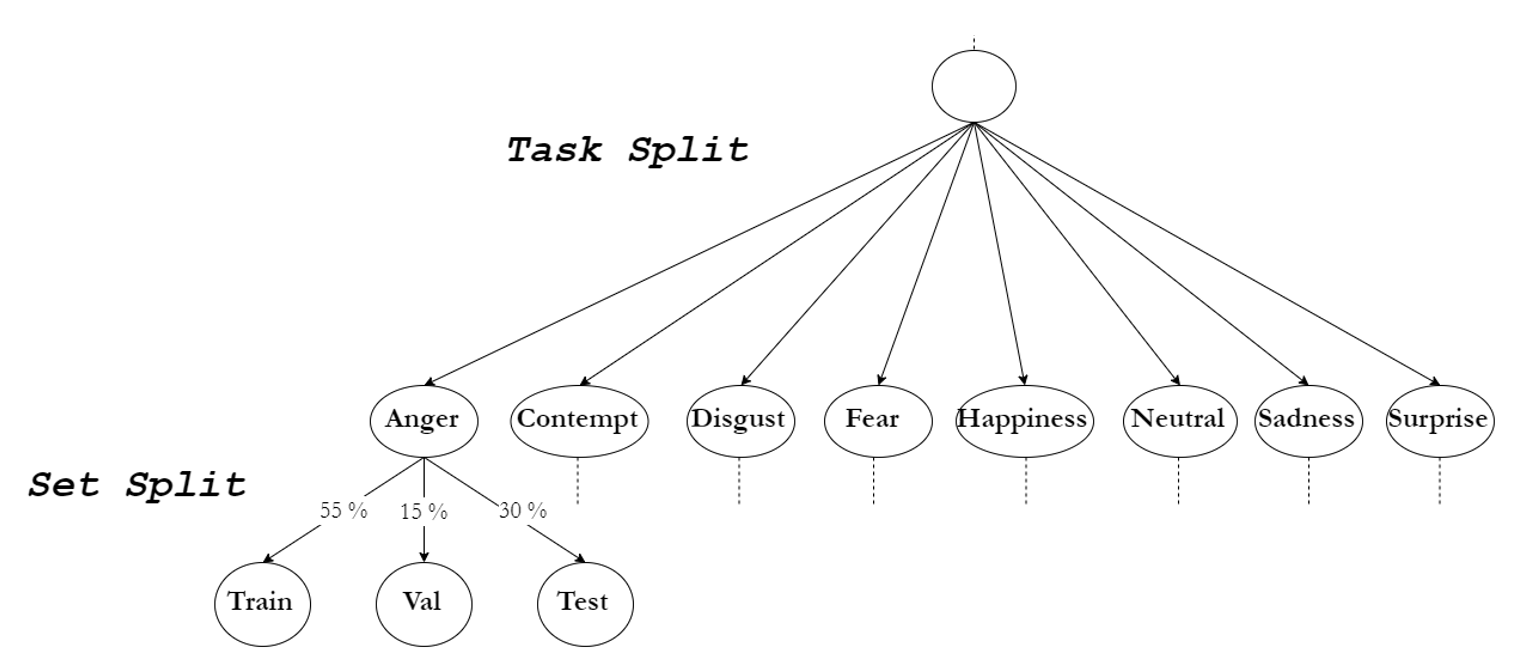}
  \caption{'Task Split' part of protocol \& partition in case of basic expressions}
  \label{protocol2}
\end{figure}

\begin{figure}[h]
\centering
  \includegraphics[width=0.47\textwidth]{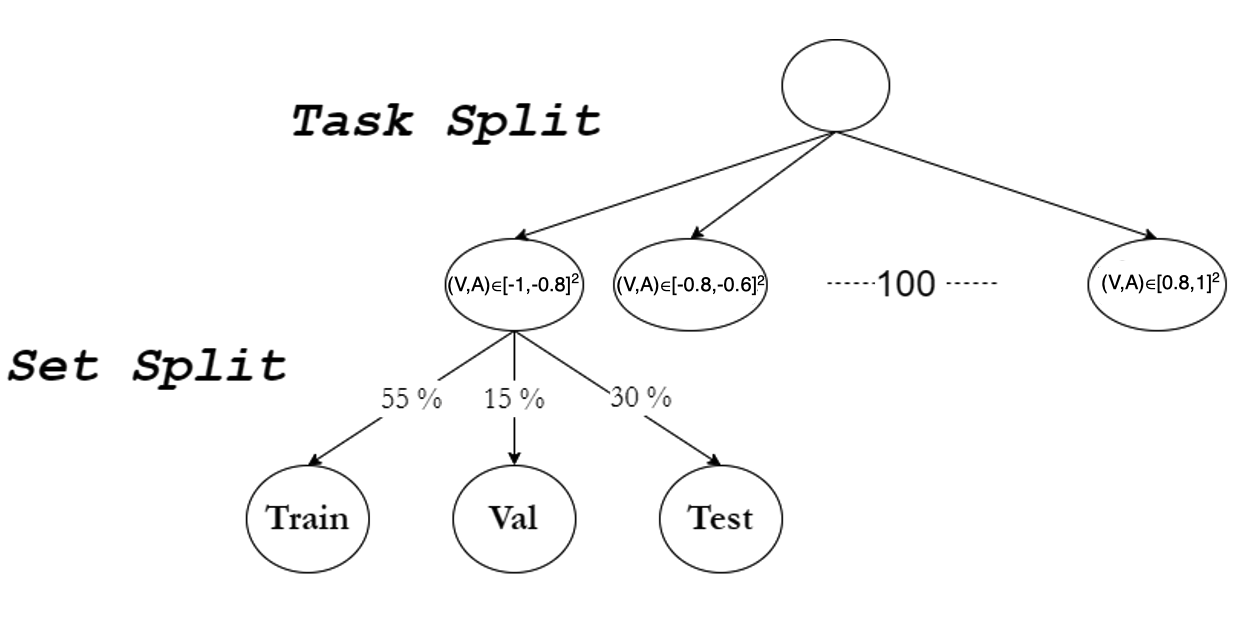}
  \vspace{-10pt}
  \caption{'Task Split' part of proposed protocol and partition in case of VA}
  \label{protocol3}
\end{figure}

Finally, for point (iii), the training, validation and test sets follow the 55\%-15\%-30\% rule, according to which the training set consists of 55\% of the data (spanning all affect labels and all race, all age groups and all genders), the validation set consists of 15\% of the data and the test set consists of the remaining 30\% of the data. Figure \ref{protocol1} illustrates the proposed protocol and database partition. Figure \ref{protocol2} illustrates the 'Task Split' part of the proposed protocol and database partition in the case of basic expressions. In the case of VA, at first we convert the continuous 2D space into distinct regions (one such region is when valence takes values in [-1,-0.8] and arousal takes values in [-0.8,-0.6]). Figure \ref{protocol3} illustrates the 'Task Split' part of the proposed protocol and database partition in the case of VA.


\subsection{Databases: Original vs New Partition}

\begin{figure}[h]   
	\centering
	\includegraphics[width=1\linewidth]{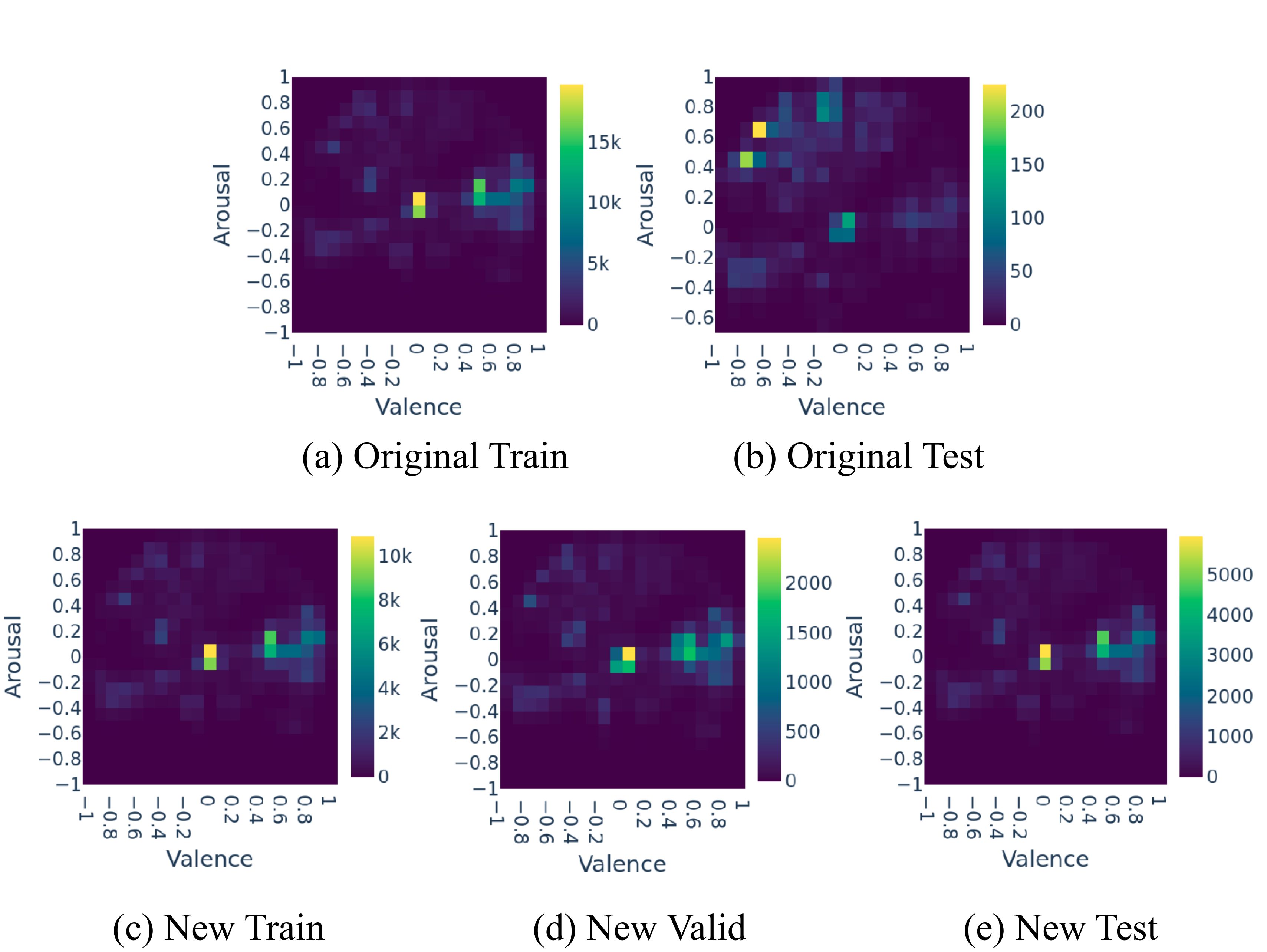}   
	\caption{2D Valence-Arousal Histogram: A Comparison Between the Original and New Partitions of AffectNet}
	\label{fig:fig1} 
\end{figure}

Tables \ref{table2}, \ref{table3} and \ref{table1}, as well as Figure \ref{fig:fig1}  present information (Label, Gender, Race, Age) regarding the original and new partitions (according to the previously mentioned protocol) of all utilized affective databases.

\begin{table}[]
\scriptsize
\caption{Data Statistics for the Original  and New Partitions of Databases Annotated in terms of basic expressions }\label{table2}
\setlength{\tabcolsep}{0.6mm}
\renewcommand{\arraystretch}{1.1} 
\scalebox{.95}{
\begin{tabular}{|cc|ccccc|ccccc|}
\hline
\multicolumn{2}{|c|}{\textbf{Database}} &
  \multicolumn{5}{c|}{\textbf{AffectNet}} &
  \multicolumn{5}{c|}{\textbf{RAF-DB}} \\ \hline
\multicolumn{2}{|c|}{\textbf{Partition}} &
  \multicolumn{2}{c|}{\textbf{Original}} &
  \multicolumn{3}{c|}{\textbf{New}} &
  \multicolumn{2}{c|}{\textbf{Original}} &
  \multicolumn{3}{c|}{\textbf{New}} \\ \hline
\multicolumn{2}{|c|}{\textbf{Set}} &
  \multicolumn{1}{c|}{\textbf{Train}} &
  \multicolumn{1}{c|}{\textbf{Test}} &
  \multicolumn{1}{c|}{\textbf{Train}} &
  \multicolumn{1}{c|}{\textbf{Valid}} &
  \textbf{Test} &
  \multicolumn{1}{c|}{\textbf{Train}} &
  \multicolumn{1}{c|}{\textbf{Test}} &
  \multicolumn{1}{c|}{\textbf{Train}} &
  \multicolumn{1}{c|}{\textbf{Valid}} &
  \textbf{Test} \\ \hline
\multicolumn{2}{|c|}{\textbf{Total}} &
  \multicolumn{1}{c|}{287651} &
  \multicolumn{1}{c|}{3999} &
  \multicolumn{1}{c|}{159540} &
  \multicolumn{1}{c|}{43330} &
  87710 &
  \multicolumn{1}{c|}{12271} &
  \multicolumn{1}{c|}{3068} &
  \multicolumn{1}{c|}{8327} &
  \multicolumn{1}{c|}{2206} &
  4806 \\ \hline
\multicolumn{1}{|c|}{\multirow{2}{*}{\textbf{Gen.}}} &
  \textbf{Female} &
  \multicolumn{1}{c|}{144422} &
  \multicolumn{1}{c|}{1794} &
  \multicolumn{1}{c|}{80281} &
  \multicolumn{1}{c|}{21808} &
  44127 &
  \multicolumn{1}{c|}{6562} &
  \multicolumn{1}{c|}{1620} &
  \multicolumn{1}{c|}{4455} &
  \multicolumn{1}{c|}{1189} &
  2538 \\ \cline{2-12} 
\multicolumn{1}{|c|}{} &
  \textbf{Male} &
  \multicolumn{1}{c|}{142172} &
  \multicolumn{1}{c|}{2192} &
  \multicolumn{1}{c|}{79259} &
  \multicolumn{1}{c|}{21522} &
  43583 &
  \multicolumn{1}{c|}{4957} &
  \multicolumn{1}{c|}{1249} &
  \multicolumn{1}{c|}{3366} &
  \multicolumn{1}{c|}{891} &
  1949 \\ \hline
\multicolumn{1}{|c|}{\multirow{4}{*}{\textbf{Race}}} &
  \textbf{Asian} &
  \multicolumn{1}{c|}{25573} &
  \multicolumn{1}{c|}{311} &
  \multicolumn{1}{c|}{14168} &
  \multicolumn{1}{c|}{3823} &
  7893 &
  \multicolumn{1}{c|}{1912} &
  \multicolumn{1}{c|}{483} &
  \multicolumn{1}{c|}{1285} &
  \multicolumn{1}{c|}{329} &
  781 \\ \cline{2-12} 
\multicolumn{1}{|c|}{} &
  \textbf{Black} &
  \multicolumn{1}{c|}{23578} &
  \multicolumn{1}{c|}{321} &
  \multicolumn{1}{c|}{13072} &
  \multicolumn{1}{c|}{3518} &
  7309 &
  \multicolumn{1}{c|}{968} &
  \multicolumn{1}{c|}{234} &
  \multicolumn{1}{c|}{631} &
  \multicolumn{1}{c|}{156} &
  415 \\ \cline{2-12} 
\multicolumn{1}{|c|}{} &
  \textbf{Indian} &
  \multicolumn{1}{c|}{17201} &
  \multicolumn{1}{c|}{234} &
  \multicolumn{1}{c|}{9526} &
  \multicolumn{1}{c|}{2552} &
  5357 &
  \multicolumn{1}{c|}{/} &
  \multicolumn{1}{c|}{/} &
  \multicolumn{1}{c|}{/} &
  \multicolumn{1}{c|}{/} &
  / \\ \cline{2-12} 
\multicolumn{1}{|c|}{} &
  \textbf{White} &
  \multicolumn{1}{c|}{220242} &
  \multicolumn{1}{c|}{3120} &
  \multicolumn{1}{c|}{122774} &
  \multicolumn{1}{c|}{33437} &
  67151 &
  \multicolumn{1}{c|}{9391} &
  \multicolumn{1}{c|}{2351} &
  \multicolumn{1}{c|}{6411} &
  \multicolumn{1}{c|}{1721} &
  3610 \\ \hline
\multicolumn{1}{|c|}{\multirow{9}{*}{\textbf{Age}}} &
  \textbf{$\le$ 2} &
  \multicolumn{1}{c|}{15418} &
  \multicolumn{1}{c|}{288} &
  \multicolumn{1}{c|}{8607} &
  \multicolumn{1}{c|}{2327} &
  4772 &
  \multicolumn{1}{c|}{1283} &
  \multicolumn{1}{c|}{329} &
  \multicolumn{1}{c|}{865} &
  \multicolumn{1}{c|}{219} &
  528 \\ \cline{2-12} 
\multicolumn{1}{|c|}{} &
  \textbf{03-09} &
  \multicolumn{1}{c|}{17794} &
  \multicolumn{1}{c|}{257} &
  \multicolumn{1}{c|}{9895} &
  \multicolumn{1}{c|}{2684} &
  5472 &
  \multicolumn{1}{c|}{\multirow{2}{*}{2171}} &
  \multicolumn{1}{c|}{\multirow{2}{*}{486}} &
  \multicolumn{1}{c|}{\multirow{2}{*}{1436}} &
  \multicolumn{1}{c|}{\multirow{2}{*}{377}} &
  \multirow{2}{*}{844} \\ \cline{2-7}
\multicolumn{1}{|c|}{} &
  \textbf{10-19} &
  \multicolumn{1}{c|}{15042} &
  \multicolumn{1}{c|}{177} &
  \multicolumn{1}{c|}{8337} &
  \multicolumn{1}{c|}{2249} &
  4633 &
  \multicolumn{1}{c|}{} &
  \multicolumn{1}{c|}{} &
  \multicolumn{1}{c|}{} &
  \multicolumn{1}{c|}{} &
   \\ \cline{2-12} 
\multicolumn{1}{|c|}{} &
  \textbf{20-29} &
  \multicolumn{1}{c|}{117914} &
  \multicolumn{1}{c|}{1464} &
  \multicolumn{1}{c|}{65624} &
  \multicolumn{1}{c|}{17881} &
  35873 &
  \multicolumn{1}{c|}{\multirow{2}{*}{6531}} &
  \multicolumn{1}{c|}{\multirow{2}{*}{1662}} &
  \multicolumn{1}{c|}{\multirow{2}{*}{4480}} &
  \multicolumn{1}{c|}{\multirow{2}{*}{1212}} &
  \multirow{2}{*}{2501} \\ \cline{2-7}
\multicolumn{1}{|c|}{} &
  \textbf{30-39} &
  \multicolumn{1}{c|}{56462} &
  \multicolumn{1}{c|}{821} &
  \multicolumn{1}{c|}{31476} &
  \multicolumn{1}{c|}{8564} &
  17243 &
  \multicolumn{1}{c|}{} &
  \multicolumn{1}{c|}{} &
  \multicolumn{1}{c|}{} &
  \multicolumn{1}{c|}{} &
   \\ \cline{2-12} 
\multicolumn{1}{|c|}{} &
  \textbf{40-49} &
  \multicolumn{1}{c|}{28280} &
  \multicolumn{1}{c|}{427} &
  \multicolumn{1}{c|}{15758} &
  \multicolumn{1}{c|}{4272} &
  8677 &
  \multicolumn{1}{c|}{\multirow{3}{*}{1920}} &
  \multicolumn{1}{c|}{\multirow{3}{*}{502}} &
  \multicolumn{1}{c|}{\multirow{3}{*}{1312}} &
  \multicolumn{1}{c|}{\multirow{3}{*}{344}} &
  \multirow{3}{*}{766} \\ \cline{2-7}
\multicolumn{1}{|c|}{} &
  \textbf{50-59} &
  \multicolumn{1}{c|}{22950} &
  \multicolumn{1}{c|}{363} &
  \multicolumn{1}{c|}{12790} &
  \multicolumn{1}{c|}{3467} &
  7056 &
  \multicolumn{1}{c|}{} &
  \multicolumn{1}{c|}{} &
  \multicolumn{1}{c|}{} &
  \multicolumn{1}{c|}{} &
   \\ \cline{2-7}
\multicolumn{1}{|c|}{} &
  \textbf{60-69} &
  \multicolumn{1}{c|}{9395} &
  \multicolumn{1}{c|}{162} &
  \multicolumn{1}{c|}{5227} &
  \multicolumn{1}{c|}{1407} &
  2923 &
  \multicolumn{1}{c|}{} &
  \multicolumn{1}{c|}{} &
  \multicolumn{1}{c|}{} &
  \multicolumn{1}{c|}{} &
   \\ \cline{2-12} 
\multicolumn{1}{|c|}{} &
  \textbf{$\ge$ 70} &
  \multicolumn{1}{c|}{3339} &
  \multicolumn{1}{c|}{27} &
  \multicolumn{1}{c|}{1826} &
  \multicolumn{1}{c|}{479} &
  1061 &
  \multicolumn{1}{c|}{366} &
  \multicolumn{1}{c|}{89} &
  \multicolumn{1}{c|}{234} &
  \multicolumn{1}{c|}{54} &
  167 \\ \hline
\multicolumn{1}{|c|}{\multirow{8}{*}{\textbf{Expr.}}} &
  \textbf{Neutral} &
  \multicolumn{1}{c|}{74874} &
  \multicolumn{1}{c|}{500} &
  \multicolumn{1}{c|}{41252} &
  \multicolumn{1}{c|}{11223} &
  22588 &
  \multicolumn{1}{c|}{2524} &
  \multicolumn{1}{c|}{680} &
  \multicolumn{1}{c|}{1744} &
  \multicolumn{1}{c|}{463} &
  997 \\ \cline{2-12} 
\multicolumn{1}{|c|}{} &
  \textbf{Happiness} &
  \multicolumn{1}{c|}{134415} &
  \multicolumn{1}{c|}{500} &
  \multicolumn{1}{c|}{73958} &
  \multicolumn{1}{c|}{20144} &
  40432 &
  \multicolumn{1}{c|}{4772} &
  \multicolumn{1}{c|}{1185} &
  \multicolumn{1}{c|}{3260} &
  \multicolumn{1}{c|}{877} &
  1820 \\ \cline{2-12} 
\multicolumn{1}{|c|}{} &
  \textbf{Sadness} &
  \multicolumn{1}{c|}{25459} &
  \multicolumn{1}{c|}{500} &
  \multicolumn{1}{c|}{14174} &
  \multicolumn{1}{c|}{3842} &
  7816 &
  \multicolumn{1}{c|}{1982} &
  \multicolumn{1}{c|}{478} &
  \multicolumn{1}{c|}{1332} &
  \multicolumn{1}{c|}{353} &
  775 \\ \cline{2-12} 
\multicolumn{1}{|c|}{} &
  \textbf{Surprise} &
  \multicolumn{1}{c|}{14090} &
  \multicolumn{1}{c|}{500} &
  \multicolumn{1}{c|}{7951} &
  \multicolumn{1}{c|}{2148} &
  4425 &
  \multicolumn{1}{c|}{1290} &
  \multicolumn{1}{c|}{329} &
  \multicolumn{1}{c|}{872} &
  \multicolumn{1}{c|}{229} &
  518 \\ \cline{2-12} 
\multicolumn{1}{|c|}{} &
  \textbf{Fear} &
  \multicolumn{1}{c|}{6378} &
  \multicolumn{1}{c|}{500} &
  \multicolumn{1}{c|}{3729} &
  \multicolumn{1}{c|}{997} &
  2113 &
  \multicolumn{1}{c|}{281} &
  \multicolumn{1}{c|}{74} &
  \multicolumn{1}{c|}{184} &
  \multicolumn{1}{c|}{44} &
  127 \\ \cline{2-12} 
\multicolumn{1}{|c|}{} &
  \textbf{Disgust} &
  \multicolumn{1}{c|}{3803} &
  \multicolumn{1}{c|}{500} &
  \multicolumn{1}{c|}{2318} &
  \multicolumn{1}{c|}{611} &
  1355 &
  \multicolumn{1}{c|}{717} &
  \multicolumn{1}{c|}{160} &
  \multicolumn{1}{c|}{471} &
  \multicolumn{1}{c|}{119} &
  287 \\ \cline{2-12} 
\multicolumn{1}{|c|}{} &
  \textbf{Anger} &
  \multicolumn{1}{c|}{24882} &
  \multicolumn{1}{c|}{500} &
  \multicolumn{1}{c|}{13859} &
  \multicolumn{1}{c|}{3755} &
  7649 &
  \multicolumn{1}{c|}{705} &
  \multicolumn{1}{c|}{162} &
  \multicolumn{1}{c|}{464} &
  \multicolumn{1}{c|}{121} &
  282 \\ \cline{2-12} 
\multicolumn{1}{|c|}{} &
  \textbf{Contempt} &
  \multicolumn{1}{c|}{3750} &
  \multicolumn{1}{c|}{499} &
  \multicolumn{1}{c|}{2299} &
  \multicolumn{1}{c|}{610} &
  1332 &
  \multicolumn{1}{c|}{/} &
  \multicolumn{1}{c|}{/} &
  \multicolumn{1}{c|}{/} &
  \multicolumn{1}{c|}{/} &
  / \\ \hline
\end{tabular}}
\end{table}

\begin{table*}[]
\centering\scriptsize
\caption{Data Statistics for the Original  and New Partitions of Databases Annotated for AUs ('Org.' denotes Original partition)}\label{table3}
\setlength{\tabcolsep}{1.3mm}
\renewcommand{\arraystretch}{1.1} 
\scalebox{1}{
\begin{tabular}{|cc|cccc|ccccc|ccccc|cccc|}
\hline
\multicolumn{2}{|c|}{\textbf{Database}} &
  \multicolumn{4}{c|}{\textbf{DISFA}} &
  \multicolumn{5}{c|}{\textbf{EmotioNet}} &
  \multicolumn{5}{c|}{\textbf{GFT}} &
  \multicolumn{4}{c|}{\textbf{RAF-AU}} \\ \hline
\multicolumn{2}{|c|}{\textbf{Partition}} &
  \multicolumn{1}{c|}{\multirow{2}{*}{\textbf{Org.}}} &
  \multicolumn{3}{c|}{\textbf{New}} &
  \multicolumn{2}{c|}{\textbf{Org.}} &
  \multicolumn{3}{c|}{\textbf{New}} &
  \multicolumn{2}{c|}{\textbf{Org.}} &
  \multicolumn{3}{c|}{\textbf{New}} &
  \multicolumn{1}{c|}{\multirow{2}{*}{\textbf{Org.}}} &
  \multicolumn{3}{c|}{\textbf{New}} \\ \cline{1-2} \cline{4-16} \cline{18-20} 
\multicolumn{2}{|c|}{\textbf{Set}} &
  \multicolumn{1}{c|}{} &
  \multicolumn{1}{c|}{\textbf{Train}} &
  \multicolumn{1}{c|}{\textbf{Valid}} &
  \textbf{Test} &
  \multicolumn{1}{c|}{\textbf{Train}} &
  \multicolumn{1}{c|}{\textbf{Test}} &
  \multicolumn{1}{c|}{\textbf{Train}} &
  \multicolumn{1}{c|}{\textbf{Valid}} &
  \textbf{Test} &
  \multicolumn{1}{c|}{\textbf{Train}} &
  \multicolumn{1}{c|}{\textbf{Test}} &
  \multicolumn{1}{c|}{\textbf{Train}} &
  \multicolumn{1}{c|}{\textbf{Valid}} &
  \textbf{Test} &
  \multicolumn{1}{c|}{} &
  \multicolumn{1}{c|}{\textbf{Train}} &
  \multicolumn{1}{c|}{\textbf{Valid}} &
  \textbf{Test} \\ \hline
\multicolumn{2}{|c|}{\textbf{Total Amount}} &
  \multicolumn{1}{c|}{261628} &
  \multicolumn{1}{c|}{141041} &
  \multicolumn{1}{c|}{29070} &
  84442 &
  \multicolumn{1}{c|}{24644} &
  \multicolumn{1}{c|}{20361} &
  \multicolumn{1}{c|}{24674} &
  \multicolumn{1}{c|}{6705} &
  13549 &
  \multicolumn{1}{c|}{108549} &
  \multicolumn{1}{c|}{24645} &
  \multicolumn{1}{c|}{69618} &
  \multicolumn{1}{c|}{23685} &
  39891 &
  \multicolumn{1}{c|}{4601} &
  \multicolumn{1}{c|}{2469} &
  \multicolumn{1}{c|}{645} &
  1426 \\ \hline
\multicolumn{1}{|c|}{\multirow{2}{*}{\textbf{Gender}}} &
  \textbf{Female} &
  \multicolumn{1}{c|}{116278} &
  \multicolumn{1}{c|}{63521} &
  \multicolumn{1}{c|}{9690} &
  38723 &
  \multicolumn{1}{c|}{11716} &
  \multicolumn{1}{c|}{10132} &
  \multicolumn{1}{c|}{11996} &
  \multicolumn{1}{c|}{3261} &
  6591 &
  \multicolumn{1}{c|}{45736} &
  \multicolumn{1}{c|}{11120} &
  \multicolumn{1}{c|}{30452} &
  \multicolumn{1}{c|}{9778} &
  16626 &
  \multicolumn{1}{c|}{2301} &
  \multicolumn{1}{c|}{1250} &
  \multicolumn{1}{c|}{328} &
  723 \\ \cline{2-20} 
\multicolumn{1}{|c|}{} &
  \textbf{Male} &
  \multicolumn{1}{c|}{145350} &
  \multicolumn{1}{c|}{77520} &
  \multicolumn{1}{c|}{19380} &
  45719 &
  \multicolumn{1}{c|}{12892} &
  \multicolumn{1}{c|}{10188} &
  \multicolumn{1}{c|}{12678} &
  \multicolumn{1}{c|}{3444} &
  6958 &
  \multicolumn{1}{c|}{62813} &
  \multicolumn{1}{c|}{13525} &
  \multicolumn{1}{c|}{39166} &
  \multicolumn{1}{c|}{13907} &
  23265 &
  \multicolumn{1}{c|}{2239} &
  \multicolumn{1}{c|}{1219} &
  \multicolumn{1}{c|}{317} &
  703 \\ \hline
\multicolumn{1}{|c|}{\multirow{4}{*}{\textbf{Race}}} &
  \textbf{Asian} &
  \multicolumn{1}{c|}{29070} &
  \multicolumn{1}{c|}{9690} &
  \multicolumn{1}{c|}{/} &
  19296 &
  \multicolumn{1}{c|}{1410} &
  \multicolumn{1}{c|}{1043} &
  \multicolumn{1}{c|}{1340} &
  \multicolumn{1}{c|}{359} &
  754 &
  \multicolumn{1}{c|}{2641} &
  \multicolumn{1}{c|}{/} &
  \multicolumn{1}{c|}{/} &
  \multicolumn{1}{c|}{/} &
  2641 &
  \multicolumn{1}{c|}{453} &
  \multicolumn{1}{c|}{243} &
  \multicolumn{1}{c|}{58} &
  152 \\ \cline{2-20} 
\multicolumn{1}{|c|}{} &
  \textbf{Black} &
  \multicolumn{1}{c|}{9690} &
  \multicolumn{1}{c|}{/} &
  \multicolumn{1}{c|}{/} &
  9690 &
  \multicolumn{1}{c|}{1979} &
  \multicolumn{1}{c|}{1935} &
  \multicolumn{1}{c|}{2144} &
  \multicolumn{1}{c|}{579} &
  1191 &
  \multicolumn{1}{c|}{7989} &
  \multicolumn{1}{c|}{4891} &
  \multicolumn{1}{c|}{9050} &
  \multicolumn{1}{c|}{/} &
  3830 &
  \multicolumn{1}{c|}{274} &
  \multicolumn{1}{c|}{144} &
  \multicolumn{1}{c|}{34} &
  96 \\ \cline{2-20} 
\multicolumn{1}{|c|}{} &
  \textbf{Indian} &
  \multicolumn{1}{c|}{9690} &
  \multicolumn{1}{c|}{/} &
  \multicolumn{1}{c|}{/} &
  9690 &
  \multicolumn{1}{c|}{1442} &
  \multicolumn{1}{c|}{1210} &
  \multicolumn{1}{c|}{1450} &
  \multicolumn{1}{c|}{388} &
  814 &
  \multicolumn{1}{c|}{/} &
  \multicolumn{1}{c|}{/} &
  \multicolumn{1}{c|}{/} &
  \multicolumn{1}{c|}{/} &
  / &
  \multicolumn{1}{c|}{252} &
  \multicolumn{1}{c|}{131} &
  \multicolumn{1}{c|}{29} &
  92 \\ \cline{2-20} 
\multicolumn{1}{|c|}{} &
  \textbf{White} &
  \multicolumn{1}{c|}{213178} &
  \multicolumn{1}{c|}{131351} &
  \multicolumn{1}{c|}{29070} &
  45766 &
  \multicolumn{1}{c|}{19777} &
  \multicolumn{1}{c|}{16132} &
  \multicolumn{1}{c|}{19740} &
  \multicolumn{1}{c|}{5379} &
  10790 &
  \multicolumn{1}{c|}{97919} &
  \multicolumn{1}{c|}{18505} &
  \multicolumn{1}{c|}{60568} &
  \multicolumn{1}{c|}{23685} &
  32171 &
  \multicolumn{1}{c|}{3561} &
  \multicolumn{1}{c|}{1951} &
  \multicolumn{1}{c|}{524} &
  1086 \\ \hline
\multicolumn{1}{|c|}{\multirow{9}{*}{\textbf{Age}}} &
  \textbf{$ \leq $ 2} &
  \multicolumn{1}{c|}{/} &
  \multicolumn{1}{c|}{/} &
  \multicolumn{1}{c|}{/} &
  / &
  \multicolumn{1}{c|}{583} &
  \multicolumn{1}{c|}{262} &
  \multicolumn{1}{c|}{460} &
  \multicolumn{1}{c|}{123} &
  262 &
  \multicolumn{1}{c|}{/} &
  \multicolumn{1}{c|}{/} &
  \multicolumn{1}{c|}{/} &
  \multicolumn{1}{c|}{/} &
  / &
  \multicolumn{1}{c|}{352} &
  \multicolumn{1}{c|}{191} &
  \multicolumn{1}{c|}{48} &
  113 \\ \cline{2-20} 
\multicolumn{1}{|c|}{} &
  \textbf{03-09} &
  \multicolumn{1}{c|}{/} &
  \multicolumn{1}{c|}{/} &
  \multicolumn{1}{c|}{/} &
  / &
  \multicolumn{1}{c|}{939} &
  \multicolumn{1}{c|}{488} &
  \multicolumn{1}{c|}{782} &
  \multicolumn{1}{c|}{210} &
  435 &
  \multicolumn{1}{c|}{/} &
  \multicolumn{1}{c|}{/} &
  \multicolumn{1}{c|}{/} &
  \multicolumn{1}{c|}{/} &
  / &
  \multicolumn{1}{c|}{585} &
  \multicolumn{1}{c|}{318} &
  \multicolumn{1}{c|}{83} &
  184 \\ \cline{2-20} 
\multicolumn{1}{|c|}{} &
  \textbf{10-19} &
  \multicolumn{1}{c|}{9690} &
  \multicolumn{1}{c|}{/} &
  \multicolumn{1}{c|}{/} &
  9690 &
  \multicolumn{1}{c|}{1047} &
  \multicolumn{1}{c|}{754} &
  \multicolumn{1}{c|}{986} &
  \multicolumn{1}{c|}{266} &
  549 &
  \multicolumn{1}{c|}{/} &
  \multicolumn{1}{c|}{/} &
  \multicolumn{1}{c|}{/} &
  \multicolumn{1}{c|}{/} &
  / &
  \multicolumn{1}{c|}{222} &
  \multicolumn{1}{c|}{117} &
  \multicolumn{1}{c|}{30} &
  75 \\ \cline{2-20} 
\multicolumn{1}{|c|}{} &
  \textbf{20-29} &
  \multicolumn{1}{c|}{213178} &
  \multicolumn{1}{c|}{131351} &
  \multicolumn{1}{c|}{29070} &
  48413 &
  \multicolumn{1}{c|}{10101} &
  \multicolumn{1}{c|}{8912} &
  \multicolumn{1}{c|}{10454} &
  \multicolumn{1}{c|}{2849} &
  5710 &
  \multicolumn{1}{c|}{108549} &
  \multicolumn{1}{c|}{24645} &
  \multicolumn{1}{c|}{69618} &
  \multicolumn{1}{c|}{23685} &
  39891 &
  \multicolumn{1}{c|}{1662} &
  \multicolumn{1}{c|}{912} &
  \multicolumn{1}{c|}{246} &
  504 \\ \cline{2-20} 
\multicolumn{1}{|c|}{} &
  \textbf{30-39} &
  \multicolumn{1}{c|}{19380} &
  \multicolumn{1}{c|}{/} &
  \multicolumn{1}{c|}{/} &
  19333 &
  \multicolumn{1}{c|}{5026} &
  \multicolumn{1}{c|}{4206} &
  \multicolumn{1}{c|}{5075} &
  \multicolumn{1}{c|}{1380} &
  2777 &
  \multicolumn{1}{c|}{/} &
  \multicolumn{1}{c|}{/} &
  \multicolumn{1}{c|}{/} &
  \multicolumn{1}{c|}{/} &
  / &
  \multicolumn{1}{c|}{1068} &
  \multicolumn{1}{c|}{584} &
  \multicolumn{1}{c|}{156} &
  328 \\ \cline{2-20} 
\multicolumn{1}{|c|}{} &
  \textbf{40-49} &
  \multicolumn{1}{c|}{19380} &
  \multicolumn{1}{c|}{9690} &
  \multicolumn{1}{c|}{/} &
  7006 &
  \multicolumn{1}{c|}{3128} &
  \multicolumn{1}{c|}{2581} &
  \multicolumn{1}{c|}{3136} &
  \multicolumn{1}{c|}{852} &
  1721 &
  \multicolumn{1}{c|}{/} &
  \multicolumn{1}{c|}{/} &
  \multicolumn{1}{c|}{/} &
  \multicolumn{1}{c|}{/} &
  / &
  \multicolumn{1}{c|}{385} &
  \multicolumn{1}{c|}{208} &
  \multicolumn{1}{c|}{53} &
  124 \\ \cline{2-20} 
\multicolumn{1}{|c|}{} &
  \textbf{50-59} &
  \multicolumn{1}{c|}{/} &
  \multicolumn{1}{c|}{/} &
  \multicolumn{1}{c|}{/} &
  / &
  \multicolumn{1}{c|}{2484} &
  \multicolumn{1}{c|}{2093} &
  \multicolumn{1}{c|}{2513} &
  \multicolumn{1}{c|}{683} &
  1381 &
  \multicolumn{1}{c|}{/} &
  \multicolumn{1}{c|}{/} &
  \multicolumn{1}{c|}{/} &
  \multicolumn{1}{c|}{/} &
  / &
  \multicolumn{1}{c|}{161} &
  \multicolumn{1}{c|}{86} &
  \multicolumn{1}{c|}{20} &
  55 \\ \cline{2-20} 
\multicolumn{1}{|c|}{} &
  \textbf{60-69} &
  \multicolumn{1}{c|}{/} &
  \multicolumn{1}{c|}{/} &
  \multicolumn{1}{c|}{/} &
  / &
  \multicolumn{1}{c|}{1041} &
  \multicolumn{1}{c|}{834} &
  \multicolumn{1}{c|}{1026} &
  \multicolumn{1}{c|}{278} &
  571 &
  \multicolumn{1}{c|}{/} &
  \multicolumn{1}{c|}{/} &
  \multicolumn{1}{c|}{/} &
  \multicolumn{1}{c|}{/} &
  / &
  \multicolumn{1}{c|}{55} &
  \multicolumn{1}{c|}{27} &
  \multicolumn{1}{c|}{5} &
  23 \\ \cline{2-20} 
\multicolumn{1}{|c|}{} &
  \textbf{$\geq $ 70} &
  \multicolumn{1}{c|}{/} &
  \multicolumn{1}{c|}{/} &
  \multicolumn{1}{c|}{/} &
  / &
  \multicolumn{1}{c|}{259} &
  \multicolumn{1}{c|}{190} &
  \multicolumn{1}{c|}{242} &
  \multicolumn{1}{c|}{64} &
  143 &
  \multicolumn{1}{c|}{/} &
  \multicolumn{1}{c|}{/} &
  \multicolumn{1}{c|}{/} &
  \multicolumn{1}{c|}{/} &
  / &
  \multicolumn{1}{c|}{50} &
  \multicolumn{1}{c|}{26} &
  \multicolumn{1}{c|}{4} &
  20 \\ \hline
\multicolumn{2}{|c|}{\textbf{AU1}} &
  \multicolumn{1}{c|}{17556} &
  \multicolumn{1}{c|}{9425} &
  \multicolumn{1}{c|}{2082} &
  5868 &
  \multicolumn{1}{c|}{1452} &
  \multicolumn{1}{c|}{1170} &
  \multicolumn{1}{c|}{1973} &
  \multicolumn{1}{c|}{371} &
  587 &
  \multicolumn{1}{c|}{4011} &
  \multicolumn{1}{c|}{23540} &
  \multicolumn{1}{c|}{2999} &
  \multicolumn{1}{c|}{431} &
  1686 &
  \multicolumn{1}{c|}{1076} &
  \multicolumn{1}{c|}{595} &
  \multicolumn{1}{c|}{164} &
  307 \\ \hline
\multicolumn{2}{|c|}{\textbf{AU2}} &
  \multicolumn{1}{c|}{14728} &
  \multicolumn{1}{c|}{7482} &
  \multicolumn{1}{c|}{512} &
  6480 &
  \multicolumn{1}{c|}{689} &
  \multicolumn{1}{c|}{713} &
  \multicolumn{1}{c|}{1051} &
  \multicolumn{1}{c|}{247} &
  383 &
  \multicolumn{1}{c|}{14527} &
  \multicolumn{1}{c|}{3009} &
  \multicolumn{1}{c|}{8817} &
  \multicolumn{1}{c|}{2920} &
  5799 &
  \multicolumn{1}{c|}{795} &
  \multicolumn{1}{c|}{412} &
  \multicolumn{1}{c|}{121} &
  254 \\ \hline
\multicolumn{2}{|c|}{\textbf{AU4}} &
  \multicolumn{1}{c|}{49188} &
  \multicolumn{1}{c|}{25692} &
  \multicolumn{1}{c|}{3118} &
  19623 &
  \multicolumn{1}{c|}{2857} &
  \multicolumn{1}{c|}{610} &
  \multicolumn{1}{c|}{3941} &
  \multicolumn{1}{c|}{675} &
  1207 &
  \multicolumn{1}{c|}{3989} &
  \multicolumn{1}{c|}{836} &
  \multicolumn{1}{c|}{3671} &
  \multicolumn{1}{c|}{466} &
  688 &
  \multicolumn{1}{c|}{1817} &
  \multicolumn{1}{c|}{994} &
  \multicolumn{1}{c|}{235} &
  576 \\ \hline
\multicolumn{2}{|c|}{\textbf{AU5}} &
  \multicolumn{1}{c|}{5458} &
  \multicolumn{1}{c|}{4081} &
  \multicolumn{1}{c|}{302} &
  954 &
  \multicolumn{1}{c|}{875} &
  \multicolumn{1}{c|}{1287} &
  \multicolumn{1}{c|}{1252} &
  \multicolumn{1}{c|}{243} &
  424 &
  \multicolumn{1}{c|}{2600} &
  \multicolumn{1}{c|}{397} &
  \multicolumn{1}{c|}{1214} &
  \multicolumn{1}{c|}{166} &
  1617 &
  \multicolumn{1}{c|}{985} &
  \multicolumn{1}{c|}{599} &
  \multicolumn{1}{c|}{140} &
  225 \\ \hline
\multicolumn{2}{|c|}{\textbf{AU6}} &
  \multicolumn{1}{c|}{38968} &
  \multicolumn{1}{c|}{24160} &
  \multicolumn{1}{c|}{2866} &
  11067 &
  \multicolumn{1}{c|}{4572} &
  \multicolumn{1}{c|}{4777} &
  \multicolumn{1}{c|}{7289} &
  \multicolumn{1}{c|}{1395} &
  2809 &
  \multicolumn{1}{c|}{30787} &
  \multicolumn{1}{c|}{6826} &
  \multicolumn{1}{c|}{20954} &
  \multicolumn{1}{c|}{7037} &
  9622 &
  \multicolumn{1}{c|}{450} &
  \multicolumn{1}{c|}{218} &
  \multicolumn{1}{c|}{60} &
  166 \\ \hline
\multicolumn{2}{|c|}{\textbf{AU7}} &
  \multicolumn{1}{c|}{/} &
  \multicolumn{1}{c|}{/} &
  \multicolumn{1}{c|}{/} &
  / &
  \multicolumn{1}{c|}{/} &
  \multicolumn{1}{c|}{/} &
  \multicolumn{1}{c|}{/} &
  \multicolumn{1}{c|}{/} &
  / &
  \multicolumn{1}{c|}{49403} &
  \multicolumn{1}{c|}{11108} &
  \multicolumn{1}{c|}{34042} &
  \multicolumn{1}{c|}{10580} &
  15889 &
  \multicolumn{1}{c|}{/} &
  \multicolumn{1}{c|}{/} &
  \multicolumn{1}{c|}{/} &
  / \\ \hline
\multicolumn{2}{|c|}{\textbf{AU9}} &
  \multicolumn{1}{c|}{14264} &
  \multicolumn{1}{c|}{7510} &
  \multicolumn{1}{c|}{1668} &
  4606 &
  \multicolumn{1}{c|}{505} &
  \multicolumn{1}{c|}{143} &
  \multicolumn{1}{c|}{494} &
  \multicolumn{1}{c|}{62} &
  34 &
  \multicolumn{1}{c|}{1519} &
  \multicolumn{1}{c|}{24267} &
  \multicolumn{1}{c|}{1289} &
  \multicolumn{1}{c|}{95} &
  513 &
  \multicolumn{1}{c|}{774} &
  \multicolumn{1}{c|}{385} &
  \multicolumn{1}{c|}{88} &
  294 \\ \hline
\multicolumn{2}{|c|}{\textbf{AU10}} &
  \multicolumn{1}{c|}{/} &
  \multicolumn{1}{c|}{/} &
  \multicolumn{1}{c|}{/} &
  / &
  \multicolumn{1}{c|}{/} &
  \multicolumn{1}{c|}{/} &
  \multicolumn{1}{c|}{/} &
  \multicolumn{1}{c|}{/} &
  / &
  \multicolumn{1}{c|}{26740} &
  \multicolumn{1}{c|}{6079} &
  \multicolumn{1}{c|}{19032} &
  \multicolumn{1}{c|}{5786} &
  8001 &
  \multicolumn{1}{c|}{1390} &
  \multicolumn{1}{c|}{661} &
  \multicolumn{1}{c|}{205} &
  509 \\ \hline
\multicolumn{2}{|c|}{\textbf{AU11}} &
  \multicolumn{1}{c|}{/} &
  \multicolumn{1}{c|}{/} &
  \multicolumn{1}{c|}{/} &
  / &
  \multicolumn{1}{c|}{/} &
  \multicolumn{1}{c|}{/} &
  \multicolumn{1}{c|}{/} &
  \multicolumn{1}{c|}{/} &
  / &
  \multicolumn{1}{c|}{14926} &
  \multicolumn{1}{c|}{3710} &
  \multicolumn{1}{c|}{11117} &
  \multicolumn{1}{c|}{3022} &
  4497 &
  \multicolumn{1}{c|}{/} &
  \multicolumn{1}{c|}{/} &
  \multicolumn{1}{c|}{/} &
  / \\ \hline
\multicolumn{2}{|c|}{\textbf{AU12}} &
  \multicolumn{1}{c|}{61588} &
  \multicolumn{1}{c|}{33754} &
  \multicolumn{1}{c|}{8828} &
  18233 &
  \multicolumn{1}{c|}{7546} &
  \multicolumn{1}{c|}{6641} &
  \multicolumn{1}{c|}{12430} &
  \multicolumn{1}{c|}{2397} &
  5009 &
  \multicolumn{1}{c|}{32021} &
  \multicolumn{1}{c|}{7186} &
  \multicolumn{1}{c|}{21578} &
  \multicolumn{1}{c|}{7426} &
  10203 &
  \multicolumn{1}{c|}{1268} &
  \multicolumn{1}{c|}{584} &
  \multicolumn{1}{c|}{178} &
  488 \\ \hline
\multicolumn{2}{|c|}{\textbf{AU15}} &
  \multicolumn{1}{c|}{15724} &
  \multicolumn{1}{c|}{8631} &
  \multicolumn{1}{c|}{1878} &
  5206 &
  \multicolumn{1}{c|}{/} &
  \multicolumn{1}{c|}{/} &
  \multicolumn{1}{c|}{/} &
  \multicolumn{1}{c|}{/} &
  / &
  \multicolumn{1}{c|}{11536} &
  \multicolumn{1}{c|}{2350} &
  \multicolumn{1}{c|}{8617} &
  \multicolumn{1}{c|}{2231} &
  3038 &
  \multicolumn{1}{c|}{/} &
  \multicolumn{1}{c|}{/} &
  \multicolumn{1}{c|}{/} &
  / \\ \hline
\multicolumn{2}{|c|}{\textbf{AU16}} &
  \multicolumn{1}{c|}{/} &
  \multicolumn{1}{c|}{/} &
  \multicolumn{1}{c|}{/} &
  / &
  \multicolumn{1}{c|}{/} &
  \multicolumn{1}{c|}{/} &
  \multicolumn{1}{c|}{/} &
  \multicolumn{1}{c|}{/} &
  / &
  \multicolumn{1}{c|}{/} &
  \multicolumn{1}{c|}{/} &
  \multicolumn{1}{c|}{/} &
  \multicolumn{1}{c|}{/} &
  / &
  \multicolumn{1}{c|}{720} &
  \multicolumn{1}{c|}{302} &
  \multicolumn{1}{c|}{107} &
  299 \\ \hline
\multicolumn{2}{|c|}{\textbf{AU17}} &
  \multicolumn{1}{c|}{25860} &
  \multicolumn{1}{c|}{12127} &
  \multicolumn{1}{c|}{2290} &
  10931 &
  \multicolumn{1}{c|}{492} &
  \multicolumn{1}{c|}{184} &
  \multicolumn{1}{c|}{515} &
  \multicolumn{1}{c|}{57} &
  58 &
  \multicolumn{1}{c|}{32723} &
  \multicolumn{1}{c|}{7984} &
  \multicolumn{1}{c|}{20395} &
  \multicolumn{1}{c|}{6502} &
  13810 &
  \multicolumn{1}{c|}{541} &
  \multicolumn{1}{c|}{321} &
  \multicolumn{1}{c|}{85} &
  135 \\ \hline
\multicolumn{2}{|c|}{\textbf{AU20}} &
  \multicolumn{1}{c|}{9064} &
  \multicolumn{1}{c|}{5216} &
  \multicolumn{1}{c|}{484} &
  2909 &
  \multicolumn{1}{c|}{134} &
  \multicolumn{1}{c|}{146} &
  \multicolumn{1}{c|}{132} &
  \multicolumn{1}{c|}{17} &
  7 &
  \multicolumn{1}{c|}{/} &
  \multicolumn{1}{c|}{/} &
  \multicolumn{1}{c|}{/} &
  \multicolumn{1}{c|}{/} &
  / &
  \multicolumn{1}{c|}{/} &
  \multicolumn{1}{c|}{/} &
  \multicolumn{1}{c|}{/} &
  / \\ \hline
\multicolumn{2}{|c|}{\textbf{AU23}} &
  \multicolumn{1}{c|}{/} &
  \multicolumn{1}{c|}{/} &
  \multicolumn{1}{c|}{/} &
  / &
  \multicolumn{1}{c|}{/} &
  \multicolumn{1}{c|}{/} &
  \multicolumn{1}{c|}{/} &
  \multicolumn{1}{c|}{/} &
  / &
  \multicolumn{1}{c|}{27009} &
  \multicolumn{1}{c|}{6281} &
  \multicolumn{1}{c|}{17051} &
  \multicolumn{1}{c|}{5666} &
  10573 &
  \multicolumn{1}{c|}{/} &
  \multicolumn{1}{c|}{/} &
  \multicolumn{1}{c|}{/} &
  / \\ \hline
\multicolumn{2}{|c|}{\textbf{AU24}} &
  \multicolumn{1}{c|}{/} &
  \multicolumn{1}{c|}{/} &
  \multicolumn{1}{c|}{/} &
  / &
  \multicolumn{1}{c|}{/} &
  \multicolumn{1}{c|}{/} &
  \multicolumn{1}{c|}{/} &
  \multicolumn{1}{c|}{/} &
  / &
  \multicolumn{1}{c|}{15477} &
  \multicolumn{1}{c|}{3480} &
  \multicolumn{1}{c|}{10108} &
  \multicolumn{1}{c|}{1899} &
  6950 &
  \multicolumn{1}{c|}{/} &
  \multicolumn{1}{c|}{/} &
  \multicolumn{1}{c|}{/} &
  / \\ \hline
\multicolumn{2}{|c|}{\textbf{AU25}} &
  \multicolumn{1}{c|}{92104} &
  \multicolumn{1}{c|}{45461} &
  \multicolumn{1}{c|}{18938} &
  26691 &
  \multicolumn{1}{c|}{11590} &
  \multicolumn{1}{c|}{9473} &
  \multicolumn{1}{c|}{18155} &
  \multicolumn{1}{c|}{3500} &
  6880 &
  \multicolumn{1}{c|}{/} &
  \multicolumn{1}{c|}{/} &
  \multicolumn{1}{c|}{/} &
  \multicolumn{1}{c|}{/} &
  / &
  \multicolumn{1}{c|}{2829} &
  \multicolumn{1}{c|}{1478} &
  \multicolumn{1}{c|}{408} &
  901 \\ \hline
\multicolumn{2}{|c|}{\textbf{AU26}} &
  \multicolumn{1}{c|}{211676} &
  \multicolumn{1}{c|}{26465} &
  \multicolumn{1}{c|}{12532} &
  10440 &
  \multicolumn{1}{c|}{2058} &
  \multicolumn{1}{c|}{1778} &
  \multicolumn{1}{c|}{3146} &
  \multicolumn{1}{c|}{651} &
  1156 &
  \multicolumn{1}{c|}{/} &
  \multicolumn{1}{c|}{/} &
  \multicolumn{1}{c|}{/} &
  \multicolumn{1}{c|}{/} &
  / &
  \multicolumn{1}{c|}{1089} &
  \multicolumn{1}{c|}{578} &
  \multicolumn{1}{c|}{160} &
  334 \\ \hline
\multicolumn{2}{|c|}{\textbf{AU27}} &
  \multicolumn{1}{c|}{/} &
  \multicolumn{1}{c|}{/} &
  \multicolumn{1}{c|}{/} &
  / &
  \multicolumn{1}{c|}{/} &
  \multicolumn{1}{c|}{/} &
  \multicolumn{1}{c|}{/} &
  \multicolumn{1}{c|}{/} &
  / &
  \multicolumn{1}{c|}{/} &
  \multicolumn{1}{c|}{/} &
  \multicolumn{1}{c|}{/} &
  \multicolumn{1}{c|}{/} &
  / &
  \multicolumn{1}{c|}{810} &
  \multicolumn{1}{c|}{382} &
  \multicolumn{1}{c|}{98} &
  313 \\ \hline
\end{tabular}}
\end{table*}

\begin{table}[]
\centering
\caption{Age Distribution of the Original and New Partitions of GFT }\label{table1}
\setlength{\tabcolsep}{1.2mm}
\renewcommand{\arraystretch}{1.2}
\begin{tabular}{|c|c|cccccccc|}
\hline
\multirow{3}{*}{\textbf{Part.}} &
  \multirow{3}{*}{\textbf{Set}} &
  \multicolumn{8}{c|}{\textbf{Age}} \\ \cline{3-10} 
 &
   &
  \multicolumn{1}{c|}{\multirow{2}{*}{\textbf{21}}} &
  \multicolumn{1}{c|}{\multirow{2}{*}{\textbf{22}}} &
  \multicolumn{1}{c|}{\multirow{2}{*}{\textbf{23}}} &
  \multicolumn{1}{c|}{\multirow{2}{*}{\textbf{24}}} &
  \multicolumn{1}{c|}{\multirow{2}{*}{\textbf{25}}} &
  \multicolumn{1}{c|}{\multirow{2}{*}{\textbf{26}}} &
  \multicolumn{1}{c|}{\multirow{2}{*}{\textbf{27}}} &
  \multirow{2}{*}{\textbf{28}} \\
 &
   &
  \multicolumn{1}{c|}{} &
  \multicolumn{1}{c|}{} &
  \multicolumn{1}{c|}{} &
  \multicolumn{1}{c|}{} &
  \multicolumn{1}{c|}{} &
  \multicolumn{1}{c|}{} &
  \multicolumn{1}{c|}{} &
   \\ \hline
\multirow{2}{*}{\textbf{Org.}} &
  Train &
  \multicolumn{1}{c|}{57179} &
  \multicolumn{1}{c|}{18644} &
  \multicolumn{1}{c|}{14004} &
  \multicolumn{1}{c|}{1486} &
  \multicolumn{1}{c|}{3896} &
  \multicolumn{1}{c|}{5797} &
  \multicolumn{1}{c|}{3274} &
  4269 \\ \cline{2-10} 
 &
  Test &
  \multicolumn{1}{c|}{9967} &
  \multicolumn{1}{c|}{10300} &
  \multicolumn{1}{c|}{916} &
  \multicolumn{1}{c|}{1249} &
  \multicolumn{1}{c|}{/} &
  \multicolumn{1}{c|}{/} &
  \multicolumn{1}{c|}{/} &
  2213 \\ \hline
\multirow{3}{*}{\textbf{New}} &
  Train &
  \multicolumn{1}{c|}{35408} &
  \multicolumn{1}{c|}{14918} &
  \multicolumn{1}{c|}{8698} &
  \multicolumn{1}{c|}{1486} &
  \multicolumn{1}{c|}{2237} &
  \multicolumn{1}{c|}{2570} &
  \multicolumn{1}{c|}{1618} &
  2683 \\ \cline{2-10} 
 &
  Valid &
  \multicolumn{1}{c|}{14145} &
  \multicolumn{1}{c|}{4818} &
  \multicolumn{1}{c|}{3096} &
  \multicolumn{1}{c|}{/} &
  \multicolumn{1}{c|}{/} &
  \multicolumn{1}{c|}{1626} &
  \multicolumn{1}{c|}{/} &
  / \\ \cline{2-10} 
 &
  Test &
  \multicolumn{1}{c|}{17593} &
  \multicolumn{1}{c|}{9208} &
  \multicolumn{1}{c|}{3126} &
  \multicolumn{1}{c|}{1249} &
  \multicolumn{1}{c|}{1659} &
  \multicolumn{1}{c|}{1601} &
  \multicolumn{1}{c|}{1656} &
  3799 \\ \hline
\end{tabular}
\end{table}


\subsection{Performance Measures}

In the following, we present the performance measures that we selected for 
validating the performance and fairness of the models on each task. \\

\noindent
\underline{1) Expression Recognition} 
\par

\textbf{F1 Score:} It represents the average F1 Score across all expression categories (i.e., macro F1 Score). 
It takes values in [0, 1]. Higher values are desirable in evaluation. 
\vspace{0.1cm}

\textbf{Statistical Parity (SP):} It quantifies the extent of disparity in model predictions across different subgroups within a demographic attribute. It is the average of absolute differences in predicted probabilities between any distinct pairs of subgroups with the same class prediction. For each demographic attribute, SP is:
\vspace{-0.6em}
\begin{align*}
\frac{2 \sum\limits_{c=1}^k \sum\limits_{i=1}^n \sum\limits_{j=i+1}^n\left|p\left(\mathbf{\hat{y}}=c \mid \mathbf{s}=s_i,\mathbf{x}\right)-p\left(\mathbf{\hat{y}}=c \mid \mathbf{s}=s_j,\mathbf{x}\right)\right|}{n(n-1)},
\end{align*}
where:
$s_i$ and $s_j$ represent distinct subgroups of a demographic attribute (e.g. $s_i$ is 'Asian', $s_j$ is 'Black' and the demographic attribute is 'Race'); 
$ p(\mathbf{\hat{y}}=c \mid \mathbf{s}=s_i,\mathbf{x}) $ denotes the probability of predicting class $c$ with sample $\mathbf{x}$ belonging to subgroup $s_i$; 
$k$ represents the number of expression categories; 
$n$ represents the number of subgroups of a demographic attribute.
SP takes values in [0, 1]. A value of 0 is desirable; values in  $[0, 
 0.1]$ indicate fair models. \\



\noindent
\underline{2) Action Unit Detection} 

\textbf{F1 Score:} It represents the average F1 Score across all AUs (i.e., binary F1 score). It takes values
in [0, 1]. Higher values are desirable in evaluation. 



\vspace{0.1cm}

\textbf{Demographic Parity Difference(DPD):} It measures the disparity in model predictions across different demographic subgroups. For each demographic attribute, DPD is:

\begin{figure}[h!]
\centering
  \includegraphics[width=0.49\textwidth]{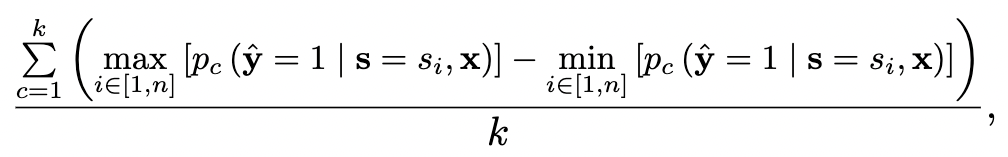}
 \\
\end{figure}

\noindent where: $k$ represents the number of AUs; $n$ denotes the number of subgroups within a demographic attribute; $p_c(\hat{\mathbf{y}}=1 \mid \mathbf{s}=s_i, \mathbf{x})$ denotes the probability of predicting activation of $AU_c$ with sample $ \mathbf{x} $ belonging to the subgroup $ s_i $; $\max\limits_{i \in[1, n]}$ and $\min\limits_{i \in[1, n]}$ represent the maximum and minimum probability across $n$ demographic subgroups, respectively. DPD takes values in [0, 1]. A value of 0 is desirable; values in  $[0, 0.1]$ indicate fair models. \\

\noindent
\underline{3) Valence-Arousal Estimation} 
\vspace{0.5em}
\par

\textbf{Concordance Correlation Coefficient (CCC):} It quantifies the agreement between predicted values and their annotations. Mathematically, CCC is:
$$
\frac{2 \cdot s_{xy}}{s_x^2 + s_y^2 + (\bar{x} - \bar{y})^2},
$$ 
where: $s_x$ and $s_y$ represent the variances of the valence/arousal annotations and predicted values, respectively; $\bar{x}$ and $\bar{y}$ denote the mean values of the corresponding annotations and predictions; $s_{xy}$ represents their covariance. CCC takes values in [-1, 1]. Higher values are desirable in evaluation. We compute the average of the CCC values for valence and arousal.
\vspace{0.1cm}

\textbf{Average CCC:} Evaluates the alignment between predicted values and annotations across diverse demographic subgroups. It is computed by averaging the mean CCC values for valence and arousal across all subgroups:
$$
\text{Average CCC} = \frac{1}{2n} \sum_{i=1}^{n} \left( CCC_{\text{valence}}^{(i)} + CCC_{\text{arousal}}^{(i)} \right)
$$
where: $i$ indicates the index of $n$ subgroups within a demographic attribute; $CCC_{\text{valence}}^{(i)}$ and $CCC_{\text{arousal}}^{(i)}$ are computed following the method of CCC defined previously. 

\section{EXPERIMENTAL RESULTS}





\begin{table*}[]
\centering
\caption{Performance comparison (in \%) between various baseline and state-of-the-art works for FER; for F1 higher values are wanted; for SP, a value of 0 is wanted, values in $[0,10]$ indicate fair models
%
}\label{table4}
\renewcommand{\arraystretch}{1.1}
\resizebox{\textwidth}{!}{
\begin{tabular}{|c|ccccccc|ccccccc|ccccccc|}
\hline
\multirow{3}{*}{\textbf{Model}} &
  \multicolumn{7}{c|}{\textbf{AffectNet-7}} &
  \multicolumn{7}{c|}{\textbf{AffectNet-8}} &
  \multicolumn{7}{c|}{\textbf{RAF-DB}} \\ \cline{2-22} 
 &
  \multicolumn{1}{c|}{\textbf{Test}} &
  \multicolumn{2}{c|}{\textbf{Age}} &
  \multicolumn{2}{c|}{\textbf{Gender}} &
  \multicolumn{2}{c|}{\textbf{Race}} &
  \multicolumn{1}{c|}{\textbf{Test}} &
  \multicolumn{2}{c|}{\textbf{Age}} &
  \multicolumn{2}{c|}{\textbf{Gender}} &
  \multicolumn{2}{c|}{\textbf{Race}} &
  \multicolumn{1}{c|}{\textbf{Test}} &
  \multicolumn{2}{c|}{\textbf{Age}} &
  \multicolumn{2}{c|}{\textbf{Gender}} &
  \multicolumn{2}{c|}{\textbf{Race}} \\ \cline{2-22} 
 &
  \multicolumn{1}{c|}{\textbf{F1}} &
  \multicolumn{1}{c|}{\textbf{SP}} &
  \multicolumn{1}{c|}{\textbf{F1}} &
  \multicolumn{1}{c|}{\textbf{SP}} &
  \multicolumn{1}{c|}{\textbf{F1}} &
  \multicolumn{1}{c|}{\textbf{SP}} &
  \textbf{F1} &
  \multicolumn{1}{c|}{\textbf{F1}} &
  \multicolumn{1}{c|}{\textbf{SP}} &
  \multicolumn{1}{c|}{\textbf{F1}} &
  \multicolumn{1}{c|}{\textbf{SP}} &
  \multicolumn{1}{c|}{\textbf{F1}} &
  \multicolumn{1}{c|}{\textbf{SP}} &
  \textbf{F1} &
  \multicolumn{1}{c|}{\textbf{F1}} &
  \multicolumn{1}{c|}{\textbf{SP}} &
  \multicolumn{1}{c|}{\textbf{F1}} &
  \multicolumn{1}{c|}{\textbf{SP}} &
  \multicolumn{1}{c|}{\textbf{F1}} &
  \multicolumn{1}{c|}{\textbf{SP}} &
  \textbf{F1} \\ \hline
\textbf{ResNet18} &
  \multicolumn{1}{c|}{58.8} &
  \multicolumn{1}{c|}{1.5} &
  \multicolumn{1}{c|}{55.9} &
  \multicolumn{1}{c|}{0.9} &
  \multicolumn{1}{c|}{56.2} &
  \multicolumn{1}{c|}{1.9} &
  56.6 &
  \multicolumn{1}{c|}{51.7} &
  \multicolumn{1}{c|}{16.4} &
  \multicolumn{1}{c|}{49.5} &
  \multicolumn{1}{c|}{18.0} &
  \multicolumn{1}{c|}{51.5} &
  \multicolumn{1}{c|}{7.2} &
  50.5 &
  \multicolumn{1}{c|}{65.7} &
  \multicolumn{1}{c|}{24.1} &
  \multicolumn{1}{c|}{57.5} &
  \multicolumn{1}{c|}{8.9} &
  \multicolumn{1}{c|}{64.8} &
  \multicolumn{1}{c|}{6.4} &
  63.2 \\ \hline
\textbf{ResNet50} &
  \multicolumn{1}{c|}{58.8} &
  \multicolumn{1}{c|}{1.8} &
  \multicolumn{1}{c|}{57.4} &
  \multicolumn{1}{c|}{0.8} &
  \multicolumn{1}{c|}{57.7} &
  \multicolumn{1}{c|}{1.6} &
  58.4 &
  \multicolumn{1}{c|}{51.4} &
  \multicolumn{1}{c|}{17.0} &
  \multicolumn{1}{c|}{49.5} &
  \multicolumn{1}{c|}{17.7} &
  \multicolumn{1}{c|}{51.1} &
  \multicolumn{1}{c|}{7.3} &
  50.9 &
  \multicolumn{1}{c|}{59.5} &
  \multicolumn{1}{c|}{23.5} &
  \multicolumn{1}{c|}{51.1} &
  \multicolumn{1}{c|}{10.0} &
  \multicolumn{1}{c|}{57.3} &
  \multicolumn{1}{c|}{7.0} &
  54.0 \\ \hline
\textbf{ResNext50} &
  \multicolumn{1}{c|}{58.6} &
  \multicolumn{1}{c|}{2.1} &
  \multicolumn{1}{c|}{58.1} &
  \multicolumn{1}{c|}{0.7} &
  \multicolumn{1}{c|}{58.6} &
  \multicolumn{1}{c|}{1.7} &
  59.0 &
  \multicolumn{1}{c|}{51.8} &
  \multicolumn{1}{c|}{15.9} &
  \multicolumn{1}{c|}{49.1} &
  \multicolumn{1}{c|}{16.5} &
  \multicolumn{1}{c|}{51.0} &
  \multicolumn{1}{c|}{6.3} &
  50.0 &
  \multicolumn{1}{c|}{56.1} &
  \multicolumn{1}{c|}{22.9} &
  \multicolumn{1}{c|}{51.3} &
  \multicolumn{1}{c|}{8.8} &
  \multicolumn{1}{c|}{56.0} &
  \multicolumn{1}{c|}{5.0} &
  53.8 \\ \hline
\textbf{DenseNet121} &
  \multicolumn{1}{c|}{59.4} &
  \multicolumn{1}{c|}{1.9} &
  \multicolumn{1}{c|}{59.1} &
  \multicolumn{1}{c|}{0.3} &
  \multicolumn{1}{c|}{59.4} &
  \multicolumn{1}{c|}{1.3} &
  60.2 &
  \multicolumn{1}{c|}{52.8} &
  \multicolumn{1}{c|}{15.9} &
  \multicolumn{1}{c|}{51.1} &
  \multicolumn{1}{c|}{16.2} &
  \multicolumn{1}{c|}{52.6} &
  \multicolumn{1}{c|}{6.5} &
  52.0 &
  \multicolumn{1}{c|}{64.1} &
  \multicolumn{1}{c|}{23.6} &
  \multicolumn{1}{c|}{56.8} &
  \multicolumn{1}{c|}{9.6} &
  \multicolumn{1}{c|}{63.1} &
  \multicolumn{1}{c|}{7.0} &
  60.8 \\ \hline
\textbf{ViT\_B\_16} &
  \multicolumn{1}{c|}{58.0} &
  \multicolumn{1}{c|}{1.8} &
  \multicolumn{1}{c|}{58.1} &
  \multicolumn{1}{c|}{0.7} &
  \multicolumn{1}{c|}{58.1} &
  \multicolumn{1}{c|}{2.0} &
  58.5 &
  \multicolumn{1}{c|}{51.3} &
  \multicolumn{1}{c|}{14.7} &
  \multicolumn{1}{c|}{49.2} &
  \multicolumn{1}{c|}{17.0} &
  \multicolumn{1}{c|}{50.9} &
  \multicolumn{1}{c|}{6.8} &
  49.8 &
  \multicolumn{1}{c|}{71.4} &
  \multicolumn{1}{c|}{24.6} &
  \multicolumn{1}{c|}{64.8} &
  \multicolumn{1}{c|}{11.4} &
  \multicolumn{1}{c|}{71.1} &
  \multicolumn{1}{c|}{7.7} &
  68.5 \\ \hline
\textbf{VGG16} &
  \multicolumn{1}{c|}{58.3} &
  \multicolumn{1}{c|}{1.8} &
  \multicolumn{1}{c|}{57.2} &
  \multicolumn{1}{c|}{0.5} &
  \multicolumn{1}{c|}{57.5} &
  \multicolumn{1}{c|}{2.0} &
  57.6 &
  \multicolumn{1}{c|}{51.1} &
  \multicolumn{1}{c|}{16.5} &
  \multicolumn{1}{c|}{49.1} &
  \multicolumn{1}{c|}{17.4} &
  \multicolumn{1}{c|}{50.9} &
  \multicolumn{1}{c|}{7.0} &
  50.1 &
  \multicolumn{1}{c|}{67.1} &
  \multicolumn{1}{c|}{23.3} &
  \multicolumn{1}{c|}{59.7} &
  \multicolumn{1}{c|}{9.7} &
  \multicolumn{1}{c|}{65.9} &
  \multicolumn{1}{c|}{8.5} &
  63.9 \\ \hline
\textbf{EffNet\_B0} &
  \multicolumn{1}{c|}{59.3} &
  \multicolumn{1}{c|}{2.0} &
  \multicolumn{1}{c|}{59.3} &
  \multicolumn{1}{c|}{0.9} &
  \multicolumn{1}{c|}{59.3} &
  \multicolumn{1}{c|}{1.6} &
  59.8 &
  \multicolumn{1}{c|}{53.1} &
  \multicolumn{1}{c|}{15.3} &
  \multicolumn{1}{c|}{49.4} &
  \multicolumn{1}{c|}{17.5} &
  \multicolumn{1}{c|}{51.7} &
  \multicolumn{1}{c|}{6.2} &
  51.2 &
  \multicolumn{1}{c|}{66.5} &
  \multicolumn{1}{c|}{23.5} &
  \multicolumn{1}{c|}{59.5} &
  \multicolumn{1}{c|}{9.9} &
  \multicolumn{1}{c|}{65.4} &
  \multicolumn{1}{c|}{7.1} &
  63.6 \\ \hline
\textbf{EffNet\_B7} &
  \multicolumn{1}{c|}{59.8} &
  \multicolumn{1}{c|}{1.9} &
  \multicolumn{1}{c|}{59.3} &
  \multicolumn{1}{c|}{0.7} &
  \multicolumn{1}{c|}{59.8} &
  \multicolumn{1}{c|}{1.6} &
  60.2 &
  \multicolumn{1}{c|}{52.9} &
  \multicolumn{1}{c|}{15.7} &
  \multicolumn{1}{c|}{49.7} &
  \multicolumn{1}{c|}{17.2} &
  \multicolumn{1}{c|}{51.7} &
  \multicolumn{1}{c|}{6.3} &
  50.9 &
  \multicolumn{1}{c|}{70.1} &
  \multicolumn{1}{c|}{22.5} &
  \multicolumn{1}{c|}{62.1} &
  \multicolumn{1}{c|}{10.8} &
  \multicolumn{1}{c|}{68.0} &
  \multicolumn{1}{c|}{7.5} &
  66.8 \\ \hline
\textbf{Swin\_B} &
  \multicolumn{1}{c|}{59.7} &
  \multicolumn{1}{c|}{2.1} &
  \multicolumn{1}{c|}{59.0} &
  \multicolumn{1}{c|}{0.5} &
  \multicolumn{1}{c|}{58.8} &
  \multicolumn{1}{c|}{1.6} &
  59.2 &
  \multicolumn{1}{c|}{52.1} &
  \multicolumn{1}{c|}{16.1} &
  \multicolumn{1}{c|}{50.9} &
  \multicolumn{1}{c|}{17.7} &
  \multicolumn{1}{c|}{52.4} &
  \multicolumn{1}{c|}{7.4} &
  51.6 &
  \multicolumn{1}{c|}{74.2} &
  \multicolumn{1}{c|}{23.6} &
  \multicolumn{1}{c|}{67.5} &
  \multicolumn{1}{c|}{11.5} &
  \multicolumn{1}{c|}{73.7} &
  \multicolumn{1}{c|}{8.0} &
  72.7 \\ \hline
\textbf{Swin\_V2\_B} &
  \multicolumn{1}{c|}{59.2} &
  \multicolumn{1}{c|}{2.0} &
  \multicolumn{1}{c|}{58.7} &
  \multicolumn{1}{c|}{0.6} &
  \multicolumn{1}{c|}{58.3} &
  \multicolumn{1}{c|}{1.5} &
  58.9 &
  \multicolumn{1}{c|}{52.0} &
  \multicolumn{1}{c|}{15.9} &
  \multicolumn{1}{c|}{50.5} &
  \multicolumn{1}{c|}{18.2} &
  \multicolumn{1}{c|}{51.9} &
  \multicolumn{1}{c|}{7.5} &
  51.4 &
  \multicolumn{1}{c|}{74.1} &
  \multicolumn{1}{c|}{24.1} &
  \multicolumn{1}{c|}{66.9} &
  \multicolumn{1}{c|}{10.7} &
  \multicolumn{1}{c|}{72.6} &
  \multicolumn{1}{c|}{9.2} &
  72.0 \\ \hline
\textbf{ConvNeXt\_Base} &
  \multicolumn{1}{c|}{60.8} &
  \multicolumn{1}{c|}{2.0} &
  \multicolumn{1}{c|}{60.2} &
  \multicolumn{1}{c|}{1.0} &
  \multicolumn{1}{c|}{60.5} &
  \multicolumn{1}{c|}{1.4} &
  60.9 &
  \multicolumn{1}{c|}{53.9} &
  \multicolumn{1}{c|}{15.4} &
  \multicolumn{1}{c|}{51.7} &
  \multicolumn{1}{c|}{18.9} &
  \multicolumn{1}{c|}{53.7} &
  \multicolumn{1}{c|}{7.4} &
  52.9 &
  \multicolumn{1}{c|}{73.2} &
  \multicolumn{1}{c|}{24.0} &
  \multicolumn{1}{c|}{67.8} &
  \multicolumn{1}{c|}{12.2} &
  \multicolumn{1}{c|}{72.4} &
  \multicolumn{1}{c|}{6.9} &
  70.3 \\ \hline
\textbf{iResNet101} &
  \multicolumn{1}{c|}{59.6} &
  \multicolumn{1}{c|}{1.7} &
  \multicolumn{1}{c|}{59.5} &
  \multicolumn{1}{c|}{0.6} &
  \multicolumn{1}{c|}{59.6} &
  \multicolumn{1}{c|}{1.6} &
  59.9 &
  \multicolumn{1}{c|}{52.0} &
  \multicolumn{1}{c|}{15.2} &
  \multicolumn{1}{c|}{49.1} &
  \multicolumn{1}{c|}{18.2} &
  \multicolumn{1}{c|}{51.3} &
  \multicolumn{1}{c|}{6.3} &
  49.9 &
  \multicolumn{1}{c|}{67.9} &
  \multicolumn{1}{c|}{24.0} &
  \multicolumn{1}{c|}{61.0} &
  \multicolumn{1}{c|}{9.7} &
  \multicolumn{1}{c|}{67.4} &
  \multicolumn{1}{c|}{7.5} &
  65.3 \\ \hline
\textbf{POSTER++} \cite{poster} &
  \multicolumn{1}{c|}{61.4} &
  \multicolumn{1}{c|}{2.0} &
  \multicolumn{1}{c|}{60.9} &
  \multicolumn{1}{c|}{0.7} &
  \multicolumn{1}{c|}{61.3} &
  \multicolumn{1}{c|}{1.8} &
  61.6 &
  \multicolumn{1}{c|}{55.4} &
  \multicolumn{1}{c|}{20.7} &
  \multicolumn{1}{c|}{52.8} &
  \multicolumn{1}{c|}{18.0} &
  \multicolumn{1}{c|}{55.2} &
  \multicolumn{1}{c|}{8.4} &
  54.9 &
  \multicolumn{1}{c|}{83.2} &
  \multicolumn{1}{c|}{24.4} &
  \multicolumn{1}{c|}{80.3} &
  \multicolumn{1}{c|}{11.5} &
  \multicolumn{1}{c|}{82.7} &
  \multicolumn{1}{c|}{7.9} &
  80.9 \\ \hline
\textbf{DAN} \cite{dan} &
  \multicolumn{1}{c|}{58.9} &
  \multicolumn{1}{c|}{2.2} &
  \multicolumn{1}{c|}{56.7} &
  \multicolumn{1}{c|}{0.5} &
  \multicolumn{1}{c|}{57.0} &
  \multicolumn{1}{c|}{2.2} &
  57.6 &
  \multicolumn{1}{c|}{51.4} &
  \multicolumn{1}{c|}{22.5} &
  \multicolumn{1}{c|}{47.7} &
  \multicolumn{1}{c|}{18.0} &
  \multicolumn{1}{c|}{49.6} &
  \multicolumn{1}{c|}{10.1} &
  49.9 &
  \multicolumn{1}{c|}{77.5} &
  \multicolumn{1}{c|}{24.3} &
  \multicolumn{1}{c|}{71.9} &
  \multicolumn{1}{c|}{12.1} &
  \multicolumn{1}{c|}{77.1} &
  \multicolumn{1}{c|}{8.1} &
  74.1 \\ \hline
\textbf{MT-EffNet} \cite{mteffnet}&
  \multicolumn{1}{c|}{57.1} &
  \multicolumn{1}{c|}{2.2} &
  \multicolumn{1}{c|}{57.1} &
  \multicolumn{1}{c|}{0.6} &
  \multicolumn{1}{c|}{57.1} &
  \multicolumn{1}{c|}{2.0} &
  57.5 &
  \multicolumn{1}{c|}{52.0} &
  \multicolumn{1}{c|}{16.2} &
  \multicolumn{1}{c|}{49.9} &
  \multicolumn{1}{c|}{18.1} &
  \multicolumn{1}{c|}{51.8} &
  \multicolumn{1}{c|}{6.9} &
  51.5 &
  \multicolumn{1}{c|}{72.6} &
  \multicolumn{1}{c|}{24.9} &
  \multicolumn{1}{c|}{68.1} &
  \multicolumn{1}{c|}{12.7} &
  \multicolumn{1}{c|}{71.6} &
  \multicolumn{1}{c|}{8.3} &
  70.4 \\ \hline
\textbf{MA-Net} \cite{manet} &
  \multicolumn{1}{c|}{62.0} &
  \multicolumn{1}{c|}{1.9} &
  \multicolumn{1}{c|}{61.9} &
  \multicolumn{1}{c|}{0.6} &
  \multicolumn{1}{c|}{62.0} &
  \multicolumn{1}{c|}{1.2} &
  62.4 &
  \multicolumn{1}{c|}{54.2} &
  \multicolumn{1}{c|}{15.2} &
  \multicolumn{1}{c|}{52.3} &
  \multicolumn{1}{c|}{17.2} &
  \multicolumn{1}{c|}{54.0} &
  \multicolumn{1}{c|}{6.5} &
  53.2 &
  \multicolumn{1}{c|}{77.2} &
  \multicolumn{1}{c|}{23.4} &
  \multicolumn{1}{c|}{72.4} &
  \multicolumn{1}{c|}{11.7} &
  \multicolumn{1}{c|}{76.5} &
  \multicolumn{1}{c|}{8.5} &
  75.1 \\ \hline
\textbf{EAC} \cite{eac} &
  \multicolumn{1}{c|}{62.4} &
  \multicolumn{1}{c|}{2.1} &
  \multicolumn{1}{c|}{61.6} &
  \multicolumn{1}{c|}{0.6} &
  \multicolumn{1}{c|}{62.4} &
  \multicolumn{1}{c|}{1.2} &
  62.7 &
  \multicolumn{1}{c|}{55.5} &
  \multicolumn{1}{c|}{14.8} &
  \multicolumn{1}{c|}{53.5} &
  \multicolumn{1}{c|}{17.2} &
  \multicolumn{1}{c|}{55.3} &
  \multicolumn{1}{c|}{6.5} &
  54.0 &
  \multicolumn{1}{c|}{81.0} &
  \multicolumn{1}{c|}{23.3} &
  \multicolumn{1}{c|}{76.5} &
  \multicolumn{1}{c|}{12.3} &
  \multicolumn{1}{c|}{80.1} &
  \multicolumn{1}{c|}{8.9} &
  79.1 \\ \hline
\textbf{DACL} \cite{dacl}&
  \multicolumn{1}{c|}{60.0} &
  \multicolumn{1}{c|}{2.0} &
  \multicolumn{1}{c|}{60.2} &
  \multicolumn{1}{c|}{0.7} &
  \multicolumn{1}{c|}{60.0} &
  \multicolumn{1}{c|}{1.8} &
  60.4 &
  \multicolumn{1}{c|}{52.3} &
  \multicolumn{1}{c|}{15.6} &
  \multicolumn{1}{c|}{49.9} &
  \multicolumn{1}{c|}{17.3} &
  \multicolumn{1}{c|}{52.1} &
  \multicolumn{1}{c|}{6.9} &
  51.2 &
  \multicolumn{1}{c|}{74.2} &
  \multicolumn{1}{c|}{23.8} &
  \multicolumn{1}{c|}{69.7} &
  \multicolumn{1}{c|}{11.2} &
  \multicolumn{1}{c|}{73.3} &
  \multicolumn{1}{c|}{8.0} &
  71.7 \\ \hline
\textbf{EmoGCN} \cite{emogcn}&
  \multicolumn{1}{c|}{58.1} &
  \multicolumn{1}{c|}{2.7} &
  \multicolumn{1}{c|}{58.2} &
  \multicolumn{1}{c|}{0.8} &
  \multicolumn{1}{c|}{58.0} &
  \multicolumn{1}{c|}{1.7} &
  57.4 &
  \multicolumn{1}{c|}{51.9} &
  \multicolumn{1}{c|}{20} &
  \multicolumn{1}{c|}{47.8} &
  \multicolumn{1}{c|}{17.7} &
  \multicolumn{1}{c|}{51.3} &
  \multicolumn{1}{c|}{8.7} &
  50.1 &
  \multicolumn{1}{c|}{70.5} &
  \multicolumn{1}{c|}{24.1} &
  \multicolumn{1}{c|}{59.7} &
  \multicolumn{1}{c|}{11.2} &
  \multicolumn{1}{c|}{67.9} &
  \multicolumn{1}{c|}{12.3} &
  66.4 \\ \hline
\textbf{FUXI} \cite{fuxi}&
  \multicolumn{1}{c|}{59.6} &
  \multicolumn{1}{c|}{2.1} &
  \multicolumn{1}{c|}{59.7} &
  \multicolumn{1}{c|}{0.6} &
  \multicolumn{1}{c|}{59.6} &
  \multicolumn{1}{c|}{1.8} &
  60.0 &
  \multicolumn{1}{c|}{52.8} &
  \multicolumn{1}{c|}{20.7} &
  \multicolumn{1}{c|}{50.0} &
  \multicolumn{1}{c|}{18.2} &
  \multicolumn{1}{c|}{51.2} &
  \multicolumn{1}{c|}{9.6} &
  51.4 &
  \multicolumn{1}{c|}{77.7} &
  \multicolumn{1}{c|}{25.1} &
  \multicolumn{1}{c|}{73.0} &
  \multicolumn{1}{c|}{11.8} &
  \multicolumn{1}{c|}{76.6} &
  \multicolumn{1}{c|}{7.7} &
  75.5 \\ \hline
\textbf{SITU} \cite{situ}&
  \multicolumn{1}{c|}{59.2} &
  \multicolumn{1}{c|}{2.0} &
  \multicolumn{1}{c|}{59.2} &
  \multicolumn{1}{c|}{0.5} &
  \multicolumn{1}{c|}{59.2} &
  \multicolumn{1}{c|}{1.3} &
  59.6 &
  \multicolumn{1}{c|}{53.8} &
  \multicolumn{1}{c|}{15.9} &
  \multicolumn{1}{c|}{51.9} &
  \multicolumn{1}{c|}{17.7} &
  \multicolumn{1}{c|}{53.6} &
  \multicolumn{1}{c|}{7.7} &
  52.6 &
  \multicolumn{1}{c|}{73.6} &
  \multicolumn{1}{c|}{24.8} &
  \multicolumn{1}{c|}{70.1} &
  \multicolumn{1}{c|}{10.4} &
  \multicolumn{1}{c|}{72.8} &
  \multicolumn{1}{c|}{7.5} &
  73.2 \\ \hline
\textbf{CTC} \cite{ctc} &
  \multicolumn{1}{c|}{58.7} &
  \multicolumn{1}{c|}{1.9} &
  \multicolumn{1}{c|}{56.6} &
  \multicolumn{1}{c|}{0.6} &
  \multicolumn{1}{c|}{56.9} &
  \multicolumn{1}{c|}{1.2} &
  57.1 &
  \multicolumn{1}{c|}{53.2} &
  \multicolumn{1}{c|}{15.6} &
  \multicolumn{1}{c|}{50.2} &
  \multicolumn{1}{c|}{16.8} &
  \multicolumn{1}{c|}{51.8} &
  \multicolumn{1}{c|}{6.9} &
  51.4 &
  \multicolumn{1}{c|}{75.6} &
  \multicolumn{1}{c|}{24.2} &
  \multicolumn{1}{c|}{72.2} &
  \multicolumn{1}{c|}{12.4} &
  \multicolumn{1}{c|}{76.2} &
  \multicolumn{1}{c|}{7.9} &
  75.7 \\ \hline
\end{tabular}}
\end{table*}




\begin{table*}[]
\scriptsize
\centering
\caption{Performance comparison (in \%) between various baseline and state-of-the-art works for AU detection; for F1 higher values are wanted; for DPD, a value of 0 is wanted, values in red {[0,10]}  indicate fair models}\label{table5}
\setlength{\tabcolsep}{2.5mm}
\renewcommand{\arraystretch}{1.1}
\scalebox{1.05}{
\begin{tabular}{|c|ccccccc|ccccccc|}
\hline
\multirow{3}{*}{\textbf{Model}} &
  \multicolumn{7}{c|}{\textbf{DISFA}} &
  \multicolumn{7}{c|}{\textbf{EmotioNet}} \\ \cline{2-15} 
 &
  \multicolumn{1}{c|}{\textbf{Test}} &
  \multicolumn{2}{c|}{\textbf{Age}} &
  \multicolumn{2}{c|}{\textbf{Gender}} &
  \multicolumn{2}{c|}{\textbf{Race}} &
  \multicolumn{1}{c|}{\textbf{Test}} &
  \multicolumn{2}{c|}{\textbf{Age}} &
  \multicolumn{2}{c|}{\textbf{Gender}} &
  \multicolumn{2}{c|}{\textbf{Race}} \\ \cline{2-15} 
 &
  \multicolumn{1}{c|}{\textbf{F1}} &
  \multicolumn{1}{c|}{\textbf{DPD}} &
  \multicolumn{1}{c|}{\textbf{F1}} &
  \multicolumn{1}{c|}{\textbf{DPD}} &
  \multicolumn{1}{c|}{\textbf{F1}} &
  \multicolumn{1}{c|}{\textbf{DPD}} &
  \textbf{F1} &
  \multicolumn{1}{c|}{\textbf{F1}} &
  \multicolumn{1}{c|}{\textbf{DPD}} &
  \multicolumn{1}{c|}{\textbf{F1}} &
  \multicolumn{1}{c|}{\textbf{DPD}} &
  \multicolumn{1}{c|}{\textbf{F1}} &
  \multicolumn{1}{c|}{\textbf{DPD}} &
  \textbf{F1} \\ \hline
\textbf{ResNet18} &
  \multicolumn{1}{c|}{42.7} &
  \multicolumn{1}{c|}{39.0} &
  \multicolumn{1}{c|}{36.2} &
  \multicolumn{1}{c|}{8.5} &
  \multicolumn{1}{c|}{40.8} &
  \multicolumn{1}{c|}{39.9} &
  35.3 &
  \multicolumn{1}{c|}{57.3} &
  \multicolumn{1}{c|}{15.3} &
  \multicolumn{1}{c|}{54.7} &
  \multicolumn{1}{c|}{5.4} &
  \multicolumn{1}{c|}{57.6} &
  \multicolumn{1}{c|}{4.2} &
  55.2 \\ \hline
\textbf{ResNet50} &
  \multicolumn{1}{c|}{39.2} &
  \multicolumn{1}{c|}{38.5} &
  \multicolumn{1}{c|}{36.3} &
  \multicolumn{1}{c|}{3.7} &
  \multicolumn{1}{c|}{40.4} &
  \multicolumn{1}{c|}{21.9} &
  36.5 &
  \multicolumn{1}{c|}{56.0} &
  \multicolumn{1}{c|}{13.3} &
  \multicolumn{1}{c|}{52.8} &
  \multicolumn{1}{c|}{5.0} &
  \multicolumn{1}{c|}{55.9} &
  \multicolumn{1}{c|}{2.9} &
  55.2 \\ \hline
\textbf{ResNext50} &
  \multicolumn{1}{c|}{38.8} &
  \multicolumn{1}{c|}{39.2} &
  \multicolumn{1}{c|}{34.9} &
  \multicolumn{1}{c|}{5.9} &
  \multicolumn{1}{c|}{38.1} &
  \multicolumn{1}{c|}{32.7} &
  35.5 &
  \multicolumn{1}{c|}{55.2} &
  \multicolumn{1}{c|}{13.7} &
  \multicolumn{1}{c|}{52.4} &
  \multicolumn{1}{c|}{4.8} &
  \multicolumn{1}{c|}{54.9} &
  \multicolumn{1}{c|}{3.6} &
  52.1 \\ \hline
\textbf{DenseNet121} &
  \multicolumn{1}{c|}{43.1} &
  \multicolumn{1}{c|}{31.3} &
  \multicolumn{1}{c|}{36.7} &
  \multicolumn{1}{c|}{4.1} &
  \multicolumn{1}{c|}{40.4} &
  \multicolumn{1}{c|}{22.7} &
  36.6 &
  \multicolumn{1}{c|}{58.4} &
  \multicolumn{1}{c|}{13.3} &
  \multicolumn{1}{c|}{55.7} &
  \multicolumn{1}{c|}{4.4} &
  \multicolumn{1}{c|}{58.3} &
  \multicolumn{1}{c|}{3.1} &
  55.7 \\ \hline
\textbf{ViT\_B\_16} &
  \multicolumn{1}{c|}{40.3} &
  \multicolumn{1}{c|}{33.9} &
  \multicolumn{1}{c|}{37.1} &
  \multicolumn{1}{c|}{6.7} &
  \multicolumn{1}{c|}{40.0} &
  \multicolumn{1}{c|}{26.0} &
  37.2 &
  \multicolumn{1}{c|}{50.8} &
  \multicolumn{1}{c|}{14.2} &
  \multicolumn{1}{c|}{47.6} &
  \multicolumn{1}{c|}{4.7} &
  \multicolumn{1}{c|}{50.8} &
  \multicolumn{1}{c|}{3.7} &
  48.8 \\ \hline
\textbf{VGG16} &
  \multicolumn{1}{c|}{43.2} &
  \multicolumn{1}{c|}{30.5} &
  \multicolumn{1}{c|}{43.6} &
  \multicolumn{1}{c|}{9.4} &
  \multicolumn{1}{c|}{42.4} &
  \multicolumn{1}{c|}{27.1} &
  38.3 &
  \multicolumn{1}{c|}{59.5} &
  \multicolumn{1}{c|}{15.2} &
  \multicolumn{1}{c|}{57.5} &
  \multicolumn{1}{c|}{5.0} &
  \multicolumn{1}{c|}{58.9} &
  \multicolumn{1}{c|}{3.4} &
  55.9 \\ \hline
\textbf{EffNet\_B0} &
  \multicolumn{1}{c|}{40.5} &
  \multicolumn{1}{c|}{38.0} &
  \multicolumn{1}{c|}{39.0} &
  \multicolumn{1}{c|}{8.4} &
  \multicolumn{1}{c|}{39.8} &
  \multicolumn{1}{c|}{33.9} &
  37.0 &
  \multicolumn{1}{c|}{56.9} &
  \multicolumn{1}{c|}{13.7} &
  \multicolumn{1}{c|}{54.5} &
  \multicolumn{1}{c|}{4.9} &
  \multicolumn{1}{c|}{57.2} &
  \multicolumn{1}{c|}{3.0} &
  56.5 \\ \hline
\textbf{EffNet\_B7} &
  \multicolumn{1}{c|}{41.4} &
  \multicolumn{1}{c|}{31.1} &
  \multicolumn{1}{c|}{35.7} &
  \multicolumn{1}{c|}{5.4} &
  \multicolumn{1}{c|}{36.9} &
  \multicolumn{1}{c|}{26.3} &
  32.5 &
  \multicolumn{1}{c|}{56.1} &
  \multicolumn{1}{c|}{12.3} &
  \multicolumn{1}{c|}{53.3} &
  \multicolumn{1}{c|}{5.1} &
  \multicolumn{1}{c|}{56.3} &
  \multicolumn{1}{c|}{3.7} &
  54.1 \\ \hline
\textbf{Swin\_B} &
  \multicolumn{1}{c|}{44.6} &
  \multicolumn{1}{c|}{36.6} &
  \multicolumn{1}{c|}{41.5} &
  \multicolumn{1}{c|}{8.9} &
  \multicolumn{1}{c|}{42.8} &
  \multicolumn{1}{c|}{28.8} &
  36.9 &
  \multicolumn{1}{c|}{56.8} &
  \multicolumn{1}{c|}{14.7} &
  \multicolumn{1}{c|}{53.3} &
  \multicolumn{1}{c|}{5.2} &
  \multicolumn{1}{c|}{56.2} &
  \multicolumn{1}{c|}{3.4} &
  53.5 \\ \hline
\textbf{Swin\_V2\_B} &
  \multicolumn{1}{c|}{43.1} &
  \multicolumn{1}{c|}{35.3} &
  \multicolumn{1}{c|}{40.7} &
  \multicolumn{1}{c|}{9.1} &
  \multicolumn{1}{c|}{43.9} &
  \multicolumn{1}{c|}{31.4} &
  38.8 &
  \multicolumn{1}{c|}{59.1} &
  \multicolumn{1}{c|}{14.2} &
  \multicolumn{1}{c|}{56.5} &
  \multicolumn{1}{c|}{5.5} &
  \multicolumn{1}{c|}{58.7} &
  \multicolumn{1}{c|}{3.5} &
  55.0 \\ \hline
\textbf{ConvNeXt\_B} &
  \multicolumn{1}{c|}{47.8} &
  \multicolumn{1}{c|}{24.3} &
  \multicolumn{1}{c|}{43.1} &
  \multicolumn{1}{c|}{6.0} &
  \multicolumn{1}{c|}{45.6} &
  \multicolumn{1}{c|}{19.1} &
  39.7 &
  \multicolumn{1}{c|}{59.4} &
  \multicolumn{1}{c|}{16.5} &
  \multicolumn{1}{c|}{54.8} &
  \multicolumn{1}{c|}{4.6} &
  \multicolumn{1}{c|}{59.0} &
  \multicolumn{1}{c|}{2.8} &
  57.1 \\ \hline
\textbf{iResNet101} &
  \multicolumn{1}{c|}{43.2} &
  \multicolumn{1}{c|}{28.8} &
  \multicolumn{1}{c|}{42.1} &
  \multicolumn{1}{c|}{6.5} &
  \multicolumn{1}{c|}{42.8} &
  \multicolumn{1}{c|}{27.5} &
  36.6 &
  \multicolumn{1}{c|}{60.1} &
  \multicolumn{1}{c|}{17.1} &
  \multicolumn{1}{c|}{56.7} &
  \multicolumn{1}{c|}{5.3} &
  \multicolumn{1}{c|}{59.9} &
  \multicolumn{1}{c|}{3.5} &
  57.0 \\ \hline
\textbf{FUXI} \cite{fuxi}&
  \multicolumn{1}{c|}{38.3} &
  \multicolumn{1}{c|}{38.0} &
  \multicolumn{1}{c|}{39.5} &
  \multicolumn{1}{c|}{8.1} &
  \multicolumn{1}{c|}{39.5} &
  \multicolumn{1}{c|}{36.6} &
  36.0 &
  \multicolumn{1}{c|}{58.8} &
  \multicolumn{1}{c|}{17.0} &
  \multicolumn{1}{c|}{54.8} &
  \multicolumn{1}{c|}{5.3} &
  \multicolumn{1}{c|}{58.7} &
  \multicolumn{1}{c|}{3.9} &
  56.8 \\ \hline
\textbf{SITU} \cite{situ}&
  \multicolumn{1}{c|}{40.6} &
  \multicolumn{1}{c|}{37.9} &
  \multicolumn{1}{c|}{39.3} &
  \multicolumn{1}{c|}{9.8} &
  \multicolumn{1}{c|}{41.2} &
  \multicolumn{1}{c|}{35.5} &
  38.4 &
  \multicolumn{1}{c|}{57.8} &
  \multicolumn{1}{c|}{14.1} &
  \multicolumn{1}{c|}{54} &
  \multicolumn{1}{c|}{4.5} &
  \multicolumn{1}{c|}{57.7} &
  \multicolumn{1}{c|}{3.3} &
  57.4 \\ \hline
\textbf{ME-GraphAU} \cite{megraphau}&
  \multicolumn{1}{c|}{46.1} &
  \multicolumn{1}{c|}{23.9} &
  \multicolumn{1}{c|}{46.0} &
  \multicolumn{1}{c|}{6.7} &
  \multicolumn{1}{c|}{47.1} &
  \multicolumn{1}{c|}{19.9} &
  42.9 &
  \multicolumn{1}{c|}{61.5} &
  \multicolumn{1}{c|}{13.1} &
  \multicolumn{1}{c|}{57.6} &
  \multicolumn{1}{c|}{5.0} &
  \multicolumn{1}{c|}{61.7} &
  \multicolumn{1}{c|}{3.3} &
  59.6 \\ \hline
\textbf{AUNets} \cite{aunets} &
  \multicolumn{1}{c|}{53.3} &
  \multicolumn{1}{c|}{27.2} &
  \multicolumn{1}{c|}{48.5} &
  \multicolumn{1}{c|}{5.1} &
  \multicolumn{1}{c|}{52.2} &
  \multicolumn{1}{c|}{24.0} &
  46.0 &
  \multicolumn{1}{c|}{62.8} &
  \multicolumn{1}{c|}{14.3} &
  \multicolumn{1}{c|}{66.5} &
  \multicolumn{1}{c|}{5.1} &
  \multicolumn{1}{c|}{67.3} &
  \multicolumn{1}{c|}{3.3} &
  64.9 \\ \hline \hline
\multirow{3}{*}{\textbf{Model}} &
  \multicolumn{7}{c|}{\textbf{GFT}} &
  \multicolumn{7}{c|}{\textbf{RAF-AU}} \\ \cline{2-15} 
 &
  \multicolumn{1}{c|}{\textbf{Test}} &
  \multicolumn{2}{c|}{\textbf{Age}} &
  \multicolumn{2}{c|}{\textbf{Gender}} &
  \multicolumn{2}{c|}{\textbf{Race}} &
  \multicolumn{1}{c|}{\textbf{Test}} &
  \multicolumn{2}{c|}{\textbf{Age}} &
  \multicolumn{2}{c|}{\textbf{Gender}} &
  \multicolumn{2}{c|}{\textbf{Race}} \\ \cline{2-15} 
 &
  \multicolumn{1}{c|}{\textbf{F1}} &
  \multicolumn{1}{c|}{\textbf{DPD}} &
  \multicolumn{1}{c|}{\textbf{F1}} &
  \multicolumn{1}{c|}{\textbf{DPD}} &
  \multicolumn{1}{c|}{\textbf{F1}} &
  \multicolumn{1}{c|}{\textbf{DPD}} &
  \textbf{F1} &
  \multicolumn{1}{c|}{\textbf{F1}} &
  \multicolumn{1}{c|}{\textbf{DPD}} &
  \multicolumn{1}{c|}{\textbf{F1}} &
  \multicolumn{1}{c|}{\textbf{DPD}} &
  \multicolumn{1}{c|}{\textbf{F1}} &
  \multicolumn{1}{c|}{\textbf{DPD}} &
  \textbf{F1} \\ \hline
\textbf{ResNet18} &
  \multicolumn{1}{c|}{41.0} &
  \multicolumn{1}{c|}{43.9} &
  \multicolumn{1}{c|}{39.3} &
  \multicolumn{1}{c|}{13.4} &
  \multicolumn{1}{c|}{41.1} &
  \multicolumn{1}{c|}{31.7} &
  39.1 &
  \multicolumn{1}{c|}{68.2} &
  \multicolumn{1}{c|}{24.6} &
  \multicolumn{1}{c|}{62.8} &
  \multicolumn{1}{c|}{2.2} &
  \multicolumn{1}{c|}{67.8} &
  \multicolumn{1}{c|}{13.8} &
  66.4 \\ \hline
\textbf{ResNet50} &
  \multicolumn{1}{c|}{41.5} &
  \multicolumn{1}{c|}{44.8} &
  \multicolumn{1}{c|}{39.0} &
  \multicolumn{1}{c|}{12.9} &
  \multicolumn{1}{c|}{41.3} &
  \multicolumn{1}{c|}{30.0} &
  39.4 &
  \multicolumn{1}{c|}{66.8} &
  \multicolumn{1}{c|}{25.4} &
  \multicolumn{1}{c|}{62.5} &
  \multicolumn{1}{c|}{2.2} &
  \multicolumn{1}{c|}{66.6} &
  \multicolumn{1}{c|}{11.3} &
  65.0 \\ \hline
\textbf{ResNext50} &
  \multicolumn{1}{c|}{39.8} &
  \multicolumn{1}{c|}{39.2} &
  \multicolumn{1}{c|}{36.5} &
  \multicolumn{1}{c|}{9.7} &
  \multicolumn{1}{c|}{39.6} &
  \multicolumn{1}{c|}{28.4} &
  36.9 &
  \multicolumn{1}{c|}{63.7} &
  \multicolumn{1}{c|}{24.7} &
  \multicolumn{1}{c|}{56.7} &
  \multicolumn{1}{c|}{2.6} &
  \multicolumn{1}{c|}{63.8} &
  \multicolumn{1}{c|}{13.0} &
  61.3 \\ \hline
\textbf{DenseNet121} &
  \multicolumn{1}{c|}{41.1} &
  \multicolumn{1}{c|}{46.6} &
  \multicolumn{1}{c|}{38.5} &
  \multicolumn{1}{c|}{12.6} &
  \multicolumn{1}{c|}{41.5} &
  \multicolumn{1}{c|}{32.5} &
  38.5 &
  \multicolumn{1}{c|}{67.9} &
  \multicolumn{1}{c|}{27.7} &
  \multicolumn{1}{c|}{62.2} &
  \multicolumn{1}{c|}{2.9} &
  \multicolumn{1}{c|}{68.1} &
  \multicolumn{1}{c|}{11.7} &
  66.8 \\ \hline
\textbf{ViT\_B\_16} &
  \multicolumn{1}{c|}{43.0} &
  \multicolumn{1}{c|}{43.1} &
  \multicolumn{1}{c|}{37.9} &
  \multicolumn{1}{c|}{12.5} &
  \multicolumn{1}{c|}{43.7} &
  \multicolumn{1}{c|}{32.8} &
  38.9 &
  \multicolumn{1}{c|}{61.1} &
  \multicolumn{1}{c|}{22.8} &
  \multicolumn{1}{c|}{55.2} &
  \multicolumn{1}{c|}{2.5} &
  \multicolumn{1}{c|}{61.5} &
  \multicolumn{1}{c|}{14.5} &
  58.7 \\ \hline
\textbf{VGG16} &
  \multicolumn{1}{c|}{38.1} &
  \multicolumn{1}{c|}{34.0} &
  \multicolumn{1}{c|}{33.9} &
  \multicolumn{1}{c|}{12.0} &
  \multicolumn{1}{c|}{37.2} &
  \multicolumn{1}{c|}{20.9} &
  34.4 &
  \multicolumn{1}{c|}{55.8} &
  \multicolumn{1}{c|}{19.2} &
  \multicolumn{1}{c|}{52.4} &
  \multicolumn{1}{c|}{1.6} &
  \multicolumn{1}{c|}{55.9} &
  \multicolumn{1}{c|}{8.7} &
  54.4 \\ \hline
\textbf{EffNet\_B0} &
  \multicolumn{1}{c|}{42.7} &
  \multicolumn{1}{c|}{40.4} &
  \multicolumn{1}{c|}{38.8} &
  \multicolumn{1}{c|}{9.8} &
  \multicolumn{1}{c|}{42.3} &
  \multicolumn{1}{c|}{30.2} &
  37.7 &
  \multicolumn{1}{c|}{63.6} &
  \multicolumn{1}{c|}{27.6} &
  \multicolumn{1}{c|}{58.9} &
  \multicolumn{1}{c|}{1.9} &
  \multicolumn{1}{c|}{63.6} &
  \multicolumn{1}{c|}{10.8} &
  61.7 \\ \hline
\textbf{EffNet\_B7} &
  \multicolumn{1}{c|}{44.8} &
  \multicolumn{1}{c|}{44.0} &
  \multicolumn{1}{c|}{41.5} &
  \multicolumn{1}{c|}{13.8} &
  \multicolumn{1}{c|}{43.7} &
  \multicolumn{1}{c|}{35.0} &
  41.3 &
  \multicolumn{1}{c|}{68.4} &
  \multicolumn{1}{c|}{26.6} &
  \multicolumn{1}{c|}{63.3} &
  \multicolumn{1}{c|}{2.5} &
  \multicolumn{1}{c|}{68.2} &
  \multicolumn{1}{c|}{12.4} &
  67.3 \\ \hline
\textbf{Swin\_B} &
  \multicolumn{1}{c|}{42.7} &
  \multicolumn{1}{c|}{42.1} &
  \multicolumn{1}{c|}{40.8} &
  \multicolumn{1}{c|}{14.7} &
  \multicolumn{1}{c|}{43.4} &
  \multicolumn{1}{c|}{26.5} &
  40.1 &
  \multicolumn{1}{c|}{70.8} &
  \multicolumn{1}{c|}{25.4} &
  \multicolumn{1}{c|}{66.7} &
  \multicolumn{1}{c|}{3.2} &
  \multicolumn{1}{c|}{70.8} &
  \multicolumn{1}{c|}{11.3} &
  69.8 \\ \hline
\textbf{Swin\_V2\_B} &
  \multicolumn{1}{c|}{44.5} &
  \multicolumn{1}{c|}{43.4} &
  \multicolumn{1}{c|}{41.1} &
  \multicolumn{1}{c|}{14.7} &
  \multicolumn{1}{c|}{44.1} &
  \multicolumn{1}{c|}{31.3} &
  39.9 &
  \multicolumn{1}{c|}{70.3} &
  \multicolumn{1}{c|}{25.6} &
  \multicolumn{1}{c|}{64.3} &
  \multicolumn{1}{c|}{2.7} &
  \multicolumn{1}{c|}{70.1} &
  \multicolumn{1}{c|}{12.6} &
  69.5 \\ \hline
\textbf{ConvNeXt\_B} &
  \multicolumn{1}{c|}{43.4} &
  \multicolumn{1}{c|}{45.8} &
  \multicolumn{1}{c|}{40.4} &
  \multicolumn{1}{c|}{15.1} &
  \multicolumn{1}{c|}{42.6} &
  \multicolumn{1}{c|}{31.5} &
  40.2 &
  \multicolumn{1}{c|}{73.0} &
  \multicolumn{1}{c|}{26.7} &
  \multicolumn{1}{c|}{68.0} &
  \multicolumn{1}{c|}{2.5} &
  \multicolumn{1}{c|}{72.7} &
  \multicolumn{1}{c|}{12.8} &
  71.0 \\ \hline
\textbf{iResNet101} &
  \multicolumn{1}{c|}{44.9} &
  \multicolumn{1}{c|}{38.6} &
  \multicolumn{1}{c|}{41.9} &
  \multicolumn{1}{c|}{11.0} &
  \multicolumn{1}{c|}{44.3} &
  \multicolumn{1}{c|}{29.2} &
  40.1 &
  \multicolumn{1}{c|}{71.9} &
  \multicolumn{1}{c|}{25.1} &
  \multicolumn{1}{c|}{66.3} &
  \multicolumn{1}{c|}{2.2} &
  \multicolumn{1}{c|}{71.9} &
  \multicolumn{1}{c|}{12.5} &
  70.3 \\ \hline
\textbf{FUXI} \cite{fuxi}&
  \multicolumn{1}{c|}{42.8} &
  \multicolumn{1}{c|}{35.9} &
  \multicolumn{1}{c|}{39.3} &
  \multicolumn{1}{c|}{10.5} &
  \multicolumn{1}{c|}{41.4} &
  \multicolumn{1}{c|}{26.3} &
  37.9 &
  \multicolumn{1}{c|}{68.0} &
  \multicolumn{1}{c|}{27.6} &
  \multicolumn{1}{c|}{62.7} &
  \multicolumn{1}{c|}{3.1} &
  \multicolumn{1}{c|}{68.3} &
  \multicolumn{1}{c|}{14.4} &
  67.2 \\ \hline
\textbf{SITU} \cite{situ}&
  \multicolumn{1}{c|}{45.1} &
  \multicolumn{1}{c|}{48.1} &
  \multicolumn{1}{c|}{39.0} &
  \multicolumn{1}{c|}{12.4} &
  \multicolumn{1}{c|}{43.5} &
  \multicolumn{1}{c|}{28.0} &
  39.3 &
  \multicolumn{1}{c|}{72.3} &
  \multicolumn{1}{c|}{28.1} &
  \multicolumn{1}{c|}{67.5} &
  \multicolumn{1}{c|}{2.5} &
  \multicolumn{1}{c|}{72.4} &
  \multicolumn{1}{c|}{12.9} &
  71.8 \\ \hline
\textbf{ME-GraphAU} \cite{megraphau}&
  \multicolumn{1}{c|}{48.6} &
  \multicolumn{1}{c|}{42.8} &
  \multicolumn{1}{c|}{43.5} &
  \multicolumn{1}{c|}{14.3} &
  \multicolumn{1}{c|}{48.1} &
  \multicolumn{1}{c|}{26.5} &
  42.7 &
  \multicolumn{1}{c|}{73.5} &
  \multicolumn{1}{c|}{25.4} &
  \multicolumn{1}{c|}{67.1} &
  \multicolumn{1}{c|}{2.9} &
  \multicolumn{1}{c|}{73.2} &
  \multicolumn{1}{c|}{12.5} &
  71.9 \\ \hline
\textbf{AUNets} \cite{aunets} &
  \multicolumn{1}{c|}{50.5} &
  \multicolumn{1}{c|}{41.0} &
  \multicolumn{1}{c|}{42.9} &
  \multicolumn{1}{c|}{14.0} &
  \multicolumn{1}{c|}{49.7} &
  \multicolumn{1}{c|}{25.5} &
  43.7 &
  \multicolumn{1}{c|}{72.4} &
  \multicolumn{1}{c|}{26.3} &
  \multicolumn{1}{c|}{65.8} &
  \multicolumn{1}{c|}{2.1} &
  \multicolumn{1}{c|}{70.9} &
  \multicolumn{1}{c|}{13.5} &
  67.7 \\ \hline
\end{tabular}}
\end{table*}

\begin{table}[]
\centering
\caption{CCC Performance comparison (in \%) between various baseline and state-of-the-art works for VA estimation, A higher value is desired}\label{table6}
\setlength{\tabcolsep}{2.4mm}
\begin{tabular}{|c|c|c|c|c|}
\hline
\textbf{Model}       & \textbf{Test} & \textbf{Age} & \textbf{Gen.} & \textbf{Race} \\ \hline
\textbf{ResNet18}    & 70.0          & 69.0         & 69.3          & 69.3          \\ \hline
\textbf{ResNet50}    & 69.3          & 68.3         & 69.0          & 68.6          \\ \hline
\textbf{resnext50}   & 69.0          & 67.3         & 68.4          & 68.3          \\ \hline
\textbf{DenseNet121} & 69.5          & 68.0         & 68.8          & 68.8          \\ \hline
\textbf{ViT\_B\_16}  & 70.5          & 69.0         & 70.1          & 69.8          \\ \hline
\textbf{VGG16}       & 70.9          & 69.7         & 70.4          & 70.2          \\ \hline
\textbf{EffNet\_B0}  & 71.2          & 69.3         & 70.0            & 69.9          \\ \hline
\textbf{EffNet\_B7}  & 71.3          & 69.9         & 70.5          & 70.2          \\ \hline
\textbf{Swin\_B}     & 71.8          & 70.5         & 70.9          & 70.6          \\ \hline
\textbf{Swin\_V2\_B} & 71.6          & 70.5         & 71.1          & 71.0          \\ \hline
\textbf{ConvNeXt\_B} & 72.5          & 70.7         & 71.0            & 71.0          \\ \hline
\textbf{iResNet101}  & 71.1          & 69.8         & 70.5          & 70.3          \\ \hline
\textbf{FUXI \cite{fuxi}}        & 74.3          & 72.8         & 73.3          & 73.3          \\ \hline
\textbf{SITU \cite{situ}}        & 72.2          & 71.4         & 72.0          & 71.7          \\ \hline
\textbf{CTC \cite{ctc}}         & 72.9          & 71.2         & 72.3          & 71.5          \\ \hline
\end{tabular}
\end{table}

We conducted experiments using datasets partitioned according to our proposed protocol to evaluate the performance and fairness of both models and datasets. These experiments incorporated prevalent baseline models (Base), top-ranking methods from the ABAW series challenges (ABAW), and state-of-the-art (SOTA) models.



\subsection{Expression Recognition}

For the FER task, we conducted experiments using the AffectNet dataset (both for 7 and 8 expression categories) and the RAF-DB dataset. We employed the F1 score on the test set (F1 Test) to assess models' overall performance. In addition, we use carefully selected fairness evaluation metrics SP and average F1 across subgroups (F1) to evaluate the fairness of the models. SP measures equality in the probability of the model predicting positive outcomes across different demographic groups. The F1 aims to complement SP by evaluating the model's prediction accuracy across various demographic subgroups. The results are summarized in Table \ref{table4}.


The F1 Test results demonstrate that SOTA models consistently exhibit superior recognition accuracy compared to baseline models. However, it is noteworthy that the ranking of models varies under our new partitioning protocol. For instance, in the original AffectNet dataset partition, models like MT-EffNet \cite{mteffnet} and DAN \cite{dan} demonstrated a clear advantage; however, their performance diminished on the newly partitioned dataset, placing them at lower positions. Conversely, models such as EAC \cite{eac} and MA-Net\cite{manet}, which previously ranked behind, exhibited improved performance on our revised dataset partition. 

To elucidate the underlying reasons for this shift, we conducted analysis using fairness evaluation metrics. Model DAN and MT-EffNet exhibit significantly higher SP across Age, Gender and Race compared to EAC and MA-Net, indicating unfairness in model predictions across these demographic attributes. Upon closer examination of the data statistic in Table \ref{table2}, we attribute this discrepancy to the small size of the previous test set, which primarily consisted of approximately 4000 images concentrated within the 20-39 age group of the White demographic. In contrast, our new partition includes a more diverse range of samples presenting various races and ages in the test set. Consequently, the limitations of models in generalizing to more complex population distributions became apparent. In such intricate scenarios, the lack of fairness in models compromises their overall recognition performance, which emphasizes the importance of addressing fairness considerations alongside overall performance.


We also observed an intriguing phenomenon with the POSTER++ \cite{poster} model. Despite its subpar performance in fairness metrics, it still achieved relatively high overall recognition rates. This suggests that fairness and overall recognition performance may not exhibit a straightforward positive or negative correlation.

To explore the relationship between the two, we analyzed the performance of all models on our newly partitioned AffectNet-7 and AffectNet-8 datasets. In the AffectNet-7 dataset, the SP values of all models across Gender and Race ranged between [0-5], with slightly higher values for Age but still under 10. This indicates that the newly partitioned AffectNet-7 dataset inherently possesses fair features, as all models trained on this dataset exhibit good fairness. Conversely, AffectNet-8 demonstrates fairness only in Gender's SP, mainly ranging between [5-10], with poorer fairness performance in other attributes. This is primarily attributed to the addition of a small number of Contempt samples to the dataset, increasing the complexity of model classification from 7 to 8 classes, thereby making it more challenging for models to maintain fairness. 

Furthermore, most models exhibiting superior fairness performance also demonstrate overall recognition proficiency (F1 Test) on both datasets. For instance,  EAC  \cite{eac} model achieves optimal performance in fairness metrics on both datasets, whilst  obtaining the highest overall recognition performance. Despite POSTER++ displaying poor fairness metrics on AffectNet-8, it still achieves results comparable to the top-performing EAC model. However, its performance on the inherently fair dataset AffectNet-7 falls short compared to the fair model EAC. This indicates that while POSTER++ may excel in accuracy for specific subgroups, it fails to showcase its superiority on fair datasets. Therefore, we can conclude that while enhancing a model's performance for the majority of groups is undoubtedly effective, enhancing its fairness is even more crucial in intricate scenarios.


To investigate the inherent fairness character of the dataset, we conducted an analysis for model performance and dataset compositions on the newly partitioned RAF-DB and AffectNet-8 in comparison to AffectNet-7, leading to the following understanding: Firstly, as models tackle more complex tasks, achieving fairness becomes increasingly challenging. A comparison between AffectNet-7 and AffectNet-8 reveals that while both datasets share identical data samples for the seven base expressions, the addition of an extra class in AffectNet-8 makes it more difficult for models to achieve fairness. Secondly, smaller dataset scales and significant disparities in demographic attribute distributions can detrimentally impact model fairness. Despite both RAF-DB and AffectNet-7 handling seven-class classification tasks, differences in dataset scale and demographic attribute distributions, as shown in Table \ref{table2}, result in varied fairness performance. For instance, due to limitations in the original dataset's size, the newly partitioned RAF-DB training set contains only 366 samples for the '70+' age group and 865 samples for the '0-3' age group, as illustrated in the fairness evaluation metrics in Table \ref{table4}, leading to noticeable unfairness in model predictions for the Age attribute.

\subsection{AU Detection}

For the AU detection task, we conducted experiments on our newly partitioned datasets: DISFA, EmotioNet, GFT, and RAF-AU. To assess the fairness of models, we carefully selected two fairness metrics: DPD to measure the disparity in positive class proportions across different attribute groups; and average F1 across subgroups (F1) to assess whether the model's performance is consistent across different demographic attributes. The experiment results are presented in Table \ref{table5}.


Analyzing the experimental results on the newly partitioned datasets revealed that all models demonstrated fairness for the Gender attribute across the EmotioNet, DISFA, GFT, and RAF-AU datasets, as evidenced by DPD values below 10. These results suggest that Gender is the attribute most conducive to achieving fairness. We attribute this mainly to the relatively balanced distribution of male and female samples within the datasets compared to other demographic attributes. While the GFT dataset displayed slight deviations from fairness in gender. Upon exploration of the dataset distributions (as shown in Table \ref{table3}), it was noted that the Gender ratios in EmotioNet and RAF-AU were close to 1:1, while GFT had ratios close to 1:1.4. For the Age attribute, F1 scores ranked the lowest across all demographic attributes, indicating considerable difficulty in achieving fairness for this attribute.

Additionally, for the EmotioNet dataset, fairness metrics across all demographic attributes outperformed the other three datasets. This can be attributed to the dataset's ``in-the-wild" image nature, which provides richer attribute diversity compared to video datasets like DISFA and GFT. Although RAF-AU shares the "in-the-wild" nature, its limited data size may have hindered model performance. Notably, compared to the FER task, all models demonstrated generally poor fairness across all datasets, particularly with respect to attributes such as Age and Race. This underscores the inherent challenges in addressing fairness in AU tasks. 


For the performance of different models across the four datasets, AUNets \cite{aunets} and ME-GraphAU \cite{megraphau} demonstrated the best overall performance (Test F1) across all datasets. Notably, while ME-GraphAU exhibited slightly weaker performance compared to AUNets on the other datasets, the fairness metric results for ME-GraphAU surpassed those of AUNets across all datasets, establishing it as the best-performing model in terms of fairness among all models. The potential reason for this phenomenon could be that ME-GraphAU may have effectively learned to mitigate biases present in the dataset and make more equitable predictions across demographic subgroups.

This observation challenges the conventional notion that achieving fairness inevitably leads to a decrease in overall performance. On the contrary, it suggests that models can achieve both fairness and better overall performance concurrently. This indicates that there is no inherent trade-off between fairness and performance, debunking a common misconception in fair model design. This insight provides valuable guidance for future fair model development, emphasizing the importance of prioritizing fairness without compromising performance.


\subsection{VA Estimation}

For the VA task, experiments were conducted on the newly partitioned AffectNet-VA datasets. Given that VA involves continuous values, traditional fairness metrics such as SP and DPD are not applicable. Instead, the average CCC between valence and arousal across subgroups within different demographic attributes (Average CCC) was employed as a fairness evaluation. The experimental results, presented in Table \ref{table6}, reveal that methods achieving excellent scores in the 5th ABAW challenge maintain favorable overall performance. Specifically, FUXI \cite{fuxi} and CTC \cite{ctc} achieved the highest performance. Interestingly, these two models also demonstrated the best performance in fairness metrics, indicating that prioritizing fairness in VA predictions can significantly enhance overall model performance. This observation highlights the importance of incorporating fairness considerations into VA prediction models, as it not only promotes equitable outcomes but also contributes to overall performance enhancement.

In terms of fairness performance across all demographic attributes, Age exhibited the lowest fairness among the three demographic attributes, consistent with findings from previous tasks. This underscores the persistent challenge of achieving fairness for Age predictions.  



\section{CONCLUSION}

In conclusion, this research underscores the critical importance of addressing biases and promoting fairness in the deployment of machine learning algorithms, particularly within the realm of automatic affect analysis. The study conducted a thorough analysis of seven affective databases, annotating demographic attributes and proposing a standardized protocol for database partitioning with a focus on fairness in evaluations. Through extensive experiments with baseline and SOTA methods, the findings reveal the significant impact of these changes and highlight the inadequacy of prior assessments. The emphasis on considering demographic attributes and fairness in affect analysis research provides a solid foundation for enhancing the model performance and the development of more equitable methodologies. 

{\small
\bibliographystyle{ieee}
\bibliography{egbib}

\begin{thebibliography}{10}\itemsep=-1pt

\bibitem{an2024learning}
R.~An, A.~Jin, W.~Chen, W.~Zhang, H.~Zeng, Z.~Deng, and Y.~Ding.
\newblock Learning facial expression-aware global-to-local representation for robust action unit detection.
\newblock {\em Applied Intelligence}, pages 1--21, 2024.

\bibitem{emogcn}
P.~Antoniadis, P.~P. Filntisis, and P.~Maragos.
\newblock Exploiting emotional dependencies with graph convolutional networks for facial expression recognition.
\newblock In {\em 2021 16th IEEE International Conference on Automatic Face and Gesture Recognition (FG 2021)}, pages 1--8, 2021.

\bibitem{emotionet2016}
C.~Benitez-Quiroz, R.~Srinivasan, and A.~Martinez.
\newblock Emotionet: An accurate, real-time algorithm for the automatic annotation of a million facial expressions in the wild.
\newblock In {\em Proceedings of IEEE International Conference on Computer Vision \& Pattern Recognition (CVPR'16)}, Las Vegas, NV, USA, June 2016.

\bibitem{chaudhari2022vitfer}
A.~Chaudhari, C.~Bhatt, A.~Krishna, and P.~L. Mazzeo.
\newblock Vitfer: facial emotion recognition with vision transformers.
\newblock {\em Applied System Innovation}, 5(4):80, 2022.

\bibitem{cui2023biomechanics}
Z.~Cui, C.~Kuang, T.~Gao, K.~Talamadupula, and Q.~Ji.
\newblock Biomechanics-guided facial action unit detection through force modeling.
\newblock In {\em Proceedings of the IEEE/CVF Conference on Computer Vision and Pattern Recognition}, pages 8694--8703, 2023.

\bibitem{ekman2002facial}
P.~Ekman.
\newblock Facial action coding system (facs).
\newblock {\em A human face}, 2002.

\bibitem{dacl}
A.~H. Farzaneh and X.~Qi.
\newblock Facial expression recognition in the wild via deep attentive center loss.
\newblock In {\em Proceedings of the IEEE/CVF winter conference on applications of computer vision}, pages 2402--2411, 2021.

\bibitem{girard2017sayette}
J.~M. Girard, W.-S. Chu, L.~A. Jeni, and J.~F. Cohn.
\newblock Sayette group formation task (gft) spontaneous facial expression database.
\newblock In {\em 2017 12th IEEE International Conference on Automatic Face \& Gesture Recognition (FG 2017)}, pages 581--588. IEEE, 2017.

\bibitem{handrich2020simultaneous}
S.~Handrich, L.~Dinges, A.~Al-Hamadi, P.~Werner, and Z.~Al~Aghbari.
\newblock Simultaneous prediction of valence/arousal and emotions on affectnet, aff-wild and afew-va.
\newblock {\em Procedia Computer Science}, 170:634--641, 2020.

\bibitem{fairface}
K.~Karkkainen and J.~Joo.
\newblock Fairface: Face attribute dataset for balanced race, gender, and age for bias measurement and mitigation.
\newblock In {\em Proceedings of the IEEE/CVF winter conference on applications of computer vision}, pages 1548--1558, 2021.

\bibitem{kossaifi2020factorized}
J.~Kossaifi, A.~Toisoul, A.~Bulat, Y.~Panagakis, T.~M. Hospedales, and M.~Pantic.
\newblock Factorized higher-order cnns with an application to spatio-temporal emotion estimation.
\newblock In {\em Proceedings of the IEEE/CVF Conference on Computer Vision and Pattern Recognition}, pages 6060--6069, 2020.

\bibitem{li2017reliable}
S.~Li, W.~Deng, and J.~Du.
\newblock Reliable crowdsourcing and deep locality-preserving learning for expression recognition in the wild.
\newblock In {\em Computer Vision and Pattern Recognition (CVPR), 2017 IEEE Conference on}, pages 2584--2593. IEEE, 2017.

\bibitem{li2019self}
Y.~Li, J.~Zeng, S.~Shan, and X.~Chen.
\newblock Self-supervised representation learning from videos for facial action unit detection.
\newblock In {\em Proceedings of the IEEE/CVF Conference on Computer vision and pattern recognition}, pages 10924--10933, 2019.

\bibitem{situ}
C.~Liu, X.~Zhang, X.~Liu, T.~Zhang, L.~Meng, Y.~Liu, Y.~Deng, and W.~Jiang.
\newblock Facial expression recognition based on multi-modal features for videos in the wild.
\newblock In {\em Proceedings of the IEEE/CVF Conference on Computer Vision and Pattern Recognition}, pages 5871--5878, 2023.

\bibitem{liu2023pose}
Y.~Liu, W.~Wang, Y.~Zhan, S.~Feng, K.~Liu, and Z.~Chen.
\newblock Pose-disentangled contrastive learning for self-supervised facial representation.
\newblock In {\em Proceedings of the IEEE/CVF Conference on Computer Vision and Pattern Recognition}, pages 9717--9728, 2023.

\bibitem{megraphau}
C.~Luo, S.~Song, W.~Xie, L.~Shen, and H.~Gunes.
\newblock Learning multi-dimensional edge feature-based au relation graph for facial action unit recognition.
\newblock In {\em Proceedings of the Thirty-First International Joint Conference on Artificial Intelligence}, pages 1239--1246, 2022.

\bibitem{poster}
J.~Mao, R.~Xu, X.~Yin, Y.~Chang, B.~Nie, and A.~Huang.
\newblock Poster v2: A simpler and stronger facial expression recognition network.
\newblock {\em arXiv preprint arXiv:2301.12149}, 2023.

\bibitem{mavadati2013disfa}
S.~M. Mavadati, M.~H. Mahoor, K.~Bartlett, P.~Trinh, and J.~F. Cohn.
\newblock Disfa: A spontaneous facial action intensity database.
\newblock {\em Affective Computing, IEEE Transactions on}, 4(2):151--160, 2013.

\bibitem{mitenkova2019valence}
A.~Mitenkova, J.~Kossaifi, Y.~Panagakis, and M.~Pantic.
\newblock Valence and arousal estimation in-the-wild with tensor methods.
\newblock In {\em 2019 14th IEEE International Conference on Automatic Face \& Gesture Recognition (FG 2019)}, pages 1--7. IEEE, 2019.

\bibitem{mollahosseini2017affectnet}
A.~Mollahosseini, B.~Hasani, and M.~H. Mahoor.
\newblock Affectnet: A database for facial expression, valence, and arousal computing in the wild.
\newblock {\em arXiv preprint arXiv:1708.03985}, 2017.

\bibitem{parameshwara2023efficient}
R.~Parameshwara, I.~Radwan, A.~Asthana, I.~Abbasnejad, R.~Subramanian, and R.~Goecke.
\newblock Efficient labelling of affective video datasets via few-shot \& multi-task contrastive learning.
\newblock In {\em Proceedings of the 31st ACM International Conference on Multimedia}, pages 6161--6170, 2023.

\bibitem{parameshwara2023examining}
R.~Parameshwara, I.~Radwan, R.~Subramanian, and R.~Goecke.
\newblock Examining subject-dependent and subject-independent human affect inference from limited video data.
\newblock In {\em 2023 IEEE 17th International Conference on Automatic Face and Gesture Recognition (FG)}, pages 1--6. IEEE, 2023.

\bibitem{aunets}
A.~Romero, J.~Le{\'o}n, and P.~Arbel{\'a}ez.
\newblock Multi-view dynamic facial action unit detection.
\newblock {\em Image and Vision Computing}, 2018.

\bibitem{russell1978evidence}
J.~A. Russell.
\newblock Evidence of convergent validity on the dimensions of affect.
\newblock {\em Journal of personality and social psychology}, 36(10):1152, 1978.

\bibitem{mteffnet}
A.~V. Savchenko, L.~V. Savchenko, and I.~Makarov.
\newblock Classifying emotions and engagement in online learning based on a single facial expression recognition neural network.
\newblock {\em IEEE Transactions on Affective Computing}, 13(4):2132--2143, 2022.

\bibitem{schaefer2008encyclopedia}
R.~T. Schaefer.
\newblock {\em Encyclopedia of race, ethnicity, and society}, volume~1.
\newblock Sage, 2008.

\bibitem{shao2024joint}
Z.~Shao, Y.~Zhou, F.~Li, H.~Zhu, and B.~Liu.
\newblock Joint facial action unit recognition and self-supervised optical flow estimation.
\newblock {\em Pattern Recognition Letters}, 2024.

\bibitem{varsha}
V.~Suresh, G.~Yeo, and D.~C~Ong.
\newblock Critically examining the domain generalizability of facial expression recognition models.
\newblock {\em arXiv preprint arXiv:2106.15453}, 2023.

\bibitem{wang2020tal}
P.~Wang, Z.~Wang, Z.~Ji, X.~Liu, S.~Yang, and Z.~Wu.
\newblock Tal emotionet challenge 2020 rethinking the model chosen problem in multi-task learning.
\newblock In {\em Proceedings of the IEEE/CVF Conference on Computer Vision and Pattern Recognition Workshops}, pages 412--413, 2020.

\bibitem{dan}
Z.~Wen, W.~Lin, T.~Wang, and G.~Xu.
\newblock Distract your attention: Multi-head cross attention network for facial expression recognition.
\newblock {\em Biomimetics}, 8(2):199, 2023.

\bibitem{werner2020facial}
P.~Werner, F.~Saxen, and A.~Al-Hamadi.
\newblock Facial action unit recognition in the wild with multi-task cnn self-training for the emotionet challenge.
\newblock In {\em Proceedings of the IEEE/CVF conference on computer vision and pattern recognition workshops}, pages 410--411, 2020.

\bibitem{yan2020raf}
W.-J. Yan, S.~Li, C.~Que, J.~Pei, and W.~Deng.
\newblock Raf-au database: in-the-wild facial expressions with subjective emotion judgement and objective au annotations.
\newblock In {\em Proceedings of the Asian conference on computer vision}, 2020.

\bibitem{ctc}
J.~Yu, R.~Li, Z.~Cai, G.~Zhao, G.~Xie, J.~Zhu, W.~Zhu, Q.~Ling, L.~Wang, C.~Wang, et~al.
\newblock Local region perception and relationship learning combined with feature fusion for facial action unit detection.
\newblock In {\em Proceedings of the IEEE/CVF Conference on Computer Vision and Pattern Recognition}, pages 5784--5791, 2023.

\bibitem{fuxi}
W.~Zhang, B.~Ma, F.~Qiu, and Y.~Ding.
\newblock Multi-modal facial affective analysis based on masked autoencoder.
\newblock In {\em Proceedings of the IEEE/CVF Conference on Computer Vision and Pattern Recognition}, pages 5792--5801, 2023.

\bibitem{zhang2024multimodal}
X.~Zhang, H.~Yang, T.~Wang, X.~Li, and L.~Yin.
\newblock Multimodal channel-mixing: Channel and spatial masked autoencoder on facial action unit detection.
\newblock In {\em Proceedings of the IEEE/CVF Winter Conference on Applications of Computer Vision}, pages 6077--6086, 2024.

\bibitem{eac}
Y.~Zhang, C.~Wang, X.~Ling, and W.~Deng.
\newblock Learn from all: Erasing attention consistency for noisy label facial expression recognition.
\newblock In {\em European Conference on Computer Vision}, pages 418--434. Springer, 2022.

\bibitem{manet}
Z.~Zhao, Q.~Liu, and S.~Wang.
\newblock Learning deep global multi-scale and local attention features for facial expression recognition in the wild.
\newblock {\em IEEE Transactions on Image Processing}, 30:6544--6556, 2021.

\bibitem{zheng2023poster}
C.~Zheng, M.~Mendieta, and C.~Chen.
\newblock Poster: A pyramid cross-fusion transformer network for facial expression recognition.
\newblock In {\em Proceedings of the IEEE/CVF International Conference on Computer Vision}, pages 3146--3155, 2023.

\end{thebibliography}
}

\end{document}


\maketitle

\begin{table}[]
\centering\scriptsize
\caption{Data Statistics for VA Databases  for the New and Original Partitions}\label{table3}
\renewcommand{\arraystretch}{1.1} 
\scalebox{1}{}
\begin{tabular}{|cc|cccc|}
\hline
\multicolumn{2}{|c|}{\textbf{Name}} &
  \multicolumn{4}{c|}{\textbf{AFEW}} \\ \hline
\multicolumn{2}{|c|}{\textbf{Version}} &
  \multicolumn{1}{c|}{\textbf{Original}} &
  \multicolumn{3}{c|}{\textbf{New}} \\ \hline
\multicolumn{2}{|c|}{\textbf{Sets}} &
  \multicolumn{1}{c|}{\textbf{Train}} &
  \multicolumn{1}{c|}{\textbf{Train}} &
  \multicolumn{1}{c|}{\textbf{Valid}} &
  \textbf{Test} \\ \hline
\multicolumn{2}{|c|}{\textbf{Total Amount}} &
  \multicolumn{1}{c|}{30051} &
  \multicolumn{1}{c|}{16566} &
  \multicolumn{1}{c|}{4498} &
  8964 \\ \hline
\multicolumn{1}{|c|}{\multirow{2}{*}{\textbf{Gender}}} &
  \textbf{Female} &
  \multicolumn{1}{c|}{15499} &
  \multicolumn{1}{c|}{8557} &
  \multicolumn{1}{c|}{2339} &
  4603 \\ \cline{2-6} 
\multicolumn{1}{|c|}{} &
  \textbf{Male} &
  \multicolumn{1}{c|}{14529} &
  \multicolumn{1}{c|}{8009} &
  \multicolumn{1}{c|}{2159} &
  4361 \\ \hline
\multicolumn{1}{|c|}{\multirow{4}{*}{\textbf{Race}}} &
  \textbf{Asian} &
  \multicolumn{1}{c|}{2481} &
  \multicolumn{1}{c|}{1526} &
  \multicolumn{1}{c|}{308} &
  647 \\ \cline{2-6} 
\multicolumn{1}{|c|}{} &
  \textbf{Black} &
  \multicolumn{1}{c|}{837} &
  \multicolumn{1}{c|}{367} &
  \multicolumn{1}{c|}{136} &
  334 \\ \cline{2-6} 
\multicolumn{1}{|c|}{} &
  \textbf{Indian} &
  \multicolumn{1}{c|}{206} &
  \multicolumn{1}{c|}{81} &
  \multicolumn{1}{c|}{88} &
  37 \\ \cline{2-6} 
\multicolumn{1}{|c|}{} &
  \textbf{\begin{tabular}[c]{@{}c@{}}White-Cauasian\end{tabular}} &
  \multicolumn{1}{c|}{26504} &
  \multicolumn{1}{c|}{14592} &
  \multicolumn{1}{c|}{3966} &
  7946 \\ \hline
\multicolumn{1}{|c|}{\multirow{9}{*}{\textbf{Age}}} &
  \textbf{0-2} &
  \multicolumn{1}{c|}{/} &
  \multicolumn{1}{c|}{/} &
  \multicolumn{1}{c|}{/} &
  / \\ \cline{2-6} 
\multicolumn{1}{|c|}{} &
  \textbf{03-09} &
  \multicolumn{1}{c|}{366} &
  \multicolumn{1}{c|}{280} &
  \multicolumn{1}{c|}{40} &
  46 \\ \cline{2-6} 
\multicolumn{1}{|c|}{} &
  \textbf{10-19} &
  \multicolumn{1}{c|}{1354} &
  \multicolumn{1}{c|}{755} &
  \multicolumn{1}{c|}{190} &
  409 \\ \cline{2-6} 
\multicolumn{1}{|c|}{} &
  \textbf{20-29} &
  \multicolumn{1}{c|}{15926} &
  \multicolumn{1}{c|}{8762} &
  \multicolumn{1}{c|}{2426} &
  4738 \\ \cline{2-6} 
\multicolumn{1}{|c|}{} &
  \textbf{30-39} &
  \multicolumn{1}{c|}{6567} &
  \multicolumn{1}{c|}{3672} &
  \multicolumn{1}{c|}{1000} &
  1895 \\ \cline{2-6} 
\multicolumn{1}{|c|}{} &
  \textbf{40-49} &
  \multicolumn{1}{c|}{4214} &
  \multicolumn{1}{c|}{2224} &
  \multicolumn{1}{c|}{600} &
  1390 \\ \cline{2-6} 
\multicolumn{1}{|c|}{} &
  \textbf{50-59} &
  \multicolumn{1}{c|}{1304} &
  \multicolumn{1}{c|}{750} &
  \multicolumn{1}{c|}{198} &
  356 \\ \cline{2-6} 
\multicolumn{1}{|c|}{} &
  \textbf{60-69} &
  \multicolumn{1}{c|}{249} &
  \multicolumn{1}{c|}{123} &
  \multicolumn{1}{c|}{44} &
  82 \\ \cline{2-6} 
\multicolumn{1}{|c|}{} &
  \textbf{70+} &
  \multicolumn{1}{c|}{48} &
  \multicolumn{1}{c|}{/} &
  \multicolumn{1}{c|}{/} &
  48 \\ \hline
\end{tabular}
\end{table}

\begin{table*}[]
\scriptsize
\centering
\caption{Performance comparison (in \%) between various baseline and state-of-the-art works for AU detection; for F1 higher values are wanted; for DPD and EOD, a value of 0 is wanted, values in \textcolor{red} {[0,10]}  indicate fair models}\label{table6}
\setlength{\tabcolsep}{0.8mm}
\renewcommand{\arraystretch}{1.05}
\scalebox{1.05}{
\begin{tabular}{|c|cccccccccc|cccccccccc|}
\hline
\multirow{3}{*}{\textbf{Model}} &
  \multicolumn{10}{c|}{\textbf{DISFA}} &
  \multicolumn{10}{c|}{\textbf{EmotioNet}} \\ \cline{2-21} 
 &
  \multicolumn{1}{c|}{\textbf{F1}} &
  \multicolumn{3}{c|}{\textbf{DPD}} &
  \multicolumn{3}{c|}{\textbf{EOD}} &
  \multicolumn{3}{c|}{\textbf{F1}} &
  \multicolumn{1}{c|}{\textbf{F1}} &
  \multicolumn{3}{c|}{\textbf{DPD}} &
  \multicolumn{3}{c|}{\textbf{EOD}} &
  \multicolumn{3}{c|}{\textbf{F1}} \\ \cline{2-21} 
 &
  \multicolumn{1}{c|}{\textbf{Test}} &
  \multicolumn{1}{c|}{\textbf{Age}} &
  \multicolumn{1}{c|}{\textbf{Gender}} &
  \multicolumn{1}{c|}{\textbf{Race}} &
  \multicolumn{1}{c|}{\textbf{Age}} &
  \multicolumn{1}{c|}{\textbf{Gender}} &
  \multicolumn{1}{c|}{\textbf{Race}} &
  \multicolumn{1}{c|}{\textbf{Age}} &
  \multicolumn{1}{c|}{\textbf{Gender}} &
  \textbf{Race} &
  \multicolumn{1}{c|}{\textbf{Test}} &
  \multicolumn{1}{c|}{\textbf{Age}} &
  \multicolumn{1}{c|}{\textbf{Gender}} &
  \multicolumn{1}{c|}{\textbf{Race}} &
  \multicolumn{1}{c|}{\textbf{Age}} &
  \multicolumn{1}{c|}{\textbf{Gender}} &
  \multicolumn{1}{c|}{\textbf{Race}} &
  \multicolumn{1}{c|}{\textbf{Age}} &
  \multicolumn{1}{c|}{\textbf{Gender}} &
  \textbf{Race} \\ \hline
\textbf{ResNet18} &
  \multicolumn{1}{c|}{42.7} &
  \multicolumn{1}{c|}{39.0} &
  \multicolumn{1}{c|}{8.5} &
  \multicolumn{1}{c|}{39.9} &
  \multicolumn{1}{c|}{57.0} &
  \multicolumn{1}{c|}{11.7} &
  \multicolumn{1}{c|}{53.3} &
  \multicolumn{1}{c|}{36.2} &
  \multicolumn{1}{c|}{40.8} &
  35.3 &
  \multicolumn{1}{c|}{57.3} &
  \multicolumn{1}{c|}{15.3} &
  \multicolumn{1}{c|}{5.4} &
  \multicolumn{1}{c|}{4.2} &
  \multicolumn{1}{c|}{44.6} &
  \multicolumn{1}{c|}{8.3} &
  \multicolumn{1}{c|}{20.4} &
  \multicolumn{1}{c|}{54.7} &
  \multicolumn{1}{c|}{57.6} &
  55.2 \\
\hline
\end{tabular}}
\end{table*}

\begin{table}[]
\centering
\caption{Performance comparison (in \%) between various baseline and state-of-the-art works for VA estimation}\label{table7}
\scalebox{.85}{
\begin{tabular}{|c|cccc|cccc|}
\hline
\multirow{2}{*}{\textbf{Model}} &
  \multicolumn{4}{c|}{\textbf{AFEW}} &
  \multicolumn{4}{c|}{\textbf{AffectNet-VA}} \\ \cline{2-9} 
 &
  \multicolumn{1}{c|}{\textbf{Test}} &
  \multicolumn{1}{c|}{\textbf{Age}} &
  \multicolumn{1}{c|}{\textbf{Gen.}} &
  \textbf{Race} &
  \multicolumn{1}{c|}{\textbf{Test}} &
  \multicolumn{1}{c|}{\textbf{Age}} &
  \multicolumn{1}{c|}{\textbf{Gen.}} &
  \textbf{Race} \\ \hline
\textbf{ResNet18} &
  \multicolumn{1}{c|}{40.1} &
  \multicolumn{1}{c|}{15.4} &
  \multicolumn{1}{c|}{29.7} &
  17.8 &
  \multicolumn{1}{c|}{70.8} &
  \multicolumn{1}{c|}{39.5} &
  \multicolumn{1}{c|}{52.8} &
  51.4 \\ \hline
\textbf{ResNet50} &
  \multicolumn{1}{c|}{42.5} &
  \multicolumn{1}{c|}{16.5} &
  \multicolumn{1}{c|}{31.4} &
  20.9 &
  \multicolumn{1}{c|}{70.8} &
  \multicolumn{1}{c|}{39.5} &
  \multicolumn{1}{c|}{52.9} &
  51.3 \\ \hline
\textbf{resnext50} &
  \multicolumn{1}{c|}{47.0} &
  \multicolumn{1}{c|}{21.1} &
  \multicolumn{1}{c|}{35.3} &
  22.5 &
  \multicolumn{1}{c|}{71.1} &
  \multicolumn{1}{c|}{39.6} &
  \multicolumn{1}{c|}{53.1} &
  51.3 \\ \hline
\textbf{DenseNet121} &
  \multicolumn{1}{c|}{46.5} &
  \multicolumn{1}{c|}{16.1} &
  \multicolumn{1}{c|}{35.1} &
  21.6 &
  \multicolumn{1}{c|}{71.3} &
  \multicolumn{1}{c|}{39.9} &
  \multicolumn{1}{c|}{53.3} &
  51.6 \\ \hline
\textbf{ViT\_B\_16} &
  \multicolumn{1}{c|}{50.5} &
  \multicolumn{1}{c|}{20.9} &
  \multicolumn{1}{c|}{36.1} &
  24.7 &
  \multicolumn{1}{c|}{71.4} &
  \multicolumn{1}{c|}{39.8} &
  \multicolumn{1}{c|}{53.3} &
  51.4 \\ \hline
\textbf{VGG16} &
  \multicolumn{1}{c|}{50.0} &
  \multicolumn{1}{c|}{18.5} &
  \multicolumn{1}{c|}{36.7} &
  22.4 &
  \multicolumn{1}{c|}{70.8} &
  \multicolumn{1}{c|}{39.4} &
  \multicolumn{1}{c|}{52.9} &
  50.8 \\
\hline
\end{tabular}}
\end{table}

\begin{table}[]
\centering
\caption{Performance comparison (in \%) between various baseline and state-of-the-art works for VA estimation}\label{table7}
\begin{tabular}{|c|l|l|l|l|}
\hline
\textbf{Model} & \multicolumn{1}{c|}{\textbf{Test}} & \multicolumn{1}{c|}{\textbf{Age}} & \multicolumn{1}{c|}{\textbf{Gen.}} & \multicolumn{1}{c|}{\textbf{Race}} \\ \hline
\textbf{ResNet18}    & 70.8 & 39.5 & 52.8 & 51.4 \\ \hline
\textbf{ResNet50}    & 70.8 & 39.5 & 52.9 & 51.3 \\ \hline
\textbf{resnext50}   & 71.1 & 39.6 & 53.1 & 51.3 \\ \hline
\textbf{DenseNet121} & 71.3 & 39.9 & 53.3 & 51.6 \\ \hline
\textbf{ViT\_B\_16}  & 71.4 & 39.8 & 53.3 & 51.4 \\ \hline
\textbf{VGG16}       & 70.8 & 39.4 & 52.9 & 50.8 \\ \hline
\textbf{EffNet\_B0}  & 71.6 & 40.1 & 53.5 & 51.6 \\ \hline
\textbf{EffNet\_B7}  & 72.0 & 40.2 & 53.7 & 51.8 \\ \hline
\textbf{Swin\_B}     & 71.9 & 40.1 & 53.6 & 51.6 \\ \hline
\textbf{Swin\_V2\_B} & 71.9 & 40.0 & 53.8 & 51.6 \\ \hline
\textbf{ConvNeXt\_B} & 72.7 & 40.6 & 54.2 & 52.3 \\ \hline
\textbf{iResNet101}  & 71.5 & 39.8 & 53.4 & 51.5 \\ \hline
\textbf{FUXI}        & 74.1 & 41.2 & 55.3 & 52.8 \\ \hline
\textbf{SITU}        & 71.9 & 40.2 & 53.7 & 52.0 \\ \hline
\textbf{CTC}         & 72.2 & 40.2 & 53.9 & 51.8 \\ \hline
\textbf{NTU}         & 71.5 & 40.1 & 52.8 & 51.5 \\ \hline
\textbf{HSE-NN}      & 70.9 & 39.8 & 52.7 & 51.0 \\ \hline
\textbf{LIVIA}       & 70.4 & 40.5 & 53.2 & 51.8 \\ \hline
\end{tabular}
\end{table}

\begin{table*}[]
\centering
\caption{Performance comparison (in \%) between various baseline and state-of-the-art works for FER; for F1 higher values are wanted; for EOP and SP, a value of 0 is wanted, values in \textcolor{red} {[0,10]} indicate fair models
%
}\label{table5}
\setlength{\tabcolsep}{0.6mm}
\renewcommand{\arraystretch}{1.2}
\resizebox{\textwidth}{!}{
\begin{tabular}{|c|cccccccccc|cccccccccc|cccccccccc|}
\hline
\multirow{3}{*}{\textbf{Model}} &
  \multicolumn{10}{c|}{\textbf{AffectNet-7}} &
  \multicolumn{10}{c|}{\textbf{AffectNet-8}} &
  \multicolumn{10}{c|}{\textbf{RAF-DB}} \\ \cline{2-31} 
 &
  \multicolumn{1}{c|}{\textbf{Test}} &
  \multicolumn{3}{c|}{\textbf{Age}} &
  \multicolumn{3}{c|}{\textbf{Gender}} &
  \multicolumn{3}{c|}{\textbf{Race}} &
  \multicolumn{1}{c|}{\textbf{Test}} &
  \multicolumn{3}{c|}{\textbf{Age}} &
  \multicolumn{3}{c|}{\textbf{Gender}} &
  \multicolumn{3}{c|}{\textbf{Race}} &
  \multicolumn{1}{c|}{\textbf{Test}} &
  \multicolumn{3}{c|}{\textbf{Age}} &
  \multicolumn{3}{c|}{\textbf{Gender}} &
  \multicolumn{3}{c|}{\textbf{Race}} \\ \cline{2-31} 
 &
  \multicolumn{1}{c|}{\textbf{F1}} &
  \multicolumn{1}{c|}{\textbf{EOP}} &
  \multicolumn{1}{c|}{\textbf{SP}} &
  \multicolumn{1}{c|}{\textbf{F1}} &
  \multicolumn{1}{c|}{\textbf{EOP}} &
  \multicolumn{1}{c|}{\textbf{SP}} &
  \multicolumn{1}{c|}{\textbf{F1}} &
  \multicolumn{1}{c|}{\textbf{EOP}} &
  \multicolumn{1}{c|}{\textbf{SP}} &
  \textbf{F1} &
  \multicolumn{1}{c|}{\textbf{F1}} &
  \multicolumn{1}{c|}{\textbf{EOP}} &
  \multicolumn{1}{c|}{\textbf{SP}} &
  \multicolumn{1}{c|}{\textbf{F1}} &
  \multicolumn{1}{c|}{\textbf{EOP}} &
  \multicolumn{1}{c|}{\textbf{SP}} &
  \multicolumn{1}{c|}{\textbf{F1}} &
  \multicolumn{1}{c|}{\textbf{EOP}} &
  \multicolumn{1}{c|}{\textbf{SP}} &
  \textbf{F1} &
  \multicolumn{1}{c|}{\textbf{F1}} &
  \multicolumn{1}{c|}{\textbf{EOP}} &
  \multicolumn{1}{c|}{\textbf{SP}} &
  \multicolumn{1}{c|}{\textbf{F1}} &
  \multicolumn{1}{c|}{\textbf{EOP}} &
  \multicolumn{1}{c|}{\textbf{SP}} &
  \multicolumn{1}{c|}{\textbf{F1}} &
  \multicolumn{1}{c|}{\textbf{EOP}} &
  \multicolumn{1}{c|}{\textbf{SP}} &
  \textbf{F1} \\
\hline
\end{tabular}}
\end{table*}

\begin{table*}[]
\centering\scriptsize
\caption{Data Statistics for AU Databases  for the New and Original Partitions}\label{table4}
\setlength{\tabcolsep}{1mm}
\scalebox{1}{
%
}}} &
  \textbf{Female} &
  \multicolumn{1}{c|}{111934} &
  \multicolumn{1}{c|}{63521} &
  \multicolumn{1}{c|}{9690} &
  38723 &
  \multicolumn{1}{c|}{11716} &
  \multicolumn{1}{c|}{10132} &
  \multicolumn{1}{c|}{11996} &
  \multicolumn{1}{c|}{3261} &
  6591 &
  \multicolumn{1}{c|}{45736} &
  \multicolumn{1}{c|}{11120} &
  \multicolumn{1}{c|}{30452} &
  \multicolumn{1}{c|}{13907} &
  23265 &
  \multicolumn{1}{c|}{2301} &
  \multicolumn{1}{c|}{1250} &
  \multicolumn{1}{c|}{328} &
  723 \\ \cline{2-20} 
\multicolumn{1}{|c|}{} &
  \textbf{Male} &
  \multicolumn{1}{c|}{142619} &
  \multicolumn{1}{c|}{77520} &
  \multicolumn{1}{c|}{19380} &
  45719 &
  \multicolumn{1}{c|}{12892} &
  \multicolumn{1}{c|}{10188} &
  \multicolumn{1}{c|}{12678} &
  \multicolumn{1}{c|}{3444} &
  6958 &
  \multicolumn{1}{c|}{62813} &
  \multicolumn{1}{c|}{13525} &
  \multicolumn{1}{c|}{39166} &
  \multicolumn{1}{c|}{9778} &
  16626 &
  \multicolumn{1}{c|}{2239} &
  \multicolumn{1}{c|}{1219} &
  \multicolumn{1}{c|}{317} &
  703 \\ \hline
\multicolumn{1}{|c|}{\multirow{4}{*}{\textbf{Race}}} &
  \textbf{Asian} &
  \multicolumn{1}{c|}{28986} &
  \multicolumn{1}{c|}{9690} &
  \multicolumn{1}{c|}{/} &
  19296 &
  \multicolumn{1}{c|}{1410} &
  \multicolumn{1}{c|}{1043} &
  \multicolumn{1}{c|}{1340} &
  \multicolumn{1}{c|}{359} &
  754 &
  \multicolumn{1}{c|}{2641} &
  \multicolumn{1}{c|}{/} &
  \multicolumn{1}{c|}{/} &
  \multicolumn{1}{c|}{/} &
  2641 &
  \multicolumn{1}{c|}{453} &
  \multicolumn{1}{c|}{243} &
  \multicolumn{1}{c|}{58} &
  152 \\ \cline{2-20} 
\multicolumn{1}{|c|}{} &
  \textbf{Black} &
  \multicolumn{1}{c|}{9690} &
  \multicolumn{1}{c|}{/} &
  \multicolumn{1}{c|}{/} &
  9690 &
  \multicolumn{1}{c|}{1979} &
  \multicolumn{1}{c|}{1935} &
  \multicolumn{1}{c|}{2144} &
  \multicolumn{1}{c|}{579} &
  1191 &
  \multicolumn{1}{c|}{7989} &
  \multicolumn{1}{c|}{4891} &
  \multicolumn{1}{c|}{9050} &
  \multicolumn{1}{c|}{/} &
  3830 &
  \multicolumn{1}{c|}{274} &
  \multicolumn{1}{c|}{144} &
  \multicolumn{1}{c|}{34} &
  96 \\ \cline{2-20} 
\multicolumn{1}{|c|}{} &
  \textbf{Indian} &
  \multicolumn{1}{c|}{9690} &
  \multicolumn{1}{c|}{/} &
  \multicolumn{1}{c|}{/} &
  9690 &
  \multicolumn{1}{c|}{1442} &
  \multicolumn{1}{c|}{1210} &
  \multicolumn{1}{c|}{1450} &
  \multicolumn{1}{c|}{388} &
  814 &
  \multicolumn{1}{c|}{/} &
  \multicolumn{1}{c|}{/} &
  \multicolumn{1}{c|}{/} &
  \multicolumn{1}{c|}{/} &
  / &
  \multicolumn{1}{c|}{252} &
  \multicolumn{1}{c|}{131} &
  \multicolumn{1}{c|}{29} &
  92 \\ \cline{2-20} 
\multicolumn{1}{|c|}{} &
  \textbf{\begin{tabular}[c]{@{}c@{}}White\end{tabular}} &
  \multicolumn{1}{c|}{206187} &
  \multicolumn{1}{c|}{131351} &
  \multicolumn{1}{c|}{29070} &
  45766 &
  \multicolumn{1}{c|}{19777} &
  \multicolumn{1}{c|}{16132} &
  \multicolumn{1}{c|}{19740} &
  \multicolumn{1}{c|}{5379} &
  10790 &
  \multicolumn{1}{c|}{97155} &
  \multicolumn{1}{c|}{18505} &
  \multicolumn{1}{c|}{60568} &
  \multicolumn{1}{c|}{23685} &
  31407 &
  \multicolumn{1}{c|}{3561} &
  \multicolumn{1}{c|}{1951} &
  \multicolumn{1}{c|}{524} &
  1086 \\ \hline
\multicolumn{1}{|c|}{\multirow{9}{*}{\textbf{Age}}} &
  \textbf{0-2} &
  \multicolumn{1}{c|}{/} &
  \multicolumn{1}{c|}{/} &
  \multicolumn{1}{c|}{/} &
  / &
  \multicolumn{1}{c|}{583} &
  \multicolumn{1}{c|}{262} &
  \multicolumn{1}{c|}{460} &
  \multicolumn{1}{c|}{123} &
  262 &
  \multicolumn{1}{c|}{/} &
  \multicolumn{1}{c|}{/} &
  \multicolumn{1}{c|}{/} &
  \multicolumn{1}{c|}{/} &
  / &
  \multicolumn{1}{c|}{352} &
  \multicolumn{1}{c|}{191} &
  \multicolumn{1}{c|}{48} &
  113 \\ \cline{2-20} 
\multicolumn{1}{|c|}{} &
  \textbf{03-09} &
  \multicolumn{1}{c|}{/} &
  \multicolumn{1}{c|}{/} &
  \multicolumn{1}{c|}{/} &
  / &
  \multicolumn{1}{c|}{939} &
  \multicolumn{1}{c|}{488} &
  \multicolumn{1}{c|}{782} &
  \multicolumn{1}{c|}{210} &
  435 &
  \multicolumn{1}{c|}{/} &
  \multicolumn{1}{c|}{/} &
  \multicolumn{1}{c|}{/} &
  \multicolumn{1}{c|}{/} &
  / &
  \multicolumn{1}{c|}{585} &
  \multicolumn{1}{c|}{318} &
  \multicolumn{1}{c|}{83} &
  184 \\ \cline{2-20} 
\multicolumn{1}{|c|}{} &
  \textbf{10-19} &
  \multicolumn{1}{c|}{9690} &
  \multicolumn{1}{c|}{/} &
  \multicolumn{1}{c|}{/} &
  9690 &
  \multicolumn{1}{c|}{1047} &
  \multicolumn{1}{c|}{754} &
  \multicolumn{1}{c|}{986} &
  \multicolumn{1}{c|}{266} &
  549 &
  \multicolumn{1}{c|}{/} &
  \multicolumn{1}{c|}{/} &
  \multicolumn{1}{c|}{/} &
  \multicolumn{1}{c|}{/} &
  / &
  \multicolumn{1}{c|}{222} &
  \multicolumn{1}{c|}{117} &
  \multicolumn{1}{c|}{30} &
  75 \\ \cline{2-20} 
\multicolumn{1}{|c|}{} &
  \textbf{20-29} &
  \multicolumn{1}{c|}{208834} &
  \multicolumn{1}{c|}{131351} &
  \multicolumn{1}{c|}{29070} &
  48413 &
  \multicolumn{1}{c|}{10101} &
  \multicolumn{1}{c|}{8912} &
  \multicolumn{1}{c|}{10454} &
  \multicolumn{1}{c|}{2849} &
  5710 &
  \multicolumn{1}{c|}{108549} &
  \multicolumn{1}{c|}{24645} &
  \multicolumn{1}{c|}{69618} &
  \multicolumn{1}{c|}{23685} &
  39891 &
  \multicolumn{1}{c|}{1662} &
  \multicolumn{1}{c|}{912} &
  \multicolumn{1}{c|}{246} &
  504 \\ \cline{2-20} 
\multicolumn{1}{|c|}{} &
  \textbf{30-39} &
  \multicolumn{1}{c|}{19333} &
  \multicolumn{1}{c|}{/} &
  \multicolumn{1}{c|}{/} &
  19333 &
  \multicolumn{1}{c|}{5026} &
  \multicolumn{1}{c|}{4206} &
  \multicolumn{1}{c|}{5075} &
  \multicolumn{1}{c|}{1380} &
  2777 &
  \multicolumn{1}{c|}{/} &
  \multicolumn{1}{c|}{/} &
  \multicolumn{1}{c|}{/} &
  \multicolumn{1}{c|}{/} &
  / &
  \multicolumn{1}{c|}{1068} &
  \multicolumn{1}{c|}{584} &
  \multicolumn{1}{c|}{156} &
  328 \\ \cline{2-20} 
\multicolumn{1}{|c|}{} &
  \textbf{40-49} &
  \multicolumn{1}{c|}{16696} &
  \multicolumn{1}{c|}{9690} &
  \multicolumn{1}{c|}{/} &
  7006 &
  \multicolumn{1}{c|}{3128} &
  \multicolumn{1}{c|}{2581} &
  \multicolumn{1}{c|}{3136} &
  \multicolumn{1}{c|}{852} &
  1721 &
  \multicolumn{1}{c|}{/} &
  \multicolumn{1}{c|}{/} &
  \multicolumn{1}{c|}{/} &
  \multicolumn{1}{c|}{/} &
  / &
  \multicolumn{1}{c|}{385} &
  \multicolumn{1}{c|}{208} &
  \multicolumn{1}{c|}{53} &
  124 \\ \cline{2-20} 
\multicolumn{1}{|c|}{} &
  \textbf{50-59} &
  \multicolumn{1}{c|}{/} &
  \multicolumn{1}{c|}{/} &
  \multicolumn{1}{c|}{/} &
  / &
  \multicolumn{1}{c|}{2484} &
  \multicolumn{1}{c|}{2093} &
  \multicolumn{1}{c|}{2513} &
  \multicolumn{1}{c|}{683} &
  1381 &
  \multicolumn{1}{c|}{/} &
  \multicolumn{1}{c|}{/} &
  \multicolumn{1}{c|}{/} &
  \multicolumn{1}{c|}{/} &
  / &
  \multicolumn{1}{c|}{161} &
  \multicolumn{1}{c|}{86} &
  \multicolumn{1}{c|}{20} &
  55 \\ \cline{2-20} 
\multicolumn{1}{|c|}{} &
  \textbf{60-69} &
  \multicolumn{1}{c|}{/} &
  \multicolumn{1}{c|}{/} &
  \multicolumn{1}{c|}{/} &
  / &
  \multicolumn{1}{c|}{1041} &
  \multicolumn{1}{c|}{834} &
  \multicolumn{1}{c|}{1026} &
  \multicolumn{1}{c|}{278} &
  571 &
  \multicolumn{1}{c|}{/} &
  \multicolumn{1}{c|}{/} &
  \multicolumn{1}{c|}{/} &
  \multicolumn{1}{c|}{/} &
  / &
  \multicolumn{1}{c|}{55} &
  \multicolumn{1}{c|}{27} &
  \multicolumn{1}{c|}{5} &
  23 \\ \cline{2-20} 
\multicolumn{1}{|c|}{} &
  \textbf{70+} &
  \multicolumn{1}{c|}{/} &
  \multicolumn{1}{c|}{/} &
  \multicolumn{1}{c|}{/} &
  / &
  \multicolumn{1}{c|}{259} &
  \multicolumn{1}{c|}{190} &
  \multicolumn{1}{c|}{242} &
  \multicolumn{1}{c|}{64} &
  143 &
  \multicolumn{1}{c|}{/} &
  \multicolumn{1}{c|}{/} &
  \multicolumn{1}{c|}{/} &
  \multicolumn{1}{c|}{/} &
  / &
  \multicolumn{1}{c|}{50} &
  \multicolumn{1}{c|}{26} &
  \multicolumn{1}{c|}{4} &
  20 \\ \hline
\multicolumn{1}{|c|}{\multirow{2}{*}{\textbf{AU1}}} &
  \textbf{0} &
  \multicolumn{1}{c|}{237262} &
  \multicolumn{1}{c|}{131616} &
  \multicolumn{1}{c|}{26988} &
  78574 &
  \multicolumn{1}{c|}{21519} &
  \multicolumn{1}{c|}{18591} &
  \multicolumn{1}{c|}{21610} &
  \multicolumn{1}{c|}{5934} &
  9965 &
  \multicolumn{1}{c|}{104538} &
  \multicolumn{1}{c|}{23540} &
  \multicolumn{1}{c|}{66619} &
  \multicolumn{1}{c|}{23254} &
  38205 &
  \multicolumn{1}{c|}{3832} &
  \multicolumn{1}{c|}{1874} &
  \multicolumn{1}{c|}{481} &
  1119 \\ \cline{2-20} 
\multicolumn{1}{|c|}{} &
  \textbf{1} &
  \multicolumn{1}{c|}{17375} &
  \multicolumn{1}{c|}{9425} &
  \multicolumn{1}{c|}{2082} &
  5868 &
  \multicolumn{1}{c|}{1452} &
  \multicolumn{1}{c|}{1170} &
  \multicolumn{1}{c|}{1419} &
  \multicolumn{1}{c|}{541} &
  532 &
  \multicolumn{1}{c|}{4011} &
  \multicolumn{1}{c|}{1105} &
  \multicolumn{1}{c|}{299} &
  \multicolumn{1}{c|}{431} &
  1686 &
  \multicolumn{1}{c|}{1076} &
  \multicolumn{1}{c|}{595} &
  \multicolumn{1}{c|}{164} &
  307 \\ \hline
\multicolumn{1}{|c|}{\multirow{2}{*}{\textbf{AU2}}} &
  \textbf{0} &
  \multicolumn{1}{c|}{240116} &
  \multicolumn{1}{c|}{133559} &
  \multicolumn{1}{c|}{28558} &
  77962 &
  \multicolumn{1}{c|}{22476} &
  \multicolumn{1}{c|}{16129} &
  \multicolumn{1}{c|}{22536} &
  \multicolumn{1}{c|}{4892} &
  11111 &
  \multicolumn{1}{c|}{94022} &
  \multicolumn{1}{c|}{21636} &
  \multicolumn{1}{c|}{60801} &
  \multicolumn{1}{c|}{20765} &
  34092 &
  \multicolumn{1}{c|}{4113} &
  \multicolumn{1}{c|}{2057} &
  \multicolumn{1}{c|}{524} &
  1172 \\ \cline{2-20} 
\multicolumn{1}{|c|}{} &
  \textbf{1} &
  \multicolumn{1}{c|}{14521} &
  \multicolumn{1}{c|}{7482} &
  \multicolumn{1}{c|}{512} &
  6480 &
  \multicolumn{1}{c|}{689} &
  \multicolumn{1}{c|}{713} &
  \multicolumn{1}{c|}{694} &
  \multicolumn{1}{c|}{307} &
  401 &
  \multicolumn{1}{c|}{14527} &
  \multicolumn{1}{c|}{3009} &
  \multicolumn{1}{c|}{8817} &
  \multicolumn{1}{c|}{2920} &
  5799 &
  \multicolumn{1}{c|}{795} &
  \multicolumn{1}{c|}{412} &
  \multicolumn{1}{c|}{121} &
  254 \\ \hline
\multicolumn{1}{|c|}{\multirow{2}{*}{\textbf{AU4}}} &
  \textbf{0} &
  \multicolumn{1}{c|}{206157} &
  \multicolumn{1}{c|}{115349} &
  \multicolumn{1}{c|}{25952} &
  64819 &
  \multicolumn{1}{c|}{21134} &
  \multicolumn{1}{c|}{8656} &
  \multicolumn{1}{c|}{21257} &
  \multicolumn{1}{c|}{6196} &
  2284 &
  \multicolumn{1}{c|}{104560} &
  \multicolumn{1}{c|}{23809} &
  \multicolumn{1}{c|}{65947} &
  \multicolumn{1}{c|}{23219} &
  39203 &
  \multicolumn{1}{c|}{3091} &
  \multicolumn{1}{c|}{1475} &
  \multicolumn{1}{c|}{410} &
  850 \\ \cline{2-20} 
\multicolumn{1}{|c|}{} &
  \textbf{1} &
  \multicolumn{1}{c|}{48480} &
  \multicolumn{1}{c|}{25692} &
  \multicolumn{1}{c|}{3118} &
  19623 &
  \multicolumn{1}{c|}{2857} &
  \multicolumn{1}{c|}{610} &
  \multicolumn{1}{c|}{2767} &
  \multicolumn{1}{c|}{487} &
  202 &
  \multicolumn{1}{c|}{3989} &
  \multicolumn{1}{c|}{836} &
  \multicolumn{1}{c|}{3671} &
  \multicolumn{1}{c|}{466} &
  688 &
  \multicolumn{1}{c|}{1817} &
  \multicolumn{1}{c|}{994} &
  \multicolumn{1}{c|}{235} &
  576 \\ \hline
\multicolumn{1}{|c|}{\multirow{2}{*}{\textbf{AU5}}} &
  \textbf{0} &
  \multicolumn{1}{c|}{249300} &
  \multicolumn{1}{c|}{136960} &
  \multicolumn{1}{c|}{28768} &
  83488 &
  \multicolumn{1}{c|}{22255} &
  \multicolumn{1}{c|}{18766} &
  \multicolumn{1}{c|}{22339} &
  \multicolumn{1}{c|}{6188} &
  12436 &
  \multicolumn{1}{c|}{105949} &
  \multicolumn{1}{c|}{24248} &
  \multicolumn{1}{c|}{68404} &
  \multicolumn{1}{c|}{23519} &
  38274 &
  \multicolumn{1}{c|}{3923} &
  \multicolumn{1}{c|}{1870} &
  \multicolumn{1}{c|}{505} &
  1201 \\ \cline{2-20} 
\multicolumn{1}{|c|}{} &
  \textbf{1} &
  \multicolumn{1}{c|}{5337} &
  \multicolumn{1}{c|}{4081} &
  \multicolumn{1}{c|}{302} &
  954 &
  \multicolumn{1}{c|}{875} &
  \multicolumn{1}{c|}{1287} &
  \multicolumn{1}{c|}{855} &
  \multicolumn{1}{c|}{411} &
  890 &
  \multicolumn{1}{c|}{2600} &
  \multicolumn{1}{c|}{397} &
  \multicolumn{1}{c|}{1214} &
  \multicolumn{1}{c|}{166} &
  1617 &
  \multicolumn{1}{c|}{985} &
  \multicolumn{1}{c|}{599} &
  \multicolumn{1}{c|}{140} &
  225 \\ \hline
\multicolumn{1}{|c|}{\multirow{2}{*}{\textbf{AU6}}} &
  \textbf{0} &
  \multicolumn{1}{c|}{216544} &
  \multicolumn{1}{c|}{116881} &
  \multicolumn{1}{c|}{26204} &
  73375 &
  \multicolumn{1}{c|}{18749} &
  \multicolumn{1}{c|}{15115} &
  \multicolumn{1}{c|}{18691} &
  \multicolumn{1}{c|}{4777} &
  10337 &
  \multicolumn{1}{c|}{77762} &
  \multicolumn{1}{c|}{17819} &
  \multicolumn{1}{c|}{48664} &
  \multicolumn{1}{c|}{16648} &
  30269 &
  \multicolumn{1}{c|}{4458} &
  \multicolumn{1}{c|}{2251} &
  \multicolumn{1}{c|}{585} &
  1260 \\ \cline{2-20} 
\multicolumn{1}{|c|}{} &
  \textbf{1} &
  \multicolumn{1}{c|}{38093} &
  \multicolumn{1}{c|}{24160} &
  \multicolumn{1}{c|}{2866} &
  11067 &
  \multicolumn{1}{c|}{4572} &
  \multicolumn{1}{c|}{4777} &
  \multicolumn{1}{c|}{4678} &
  \multicolumn{1}{c|}{1779} &
  2883 &
  \multicolumn{1}{c|}{30787} &
  \multicolumn{1}{c|}{6826} &
  \multicolumn{1}{c|}{20954} &
  \multicolumn{1}{c|}{7037} &
  9622 &
  \multicolumn{1}{c|}{450} &
  \multicolumn{1}{c|}{218} &
  \multicolumn{1}{c|}{60} &
  166 \\ \hline
\multicolumn{1}{|c|}{\multirow{2}{*}{\textbf{AU7}}} &
  \textbf{0} &
  \multicolumn{1}{c|}{/} &
  \multicolumn{1}{c|}{/} &
  \multicolumn{1}{c|}{/} &
  / &
  \multicolumn{1}{c|}{/} &
  \multicolumn{1}{c|}{/} &
  \multicolumn{1}{c|}{/} &
  \multicolumn{1}{c|}{/} &
  / &
  \multicolumn{1}{c|}{59146} &
  \multicolumn{1}{c|}{13537} &
  \multicolumn{1}{c|}{35576} &
  \multicolumn{1}{c|}{13105} &
  24002 &
  \multicolumn{1}{c|}{/} &
  \multicolumn{1}{c|}{/} &
  \multicolumn{1}{c|}{/} &
  / \\ \cline{2-20} 
\multicolumn{1}{|c|}{} &
  \textbf{1} &
  \multicolumn{1}{c|}{/} &
  \multicolumn{1}{c|}{/} &
  \multicolumn{1}{c|}{/} &
  / &
  \multicolumn{1}{c|}{/} &
  \multicolumn{1}{c|}{/} &
  \multicolumn{1}{c|}{/} &
  \multicolumn{1}{c|}{/} &
  / &
  \multicolumn{1}{c|}{49403} &
  \multicolumn{1}{c|}{11108} &
  \multicolumn{1}{c|}{34042} &
  \multicolumn{1}{c|}{10580} &
  15889 &
  \multicolumn{1}{c|}{/} &
  \multicolumn{1}{c|}{/} &
  \multicolumn{1}{c|}{/} &
  / \\ \hline
\multicolumn{1}{|c|}{\multirow{2}{*}{\textbf{AU9}}} &
  \textbf{0} &
  \multicolumn{1}{c|}{240853} &
  \multicolumn{1}{c|}{133531} &
  \multicolumn{1}{c|}{27402} &
  79836 &
  \multicolumn{1}{c|}{23208} &
  \multicolumn{1}{c|}{19942} &
  \multicolumn{1}{c|}{23300} &
  \multicolumn{1}{c|}{6422} &
  13362 &
  \multicolumn{1}{c|}{107030} &
  \multicolumn{1}{c|}{24267} &
  \multicolumn{1}{c|}{68329} &
  \multicolumn{1}{c|}{23590} &
  39378 &
  \multicolumn{1}{c|}{4134} &
  \multicolumn{1}{c|}{2084} &
  \multicolumn{1}{c|}{557} &
  1132 \\ \cline{2-20} 
\multicolumn{1}{|c|}{} &
  \textbf{1} &
  \multicolumn{1}{c|}{13784} &
  \multicolumn{1}{c|}{7510} &
  \multicolumn{1}{c|}{1668} &
  4606 &
  \multicolumn{1}{c|}{505} &
  \multicolumn{1}{c|}{143} &
  \multicolumn{1}{c|}{467} &
  \multicolumn{1}{c|}{109} &
  68 &
  \multicolumn{1}{c|}{1519} &
  \multicolumn{1}{c|}{378} &
  \multicolumn{1}{c|}{1289} &
  \multicolumn{1}{c|}{95} &
  513 &
  \multicolumn{1}{c|}{774} &
  \multicolumn{1}{c|}{385} &
  \multicolumn{1}{c|}{88} &
  294 \\ \hline
\multicolumn{1}{|c|}{\multirow{2}{*}{\textbf{AU10}}} &
  \textbf{0} &
  \multicolumn{1}{c|}{/} &
  \multicolumn{1}{c|}{/} &
  \multicolumn{1}{c|}{/} &
  / &
  \multicolumn{1}{c|}{/} &
  \multicolumn{1}{c|}{/} &
  \multicolumn{1}{c|}{/} &
  \multicolumn{1}{c|}{/} &
  / &
  \multicolumn{1}{c|}{81809} &
  \multicolumn{1}{c|}{18566} &
  \multicolumn{1}{c|}{50586} &
  \multicolumn{1}{c|}{17899} &
  31890 &
  \multicolumn{1}{c|}{3518} &
  \multicolumn{1}{c|}{1808} &
  \multicolumn{1}{c|}{440} &
  917 \\ \cline{2-20} 
\multicolumn{1}{|c|}{} &
  \textbf{1} &
  \multicolumn{1}{c|}{/} &
  \multicolumn{1}{c|}{/} &
  \multicolumn{1}{c|}{/} &
  / &
  \multicolumn{1}{c|}{/} &
  \multicolumn{1}{c|}{/} &
  \multicolumn{1}{c|}{/} &
  \multicolumn{1}{c|}{/} &
  / &
  \multicolumn{1}{c|}{26740} &
  \multicolumn{1}{c|}{6079} &
  \multicolumn{1}{c|}{19032} &
  \multicolumn{1}{c|}{5786} &
  8001 &
  \multicolumn{1}{c|}{1390} &
  \multicolumn{1}{c|}{661} &
  \multicolumn{1}{c|}{205} &
  509 \\ \hline
\multicolumn{1}{|c|}{\multirow{2}{*}{\textbf{AU11}}} &
  \textbf{0} &
  \multicolumn{1}{c|}{/} &
  \multicolumn{1}{c|}{/} &
  \multicolumn{1}{c|}{/} &
  / &
  \multicolumn{1}{c|}{/} &
  \multicolumn{1}{c|}{/} &
  \multicolumn{1}{c|}{/} &
  \multicolumn{1}{c|}{/} &
  / &
  \multicolumn{1}{c|}{93623} &
  \multicolumn{1}{c|}{20935} &
  \multicolumn{1}{c|}{58501} &
  \multicolumn{1}{c|}{20663} &
  35394 &
  \multicolumn{1}{c|}{/} &
  \multicolumn{1}{c|}{/} &
  \multicolumn{1}{c|}{/} &
  / \\ \cline{2-20} 
\multicolumn{1}{|c|}{} &
  \textbf{1} &
  \multicolumn{1}{c|}{/} &
  \multicolumn{1}{c|}{/} &
  \multicolumn{1}{c|}{/} &
  / &
  \multicolumn{1}{c|}{/} &
  \multicolumn{1}{c|}{/} &
  \multicolumn{1}{c|}{/} &
  \multicolumn{1}{c|}{/} &
  / &
  \multicolumn{1}{c|}{14926} &
  \multicolumn{1}{c|}{3710} &
  \multicolumn{1}{c|}{11117} &
  \multicolumn{1}{c|}{3022} &
  4497 &
  \multicolumn{1}{c|}{/} &
  \multicolumn{1}{c|}{/} &
  \multicolumn{1}{c|}{/} &
  / \\ \hline
\multicolumn{1}{|c|}{\multirow{2}{*}{\textbf{AU12}}} &
  \textbf{0} &
  \multicolumn{1}{c|}{/} &
  \multicolumn{1}{c|}{/} &
  \multicolumn{1}{c|}{/} &
  / &
  \multicolumn{1}{c|}{12199} &
  \multicolumn{1}{c|}{9962} &
  \multicolumn{1}{c|}{12147} &
  \multicolumn{1}{c|}{2520} &
  7465 &
  \multicolumn{1}{c|}{76528} &
  \multicolumn{1}{c|}{17459} &
  \multicolumn{1}{c|}{48040} &
  \multicolumn{1}{c|}{16259} &
  29688 &
  \multicolumn{1}{c|}{3640} &
  \multicolumn{1}{c|}{1885} &
  \multicolumn{1}{c|}{467} &
  938 \\ \cline{2-20} 
\multicolumn{1}{|c|}{} &
  \textbf{1} &
  \multicolumn{1}{c|}{/} &
  \multicolumn{1}{c|}{/} &
  \multicolumn{1}{c|}{/} &
  / &
  \multicolumn{1}{c|}{7546} &
  \multicolumn{1}{c|}{6641} &
  \multicolumn{1}{c|}{7633} &
  \multicolumn{1}{c|}{1703} &
  4838 &
  \multicolumn{1}{c|}{32021} &
  \multicolumn{1}{c|}{7186} &
  \multicolumn{1}{c|}{21578} &
  \multicolumn{1}{c|}{7426} &
  10203 &
  \multicolumn{1}{c|}{1268} &
  \multicolumn{1}{c|}{584} &
  \multicolumn{1}{c|}{178} &
  488 \\ \hline
\multicolumn{1}{|c|}{\multirow{2}{*}{\textbf{AU15}}} &
  \textbf{0} &
  \multicolumn{1}{c|}{193822} &
  \multicolumn{1}{c|}{107287} &
  \multicolumn{1}{c|}{20242} &
  66209 &
  \multicolumn{1}{c|}{/} &
  \multicolumn{1}{c|}{/} &
  \multicolumn{1}{c|}{/} &
  \multicolumn{1}{c|}{/} &
  / &
  \multicolumn{1}{c|}{97013} &
  \multicolumn{1}{c|}{22295} &
  \multicolumn{1}{c|}{61001} &
  \multicolumn{1}{c|}{21454} &
  36853 &
  \multicolumn{1}{c|}{/} &
  \multicolumn{1}{c|}{/} &
  \multicolumn{1}{c|}{/} &
  / \\ \cline{2-20} 
\multicolumn{1}{|c|}{} &
  \textbf{1} &
  \multicolumn{1}{c|}{60815} &
  \multicolumn{1}{c|}{33754} &
  \multicolumn{1}{c|}{8828} &
  18233 &
  \multicolumn{1}{c|}{/} &
  \multicolumn{1}{c|}{/} &
  \multicolumn{1}{c|}{/} &
  \multicolumn{1}{c|}{/} &
  / &
  \multicolumn{1}{c|}{11536} &
  \multicolumn{1}{c|}{2350} &
  \multicolumn{1}{c|}{8617} &
  \multicolumn{1}{c|}{2231} &
  3038 &
  \multicolumn{1}{c|}{/} &
  \multicolumn{1}{c|}{/} &
  \multicolumn{1}{c|}{/} &
  / \\ \hline
\multicolumn{1}{|c|}{\multirow{2}{*}{\textbf{AU16}}} &
  \textbf{0} &
  \multicolumn{1}{c|}{/} &
  \multicolumn{1}{c|}{/} &
  \multicolumn{1}{c|}{/} &
  / &
  \multicolumn{1}{c|}{/} &
  \multicolumn{1}{c|}{/} &
  \multicolumn{1}{c|}{/} &
  \multicolumn{1}{c|}{/} &
  / &
  \multicolumn{1}{c|}{/} &
  \multicolumn{1}{c|}{/} &
  \multicolumn{1}{c|}{/} &
  \multicolumn{1}{c|}{/} &
  / &
  \multicolumn{1}{c|}{4188} &
  \multicolumn{1}{c|}{2167} &
  \multicolumn{1}{c|}{538} &
  1127 \\ \cline{2-20} 
\multicolumn{1}{|c|}{} &
  \textbf{1} &
  \multicolumn{1}{c|}{/} &
  \multicolumn{1}{c|}{/} &
  \multicolumn{1}{c|}{/} &
  / &
  \multicolumn{1}{c|}{/} &
  \multicolumn{1}{c|}{/} &
  \multicolumn{1}{c|}{/} &
  \multicolumn{1}{c|}{/} &
  / &
  \multicolumn{1}{c|}{/} &
  \multicolumn{1}{c|}{/} &
  \multicolumn{1}{c|}{/} &
  \multicolumn{1}{c|}{/} &
  / &
  \multicolumn{1}{c|}{720} &
  \multicolumn{1}{c|}{302} &
  \multicolumn{1}{c|}{107} &
  299 \\ \hline
\multicolumn{1}{|c|}{\multirow{2}{*}{\textbf{AU17}}} &
  \textbf{0} &
  \multicolumn{1}{c|}{229289} &
  \multicolumn{1}{c|}{128914} &
  \multicolumn{1}{c|}{26780} &
  73511 &
  \multicolumn{1}{c|}{22288} &
  \multicolumn{1}{c|}{19674} &
  \multicolumn{1}{c|}{22378} &
  \multicolumn{1}{c|}{6302} &
  13224 &
  \multicolumn{1}{c|}{75826} &
  \multicolumn{1}{c|}{16661} &
  \multicolumn{1}{c|}{49223} &
  \multicolumn{1}{c|}{17183} &
  26081 &
  \multicolumn{1}{c|}{4367} &
  \multicolumn{1}{c|}{2148} &
  \multicolumn{1}{c|}{560} &
  1291 \\ \cline{2-20} 
\multicolumn{1}{|c|}{} &
  \textbf{1} &
  \multicolumn{1}{c|}{25348} &
  \multicolumn{1}{c|}{12127} &
  \multicolumn{1}{c|}{2290} &
  10931 &
  \multicolumn{1}{c|}{492} &
  \multicolumn{1}{c|}{184} &
  \multicolumn{1}{c|}{463} &
  \multicolumn{1}{c|}{102} &
  107 &
  \multicolumn{1}{c|}{32723} &
  \multicolumn{1}{c|}{7984} &
  \multicolumn{1}{c|}{20395} &
  \multicolumn{1}{c|}{6502} &
  13810 &
  \multicolumn{1}{c|}{541} &
  \multicolumn{1}{c|}{321} &
  \multicolumn{1}{c|}{85} &
  135 \\ \hline
\multicolumn{1}{|c|}{\multirow{2}{*}{\textbf{AU20}}} &
  \textbf{0} &
  \multicolumn{1}{c|}{246028} &
  \multicolumn{1}{c|}{135825} &
  \multicolumn{1}{c|}{28586} &
  81533 &
  \multicolumn{1}{c|}{24376} &
  \multicolumn{1}{c|}{20049} &
  \multicolumn{1}{c|}{24421} &
  \multicolumn{1}{c|}{6599} &
  13335 &
  \multicolumn{1}{c|}{/} &
  \multicolumn{1}{c|}{/} &
  \multicolumn{1}{c|}{/} &
  \multicolumn{1}{c|}{/} &
  / &
  \multicolumn{1}{c|}{/} &
  \multicolumn{1}{c|}{/} &
  \multicolumn{1}{c|}{/} &
  / \\ \cline{2-20} 
\multicolumn{1}{|c|}{} &
  \textbf{1} &
  \multicolumn{1}{c|}{8609} &
  \multicolumn{1}{c|}{5216} &
  \multicolumn{1}{c|}{484} &
  2909 &
  \multicolumn{1}{c|}{134} &
  \multicolumn{1}{c|}{146} &
  \multicolumn{1}{c|}{126} &
  \multicolumn{1}{c|}{56} &
  97 &
  \multicolumn{1}{c|}{/} &
  \multicolumn{1}{c|}{/} &
  \multicolumn{1}{c|}{/} &
  \multicolumn{1}{c|}{/} &
  / &
  \multicolumn{1}{c|}{/} &
  \multicolumn{1}{c|}{/} &
  \multicolumn{1}{c|}{/} &
  / \\ \hline
\multicolumn{1}{|c|}{\multirow{2}{*}{\textbf{AU23}}} &
  \textbf{0} &
  \multicolumn{1}{c|}{/} &
  \multicolumn{1}{c|}{/} &
  \multicolumn{1}{c|}{/} &
  / &
  \multicolumn{1}{c|}{/} &
  \multicolumn{1}{c|}{/} &
  \multicolumn{1}{c|}{/} &
  \multicolumn{1}{c|}{/} &
  / &
  \multicolumn{1}{c|}{81540} &
  \multicolumn{1}{c|}{18364} &
  \multicolumn{1}{c|}{52567} &
  \multicolumn{1}{c|}{18019} &
  29318 &
  \multicolumn{1}{c|}{/} &
  \multicolumn{1}{c|}{/} &
  \multicolumn{1}{c|}{/} &
  / \\ \cline{2-20} 
\multicolumn{1}{|c|}{} &
  \textbf{1} &
  \multicolumn{1}{c|}{/} &
  \multicolumn{1}{c|}{/} &
  \multicolumn{1}{c|}{/} &
  / &
  \multicolumn{1}{c|}{/} &
  \multicolumn{1}{c|}{/} &
  \multicolumn{1}{c|}{/} &
  \multicolumn{1}{c|}{/} &
  / &
  \multicolumn{1}{c|}{27009} &
  \multicolumn{1}{c|}{6281} &
  \multicolumn{1}{c|}{17051} &
  \multicolumn{1}{c|}{5666} &
  10573 &
  \multicolumn{1}{c|}{/} &
  \multicolumn{1}{c|}{/} &
  \multicolumn{1}{c|}{/} &
  / \\ \hline
\multicolumn{1}{|c|}{\multirow{2}{*}{\textbf{AU24}}} &
  \textbf{0} &
  \multicolumn{1}{c|}{/} &
  \multicolumn{1}{c|}{/} &
  \multicolumn{1}{c|}{/} &
  / &
  \multicolumn{1}{c|}{/} &
  \multicolumn{1}{c|}{/} &
  \multicolumn{1}{c|}{/} &
  \multicolumn{1}{c|}{/} &
  / &
  \multicolumn{1}{c|}{93072} &
  \multicolumn{1}{c|}{21165} &
  \multicolumn{1}{c|}{59510} &
  \multicolumn{1}{c|}{21786} &
  32941 &
  \multicolumn{1}{c|}{/} &
  \multicolumn{1}{c|}{/} &
  \multicolumn{1}{c|}{/} &
  / \\ \cline{2-20} 
\multicolumn{1}{|c|}{} &
  \textbf{1} &
  \multicolumn{1}{c|}{/} &
  \multicolumn{1}{c|}{/} &
  \multicolumn{1}{c|}{/} &
  / &
  \multicolumn{1}{c|}{/} &
  \multicolumn{1}{c|}{/} &
  \multicolumn{1}{c|}{/} &
  \multicolumn{1}{c|}{/} &
  / &
  \multicolumn{1}{c|}{15477} &
  \multicolumn{1}{c|}{3480} &
  \multicolumn{1}{c|}{10108} &
  \multicolumn{1}{c|}{1899} &
  6950 &
  \multicolumn{1}{c|}{/} &
  \multicolumn{1}{c|}{/} &
  \multicolumn{1}{c|}{/} &
  / \\ \hline
\multicolumn{1}{|c|}{\multirow{2}{*}{\textbf{AU25}}} &
  \textbf{0} &
  \multicolumn{1}{c|}{163547} &
  \multicolumn{1}{c|}{95580} &
  \multicolumn{1}{c|}{10132} &
  57751 &
  \multicolumn{1}{c|}{11882} &
  \multicolumn{1}{c|}{10430} &
  \multicolumn{1}{c|}{11898} &
  \multicolumn{1}{c|}{3236} &
  7142 &
  \multicolumn{1}{c|}{/} &
  \multicolumn{1}{c|}{/} &
  \multicolumn{1}{c|}{/} &
  \multicolumn{1}{c|}{/} &
  / &
  \multicolumn{1}{c|}{2079} &
  \multicolumn{1}{c|}{991} &
  \multicolumn{1}{c|}{237} &
  525 \\ \cline{2-20} 
\multicolumn{1}{|c|}{} &
  \textbf{1} &
  \multicolumn{1}{c|}{91090} &
  \multicolumn{1}{c|}{45461} &
  \multicolumn{1}{c|}{18938} &
  26691 &
  \multicolumn{1}{c|}{11590} &
  \multicolumn{1}{c|}{9473} &
  \multicolumn{1}{c|}{11641} &
  \multicolumn{1}{c|}{3278} &
  6113 &
  \multicolumn{1}{c|}{/} &
  \multicolumn{1}{c|}{/} &
  \multicolumn{1}{c|}{/} &
  \multicolumn{1}{c|}{/} &
  / &
  \multicolumn{1}{c|}{2829} &
  \multicolumn{1}{c|}{1478} &
  \multicolumn{1}{c|}{408} &
  901 \\ \hline
\multicolumn{1}{|c|}{\multirow{2}{*}{\textbf{AU26}}} &
  \textbf{0} &
  \multicolumn{1}{c|}{205200} &
  \multicolumn{1}{c|}{114576} &
  \multicolumn{1}{c|}{16538} &
  74002 &
  \multicolumn{1}{c|}{21676} &
  \multicolumn{1}{c|}{18330} &
  \multicolumn{1}{c|}{21721} &
  \multicolumn{1}{c|}{6008} &
  12221 &
  \multicolumn{1}{c|}{/} &
  \multicolumn{1}{c|}{/} &
  \multicolumn{1}{c|}{/} &
  \multicolumn{1}{c|}{/} &
  / &
  \multicolumn{1}{c|}{3819} &
  \multicolumn{1}{c|}{1891} &
  \multicolumn{1}{c|}{485} &
  1092 \\ \cline{2-20} 
\multicolumn{1}{|c|}{} &
  \textbf{1} &
  \multicolumn{1}{c|}{49437} &
  \multicolumn{1}{c|}{26465} &
  \multicolumn{1}{c|}{12532} &
  10440 &
  \multicolumn{1}{c|}{2058} &
  \multicolumn{1}{c|}{1778} &
  \multicolumn{1}{c|}{2059} &
  \multicolumn{1}{c|}{624} &
  1145 &
  \multicolumn{1}{c|}{/} &
  \multicolumn{1}{c|}{/} &
  \multicolumn{1}{c|}{/} &
  \multicolumn{1}{c|}{/} &
  / &
  \multicolumn{1}{c|}{1089} &
  \multicolumn{1}{c|}{578} &
  \multicolumn{1}{c|}{160} &
  334 \\ \hline
\multicolumn{1}{|c|}{\multirow{2}{*}{\textbf{AU27}}} &
  \textbf{0} &
  \multicolumn{1}{c|}{/} &
  \multicolumn{1}{c|}{/} &
  \multicolumn{1}{c|}{/} &
  / &
  \multicolumn{1}{c|}{/} &
  \multicolumn{1}{c|}{/} &
  \multicolumn{1}{c|}{/} &
  \multicolumn{1}{c|}{/} &
  / &
  \multicolumn{1}{c|}{/} &
  \multicolumn{1}{c|}{/} &
  \multicolumn{1}{c|}{/} &
  \multicolumn{1}{c|}{/} &
  / &
  \multicolumn{1}{c|}{4098} &
  \multicolumn{1}{c|}{2087} &
  \multicolumn{1}{c|}{547} &
  1113 \\ \cline{2-20} 
\multicolumn{1}{|c|}{} &
  \textbf{1} &
  \multicolumn{1}{c|}{/} &
  \multicolumn{1}{c|}{/} &
  \multicolumn{1}{c|}{/} &
  / &
  \multicolumn{1}{c|}{/} &
  \multicolumn{1}{c|}{/} &
  \multicolumn{1}{c|}{/} &
  \multicolumn{1}{c|}{/} &
  / &
  \multicolumn{1}{c|}{/} &
  \multicolumn{1}{c|}{/} &
  \multicolumn{1}{c|}{/} &
  \multicolumn{1}{c|}{/} &
  / &
  \multicolumn{1}{c|}{810} &
  \multicolumn{1}{c|}{382} &
  \multicolumn{1}{c|}{98} &
  313 \\ \hline
\end{tabular}
}
\end{table*}


\begin{table*}[]
\caption{Performance comparison in test and valid set (in \%) between various baseline works for FER; for ACC and F1 higher values are wanted}\label{table1}
\renewcommand{\arraystretch}{1.2}
\begin{tabular}{|c|cccc|cccc|cccc|}
\hline
\multirow{3}{*}{\textbf{Model}} &
  \multicolumn{4}{c|}{\textbf{AffectNet-7}} &
  \multicolumn{4}{c|}{\textbf{AffectNet-8}} &
  \multicolumn{4}{c|}{\textbf{RAF-DB}} \\ \cline{2-13} 
 &
  \multicolumn{2}{c|}{\textbf{ACC}} &
  \multicolumn{2}{c|}{\textbf{MACRO F1}} &
  \multicolumn{2}{c|}{\textbf{ACC}} &
  \multicolumn{2}{c|}{\textbf{MACRO F1}} &
  \multicolumn{2}{c|}{\textbf{ACC}} &
  \multicolumn{2}{c|}{\textbf{MACRO F1}} \\ \cline{2-13} 
 &
  \multicolumn{1}{c|}{\textbf{Valid}} &
  \multicolumn{1}{c|}{\textbf{Test}} &
  \multicolumn{1}{c|}{\textbf{Valid}} &
  \textbf{Test} &
  \multicolumn{1}{c|}{\textbf{Valid}} &
  \multicolumn{1}{c|}{\textbf{Test}} &
  \multicolumn{1}{c|}{\textbf{Valid}} &
  \textbf{Test} &
  \multicolumn{1}{c|}{\textbf{Valid}} &
  \multicolumn{1}{c|}{\textbf{Test}} &
  \multicolumn{1}{c|}{\textbf{Valid}} &
  \textbf{Test} \\ \hline
\textbf{ResNet18} &
  \multicolumn{1}{c|}{57.36} &
  \multicolumn{1}{c|}{56.80} &
  \multicolumn{1}{c|}{59.15} &
  58.83 &
  \multicolumn{1}{c|}{50.53} &
  \multicolumn{1}{c|}{49.96} &
  \multicolumn{1}{c|}{52.23} &
  51.72 &
  \multicolumn{1}{c|}{65.94} &
  \multicolumn{1}{c|}{64.07} &
  \multicolumn{1}{c|}{67.37} &
  65.72 \\ \hline
\textbf{ResNet50} &
  \multicolumn{1}{c|}{56.85} &
  \multicolumn{1}{c|}{56.77} &
  \multicolumn{1}{c|}{58.84} &
  58.76 &
  \multicolumn{1}{c|}{50.81} &
  \multicolumn{1}{c|}{50.14} &
  \multicolumn{1}{c|}{51.83} &
  51.35 &
  \multicolumn{1}{c|}{62.53} &
  \multicolumn{1}{c|}{57.31} &
  \multicolumn{1}{c|}{64.08} &
  59.53 \\ \hline
\textbf{resnext50\_32x4d} &
  \multicolumn{1}{c|}{56.17} &
  \multicolumn{1}{c|}{55.68} &
  \multicolumn{1}{c|}{58.97} &
  58.58 &
  \multicolumn{1}{c|}{50.76} &
  \multicolumn{1}{c|}{50.11} &
  \multicolumn{1}{c|}{51.94} &
  51.78 &
  \multicolumn{1}{c|}{58.61} &
  \multicolumn{1}{c|}{54.97} &
  \multicolumn{1}{c|}{59.63} &
  56.12 \\ \hline
\textbf{DenseNet121} &
  \multicolumn{1}{c|}{57.44} &
  \multicolumn{1}{c|}{57.22} &
  \multicolumn{1}{c|}{59.48} &
  59.43 &
  \multicolumn{1}{c|}{50.69} &
  \multicolumn{1}{c|}{50.20} &
  \multicolumn{1}{c|}{53.07} &
  52.82 &
  \multicolumn{1}{c|}{66.08} &
  \multicolumn{1}{c|}{63.11} &
  \multicolumn{1}{c|}{66.61} &
  64.13 \\ \hline
\textbf{ViT\_B\_16} &
  \multicolumn{1}{c|}{57.08} &
  \multicolumn{1}{c|}{57.18} &
  \multicolumn{1}{c|}{58.28} &
  58.02 &
  \multicolumn{1}{c|}{50.58} &
  \multicolumn{1}{c|}{50.20} &
  \multicolumn{1}{c|}{51.41} &
  51.26 &
  \multicolumn{1}{c|}{71.76} &
  \multicolumn{1}{c|}{70.46} &
  \multicolumn{1}{c|}{72.28} &
  71.39 \\ \hline
\textbf{VGG16} &
  \multicolumn{1}{c|}{57.12} &
  \multicolumn{1}{c|}{56.02} &
  \multicolumn{1}{c|}{59.07} &
  58.29 &
  \multicolumn{1}{c|}{51.05} &
  \multicolumn{1}{c|}{50.20} &
  \multicolumn{1}{c|}{51.61} &
  51.05 &
  \multicolumn{1}{c|}{69.87} &
  \multicolumn{1}{c|}{66.01} &
  \multicolumn{1}{c|}{69.90} &
  67.06 \\ \hline
\textbf{EfficientNet\_B0} &
  \multicolumn{1}{c|}{58.57} &
  \multicolumn{1}{c|}{57.69} &
  \multicolumn{1}{c|}{60.01} &
  59.26 &
  \multicolumn{1}{c|}{50.91} &
  \multicolumn{1}{c|}{50.33} &
  \multicolumn{1}{c|}{52.93} &
  53.05 &
  \multicolumn{1}{c|}{69.45} &
  \multicolumn{1}{c|}{65.06} &
  \multicolumn{1}{c|}{69.40} &
  66.48 \\ \hline
\textbf{EfficientNet\_B7} &
  \multicolumn{1}{c|}{59.04} &
  \multicolumn{1}{c|}{58.81} &
  \multicolumn{1}{c|}{60.26} &
  59.80 &
  \multicolumn{1}{c|}{51.93} &
  \multicolumn{1}{c|}{51.10} &
  \multicolumn{1}{c|}{52.97} &
  52.94 &
  \multicolumn{1}{c|}{72.32} &
  \multicolumn{1}{c|}{69.36} &
  \multicolumn{1}{c|}{72.70} &
  70.11 \\ \hline
\textbf{Swin\_B} &
  \multicolumn{1}{c|}{59.76} &
  \multicolumn{1}{c|}{58.36} &
  \multicolumn{1}{c|}{60.42} &
  59.71 &
  \multicolumn{1}{c|}{52.60} &
  \multicolumn{1}{c|}{51.96} &
  \multicolumn{1}{c|}{53.28} &
  52.07 &
  \multicolumn{1}{c|}{76.46} &
  \multicolumn{1}{c|}{74.03} &
  \multicolumn{1}{c|}{76.23} &
  74.15 \\ \hline
\textbf{Swin\_V2\_B} &
  \multicolumn{1}{c|}{59.82} &
  \multicolumn{1}{c|}{59.15} &
  \multicolumn{1}{c|}{59.88} &
  59.17 &
  \multicolumn{1}{c|}{51.71} &
  \multicolumn{1}{c|}{51.50} &
  \multicolumn{1}{c|}{52.38} &
  51.96 &
  \multicolumn{1}{c|}{77.39} &
  \multicolumn{1}{c|}{73.04} &
  \multicolumn{1}{c|}{76.19} &
  74.07 \\ \hline
\textbf{ConvNeXt\_Base} &
  \multicolumn{1}{c|}{60.02} &
  \multicolumn{1}{c|}{59.66} &
  \multicolumn{1}{c|}{60.57} &
  60.76 &
  \multicolumn{1}{c|}{51.96} &
  \multicolumn{1}{c|}{51.37} &
  \multicolumn{1}{c|}{54.31} &
  53.93 &
  \multicolumn{1}{c|}{78.29} &
  \multicolumn{1}{c|}{73.25} &
  \multicolumn{1}{c|}{77.56} &
  73.20 \\ \hline
\textbf{iResNet101} &
  \multicolumn{1}{c|}{57.58} &
  \multicolumn{1}{c|}{57.44} &
  \multicolumn{1}{c|}{59.44} &
  59.61 &
  \multicolumn{1}{c|}{51.31} &
  \multicolumn{1}{c|}{50.08} &
  \multicolumn{1}{c|}{52.83} &
  52.01 &
  \multicolumn{1}{c|}{68.73} &
  \multicolumn{1}{c|}{66.07} &
  \multicolumn{1}{c|}{69.86} &
  67.91 \\ \hline
\textbf{ResNet34} &
  \multicolumn{1}{c|}{59.36} &
  \multicolumn{1}{c|}{58.65} &
  \multicolumn{1}{c|}{59.73} &
  58.98 &
  \multicolumn{1}{c|}{51.66} &
  \multicolumn{1}{c|}{50.51} &
  \multicolumn{1}{c|}{52.64} &
  51.73 &
  \multicolumn{1}{c|}{66.78} &
  \multicolumn{1}{c|}{63.99} &
  \multicolumn{1}{c|}{67.40} &
  65.81 \\ \hline
\textbf{ResNet101} &
  \multicolumn{1}{c|}{56.83} &
  \multicolumn{1}{c|}{56.01} &
  \multicolumn{1}{c|}{58.99} &
  58.61 &
  \multicolumn{1}{c|}{51.37} &
  \multicolumn{1}{c|}{51.33} &
  \multicolumn{1}{c|}{52.50} &
  51.85 &
  \multicolumn{1}{c|}{60.07} &
  \multicolumn{1}{c|}{55.27} &
  \multicolumn{1}{c|}{60.45} &
  56.70 \\ \hline
\textbf{ResNet152} &
  \multicolumn{1}{c|}{58.77} &
  \multicolumn{1}{c|}{58.19} &
  \multicolumn{1}{c|}{60.03} &
  59.91 &
  \multicolumn{1}{c|}{51.25} &
  \multicolumn{1}{c|}{51.07} &
  \multicolumn{1}{c|}{51.07} &
  52.54 &
  \multicolumn{1}{c|}{60.99} &
  \multicolumn{1}{c|}{56.97} &
  \multicolumn{1}{c|}{62.11} &
  58.98 \\ \hline
\textbf{resnext101\_32x8d} &
  \multicolumn{1}{c|}{56.78} &
  \multicolumn{1}{c|}{56.18} &
  \multicolumn{1}{c|}{59.79} &
  59.65 &
  \multicolumn{1}{c|}{50.79} &
  \multicolumn{1}{c|}{50.24} &
  \multicolumn{1}{c|}{52.54} &
  51.63 &
  \multicolumn{1}{c|}{59.15} &
  \multicolumn{1}{c|}{55.66} &
  \multicolumn{1}{c|}{59.58} &
  57.51 \\ \hline
\textbf{resnext101\_64x4d} &
  \multicolumn{1}{c|}{57.66} &
  \multicolumn{1}{c|}{56.74} &
  \multicolumn{1}{c|}{60.30} &
  59.55 &
  \multicolumn{1}{c|}{50.85} &
  \multicolumn{1}{c|}{50.24} &
  \multicolumn{1}{c|}{50.24} &
  52.62 &
  \multicolumn{1}{c|}{59.57} &
  \multicolumn{1}{c|}{54.83} &
  \multicolumn{1}{c|}{55.37} &
  60.97 \\ \hline
\textbf{DenseNet161} &
  \multicolumn{1}{c|}{58.84} &
  \multicolumn{1}{c|}{58.04} &
  \multicolumn{1}{c|}{60.33} &
  59.92 &
  \multicolumn{1}{c|}{51.51} &
  \multicolumn{1}{c|}{50.67} &
  \multicolumn{1}{c|}{53.73} &
  52.89 &
  \multicolumn{1}{c|}{71.51} &
  \multicolumn{1}{c|}{69.41} &
  \multicolumn{1}{c|}{71.93} &
  70.61 \\ \hline
\textbf{DenseNet201} &
  \multicolumn{1}{c|}{57.45} &
  \multicolumn{1}{c|}{56.93} &
  \multicolumn{1}{c|}{60.00} &
  59.58 &
  \multicolumn{1}{c|}{51.73} &
  \multicolumn{1}{c|}{51.07} &
  \multicolumn{1}{c|}{53.76} &
  53.16 &
  \multicolumn{1}{c|}{70.22} &
  \multicolumn{1}{c|}{64.86} &
  \multicolumn{1}{c|}{71.01} &
  67.68 \\ \hline
\textbf{ViT\_B\_32} &
  \multicolumn{1}{c|}{54.26} &
  \multicolumn{1}{c|}{53.70} &
  \multicolumn{1}{c|}{55.89} &
  55.51 &
  \multicolumn{1}{c|}{48.52} &
  \multicolumn{1}{c|}{47.38} &
  \multicolumn{1}{c|}{49.44} &
  48.52 &
  \multicolumn{1}{c|}{60.69} &
  \multicolumn{1}{c|}{58.80} &
  \multicolumn{1}{c|}{61.58} &
  59.98 \\ \hline
\textbf{ViT\_L\_32} &
  \multicolumn{1}{c|}{54.94} &
  \multicolumn{1}{c|}{56.06} &
  \multicolumn{1}{c|}{54.44} &
  56.19 &
  \multicolumn{1}{c|}{49.19} &
  \multicolumn{1}{c|}{50.54} &
  \multicolumn{1}{c|}{48.88} &
  50.03 &
  \multicolumn{1}{c|}{71.55} &
  \multicolumn{1}{c|}{71.27} &
  \multicolumn{1}{c|}{70.07} &
  71.17 \\ \hline
\textbf{VGG11} &
  \multicolumn{1}{c|}{55.66} &
  \multicolumn{1}{c|}{55.16} &
  \multicolumn{1}{c|}{58.05} &
  57.89 &
  \multicolumn{1}{c|}{49.58} &
  \multicolumn{1}{c|}{48.93} &
  \multicolumn{1}{c|}{51.23} &
  50.57 &
  \multicolumn{1}{c|}{70.85} &
  \multicolumn{1}{c|}{68.58} &
  \multicolumn{1}{c|}{70.90} &
  69.27 \\ \hline
\textbf{VGG19} &
  \multicolumn{1}{c|}{58.03} &
  \multicolumn{1}{c|}{57.33} &
  \multicolumn{1}{c|}{58.85} &
  58.67 &
  \multicolumn{1}{c|}{51.60} &
  \multicolumn{1}{c|}{51.35} &
  \multicolumn{1}{c|}{52.00} &
  51.44 &
  \multicolumn{1}{c|}{69.21} &
  \multicolumn{1}{c|}{66.76} &
  \multicolumn{1}{c|}{67.60} &
  66.74 \\ \hline
\textbf{EfficientNet\_B1} &
  \multicolumn{1}{c|}{58.03} &
  \multicolumn{1}{c|}{57.42} &
  \multicolumn{1}{c|}{60.26} &
  60.01 &
  \multicolumn{1}{c|}{51.72} &
  \multicolumn{1}{c|}{51.22} &
  \multicolumn{1}{c|}{53.10} &
  53.04 &
  \multicolumn{1}{c|}{67.02} &
  \multicolumn{1}{c|}{63.01} &
  \multicolumn{1}{c|}{66.60} &
  64.60 \\ \hline
\textbf{EfficientNet\_B2} &
  \multicolumn{1}{c|}{59.39} &
  \multicolumn{1}{c|}{57.81} &
  \multicolumn{1}{c|}{60.55} &
  60.36 &
  \multicolumn{1}{c|}{52.32} &
  \multicolumn{1}{c|}{51.70} &
  \multicolumn{1}{c|}{53.65} &
  53.52 &
  \multicolumn{1}{c|}{68.22} &
  \multicolumn{1}{c|}{64.69} &
  \multicolumn{1}{c|}{68.49} &
  66.62 \\ \hline
\textbf{EfficientNet\_B6} &
  \multicolumn{1}{c|}{57.46} &
  \multicolumn{1}{c|}{57.21} &
  \multicolumn{1}{c|}{60.36} &
  60.41 &
  \multicolumn{1}{c|}{51.48} &
  \multicolumn{1}{c|}{51.08} &
  \multicolumn{1}{c|}{53.59} &
  53.15 &
  \multicolumn{1}{c|}{70.42} &
  \multicolumn{1}{c|}{66.55} &
  \multicolumn{1}{c|}{70.61} &
  68.16 \\ \hline
\textbf{EfficientNet\_V2\_S} &
  \multicolumn{1}{c|}{59.72} &
  \multicolumn{1}{c|}{58.66} &
  \multicolumn{1}{c|}{61.08} &
  60.62 &
  \multicolumn{1}{c|}{51.79} &
  \multicolumn{1}{c|}{51.21} &
  \multicolumn{1}{c|}{53.38} &
  52.57 &
  \multicolumn{1}{c|}{73.71} &
  \multicolumn{1}{c|}{71.84} &
  \multicolumn{1}{c|}{73.94} &
  72.52 \\ \hline
\textbf{EfficientNet\_V2\_M} &
  \multicolumn{1}{c|}{58.08} &
  \multicolumn{1}{c|}{57.49} &
  \multicolumn{1}{c|}{60.11} &
  60.17 &
  \multicolumn{1}{c|}{52.58} &
  \multicolumn{1}{c|}{52.08} &
  \multicolumn{1}{c|}{53.32} &
  53.14 &
  \multicolumn{1}{c|}{69.64} &
  \multicolumn{1}{c|}{66.76} &
  \multicolumn{1}{c|}{70.58} &
  69.19 \\ \hline
\textbf{EfficientNet\_V2\_L} &
  \multicolumn{1}{c|}{59.09} &
  \multicolumn{1}{c|}{58.69} &
  \multicolumn{1}{c|}{60.07} &
  60.08 &
  \multicolumn{1}{c|}{52.20} &
  \multicolumn{1}{c|}{51.25} &
  \multicolumn{1}{c|}{54.00} &
  53.64 &
  \multicolumn{1}{c|}{70.02} &
  \multicolumn{1}{c|}{66.23} &
  \multicolumn{1}{c|}{69.16} &
  67.32 \\ \hline
\textbf{Swin\_T} &
  \multicolumn{1}{c|}{59.31} &
  \multicolumn{1}{c|}{58.03} &
  \multicolumn{1}{c|}{61.09} &
  60.16 &
  \multicolumn{1}{c|}{52.72} &
  \multicolumn{1}{c|}{51.85} &
  \multicolumn{1}{c|}{53.78} &
  52.97 &
  \multicolumn{1}{c|}{73.21} &
  \multicolumn{1}{c|}{71.75} &
  \multicolumn{1}{c|}{72.77} &
  72.17 \\ \hline
\textbf{Swin\_S} &
  \multicolumn{1}{c|}{57.94} &
  \multicolumn{1}{c|}{57.08} &
  \multicolumn{1}{c|}{60.07} &
  59.47 &
  \multicolumn{1}{c|}{51.95} &
  \multicolumn{1}{c|}{51.04} &
  \multicolumn{1}{c|}{53.88} &
  53.33 &
  \multicolumn{1}{c|}{75.44} &
  \multicolumn{1}{c|}{71.84} &
  \multicolumn{1}{c|}{73.16} &
  72.38 \\ \hline
\textbf{Swin\_V2\_T} &
  \multicolumn{1}{c|}{57.45} &
  \multicolumn{1}{c|}{57.16} &
  \multicolumn{1}{c|}{60.11} &
  60.21 &
  \multicolumn{1}{c|}{52.33} &
  \multicolumn{1}{c|}{51.83} &
  \multicolumn{1}{c|}{52.60} &
  52.18 &
  \multicolumn{1}{c|}{73.51} &
  \multicolumn{1}{c|}{70.08} &
  \multicolumn{1}{c|}{73.17} &
  71.59 \\ \hline
\textbf{Swin\_V2\_S} &
  \multicolumn{1}{c|}{59.16} &
  \multicolumn{1}{c|}{59.15} &
  \multicolumn{1}{c|}{60.21} &
  59.89 &
  \multicolumn{1}{c|}{51.45} &
  \multicolumn{1}{c|}{50.81} &
  \multicolumn{1}{c|}{53.05} &
  52.48 &
  \multicolumn{1}{c|}{76.18} &
  \multicolumn{1}{c|}{72.22} &
  \multicolumn{1}{c|}{75.76} &
  72.87 \\ \hline
\textbf{ConvNeXt\_Tiny} &
  \multicolumn{1}{c|}{59.86} &
  \multicolumn{1}{c|}{58.89} &
  \multicolumn{1}{c|}{61.02} &
  60.50 &
  \multicolumn{1}{c|}{52.94} &
  \multicolumn{1}{c|}{52.35} &
  \multicolumn{1}{c|}{54.62} &
  53.67 &
  \multicolumn{1}{c|}{73.13} &
  \multicolumn{1}{c|}{71.50} &
  \multicolumn{1}{c|}{72.98} &
  73.03 \\ \hline
\textbf{ConvNeXt\_Small} &
  \multicolumn{1}{c|}{59.50} &
  \multicolumn{1}{c|}{59.06} &
  \multicolumn{1}{c|}{60.26} &
  60.06 &
  \multicolumn{1}{c|}{53.82} &
  \multicolumn{1}{c|}{52.79} &
  \multicolumn{1}{c|}{54.33} &
  53.65 &
  \multicolumn{1}{c|}{76.97} &
  \multicolumn{1}{c|}{73.30} &
  \multicolumn{1}{c|}{75.84} &
  74.34 \\ \hline
\textbf{ConvNeXt\_Large} &
  \multicolumn{1}{c|}{59.33} &
  \multicolumn{1}{c|}{58.30} &
  \multicolumn{1}{c|}{60.44} &
  60.13 &
  \multicolumn{1}{c|}{51.97} &
  \multicolumn{1}{c|}{51.23} &
  \multicolumn{1}{c|}{52.96} &
  52.56 &
  \multicolumn{1}{c|}{77.82} &
  \multicolumn{1}{c|}{74.10} &
  \multicolumn{1}{c|}{76.19} &
  74.78 \\ \hline
\textbf{Multi-task EfficientNet-B2} &
  \multicolumn{1}{c|}{64.88} &
  \multicolumn{1}{c|}{64.89} &
  \multicolumn{1}{c|}{57.74} &
  57.32 &
  \multicolumn{1}{c|}{61.31} &
  \multicolumn{1}{c|}{60.10} &
  \multicolumn{1}{c|}{52.26} &
  51.53 &
  \multicolumn{1}{c|}{73.85} &
  \multicolumn{1}{c|}{72.08} &
  \multicolumn{1}{c|}{67.17} &
  68.15 \\ \hline
\end{tabular}
\end{table*}

\begin{table*}[]
\centering
\caption{Performance comparison (in \%) between various baseline and state-of-the-art works for VA estimation}\label{table7}
\setlength{\tabcolsep}{1mm}
\scalebox{.85}{
\begin{tabular}{|c|c|c|c|c|c|c|c|c|c|c|c|c|c|c|c|c|}
\hline
\textbf{Model} &
  \textbf{ResNet18} &
  \textbf{ResNet50} &
  \textbf{resnext50} &
  \textbf{DenseNet121} &
  \textbf{ViT\_B\_16} &
  \textbf{VGG16} &
  \textbf{EffNet\_B0} &
  \textbf{EffNet\_B7} &
  \textbf{Swin\_B} &
  \textbf{Swin\_V2\_B} &
  \textbf{ConvNeXt\_B} &
  \textbf{iResNet101} &
  \textbf{FUXI} &
  \textbf{SITU} &
  \textbf{CTC} &
  \textbf{NTU} \\ \hline
\textbf{Test} & 70.8 & 70.8 & 71.1 & 71.3 & 71.4 & 70.8 & 71.6 & 72.0 & 71.9 & 71.9 & 72.7 & 71.5 & 74.1 & 71.9 & 72.2 & 71.5 \\ \hline
\textbf{Age}  & 39.5 & 39.5 & 39.6 & 39.9 & 39.8 & 39.4 & 40.1 & 40.2 & 40.1 & 40.0 & 40.6 & 39.8 & 41.2 & 40.2 & 40.2 & 40.1 \\ \hline
\textbf{Gen.} & 52.8 & 52.9 & 53.1 & 53.3 & 53.3 & 52.9 & 53.5 & 53.7 & 53.6 & 53.8 & 54.2 & 53.4 & 55.3 & 53.7 & 53.9 & 52.8 \\ \hline
\textbf{Race} & 51.4 & 51.3 & 51.3 & 51.6 & 51.4 & 50.8 & 51.6 & 51.8 & 51.6 & 51.6 & 52.3 & 51.5 & 52.8 & 52.0 & 51.8 & 51.5 \\ \hline
\end{tabular}}
\end{table*}

\begin{table*}[]
\centering
\caption{Performance comparison in test and valid set (in \%) between various baseline for AU detection; for F1 higher values are wanted}\label{table6}
\renewcommand{\arraystretch}{1.2}

\begin{tabular}{|c|cc|cc|cc|cc|}
\hline
\multirow{3}{*}{\textbf{Model}} &
  \multicolumn{2}{c|}{\textbf{DISFA}} &
  \multicolumn{2}{c|}{\textbf{EmotioNet}} &
  \multicolumn{2}{c|}{\textbf{RAF-AU}} &
  \multicolumn{2}{c|}{\textbf{GFT}} \\ \cline{2-9} 
 &
  \multicolumn{1}{c|}{\multirow{2}{*}{\textbf{Valid F1}}} &
  \multirow{2}{*}{\textbf{Test F1}} &
  \multicolumn{1}{c|}{\multirow{2}{*}{\textbf{Valid F1}}} &
  \multirow{2}{*}{\textbf{Test F1}} &
  \multicolumn{1}{c|}{\multirow{2}{*}{\textbf{Valid F1}}} &
  \multirow{2}{*}{\textbf{Test F1}} &
  \multicolumn{1}{c|}{\multirow{2}{*}{\textbf{Valid F1}}} &
  \multirow{2}{*}{\textbf{Test F1}} \\
 &
  \multicolumn{1}{c|}{} &
   &
  \multicolumn{1}{c|}{} &
   &
  \multicolumn{1}{c|}{} &
   &
  \multicolumn{1}{c|}{} &
   \\ \hline
\textbf{ResNet18} &
  \multicolumn{1}{c|}{55.6} &
  42.7 &
  \multicolumn{1}{c|}{62.9} &
  57.3 &
  \multicolumn{1}{c|}{68.9} &
  68.2 &
  \multicolumn{1}{c|}{43.8} &
  41.0 \\ \hline
\textbf{ResNet50} &
  \multicolumn{1}{c|}{52.5} &
  39.2 &
  \multicolumn{1}{c|}{60.9} &
  56.0 &
  \multicolumn{1}{c|}{67.1} &
  66.8 &
  \multicolumn{1}{c|}{44.6} &
  41.5 \\ \hline
\textbf{resnext50\_32x4d} &
  \multicolumn{1}{c|}{51.9} &
  38.8 &
  \multicolumn{1}{c|}{59.6} &
  55.2 &
  \multicolumn{1}{c|}{64.5} &
  63.7 &
  \multicolumn{1}{c|}{43.8} &
  39.8 \\ \hline
\textbf{DenseNet121} &
  \multicolumn{1}{c|}{52.6} &
  43.1 &
  \multicolumn{1}{c|}{63.6} &
  58.4 &
  \multicolumn{1}{c|}{69.2} &
  67.9 &
  \multicolumn{1}{c|}{42.7} &
  41.1 \\ \hline
\textbf{ViT\_B\_16} &
  \multicolumn{1}{c|}{52.8} &
  40.3 &
  \multicolumn{1}{c|}{57.3} &
  50.8 &
  \multicolumn{1}{c|}{61.9} &
  61.1 &
  \multicolumn{1}{c|}{43.7} &
  43.0 \\ \hline
\textbf{VGG16} &
  \multicolumn{1}{c|}{56.1} &
  43.2 &
  \multicolumn{1}{c|}{64.7} &
  59.5 &
  \multicolumn{1}{c|}{54.9} &
  55.8 &
  \multicolumn{1}{c|}{41.3} &
  38.1 \\ \hline
\textbf{EfficientNet\_B0} &
  \multicolumn{1}{c|}{55.7} &
  40.5 &
  \multicolumn{1}{c|}{63.4} &
  56.9 &
  \multicolumn{1}{c|}{63.3} &
  63.6 &
  \multicolumn{1}{c|}{45.7} &
  42.7 \\ \hline
\textbf{EfficientNet\_B7} &
  \multicolumn{1}{c|}{50.9} &
  41.4 &
  \multicolumn{1}{c|}{62.4} &
  56.1 &
  \multicolumn{1}{c|}{69.4} &
  68.4 &
  \multicolumn{1}{c|}{42.6} &
  44.8 \\ \hline
\textbf{Swin\_B} &
  \multicolumn{1}{c|}{52.6} &
  44.6 &
  \multicolumn{1}{c|}{62.5} &
  56.8 &
  \multicolumn{1}{c|}{71.4} &
  70.8 &
  \multicolumn{1}{c|}{45.9} &
  42.7 \\ \hline
\textbf{Swin\_V2\_B} &
  \multicolumn{1}{c|}{57.3} &
  43.1 &
  \multicolumn{1}{c|}{63.4} &
  59.1 &
  \multicolumn{1}{c|}{70.5} &
  70.3 &
  \multicolumn{1}{c|}{46.0} &
  44.5 \\ \hline
\textbf{ConvNeXt\_Base} &
  \multicolumn{1}{c|}{58.3} &
  47.8 &
  \multicolumn{1}{c|}{64.6} &
  59.4 &
  \multicolumn{1}{c|}{73.6} &
  73.0 &
  \multicolumn{1}{c|}{46.5} &
  43.4 \\ \hline
\textbf{iResNet101} &
  \multicolumn{1}{c|}{54.6} &
  43.2 &
  \multicolumn{1}{c|}{66.5} &
  60.1 &
  \multicolumn{1}{c|}{72.5} &
  71.9 &
  \multicolumn{1}{c|}{47.2} &
  44.9 \\ \hline
\textbf{ResNet34} &
  \multicolumn{1}{c|}{55.55} &
  43.02 &
  \multicolumn{1}{c|}{63.96} &
  58.58 &
  \multicolumn{1}{c|}{70.32} &
  69.76 &
  \multicolumn{1}{c|}{43.43} &
  42.86 \\ \hline
\textbf{ResNet101} &
  \multicolumn{1}{c|}{50.94} &
  40.99 &
  \multicolumn{1}{c|}{60.79} &
  55.68 &
  \multicolumn{1}{c|}{67.22} &
  67.67 &
  \multicolumn{1}{c|}{44.62} &
  41.47 \\ \hline
\textbf{ResNet152} &
  \multicolumn{1}{c|}{53.81} &
  40.15 &
  \multicolumn{1}{c|}{62.50} &
  56.76 &
  \multicolumn{1}{c|}{68.75} &
  68.26 &
  \multicolumn{1}{c|}{44.86} &
  42.40 \\ \hline
\textbf{resnext101\_32x8d} &
  \multicolumn{1}{c|}{54.35} &
  40.39 &
  \multicolumn{1}{c|}{61.39} &
  56.28 &
  \multicolumn{1}{c|}{67.75} &
  67.27 &
  \multicolumn{1}{c|}{44.88} &
  43.29 \\ \hline
\textbf{resnext101\_64x4d} &
  \multicolumn{1}{c|}{55.97} &
  38.37 &
  \multicolumn{1}{c|}{62.37} &
  55.95 &
  \multicolumn{1}{c|}{67.07} &
  68.19 &
  \multicolumn{1}{c|}{43.90} &
  41.39 \\ \hline
\textbf{DenseNet161} &
  \multicolumn{1}{c|}{55.83} &
  41.65 &
  \multicolumn{1}{c|}{65.24} &
  60.62 &
  \multicolumn{1}{c|}{73.32} &
  72.17 &
  \multicolumn{1}{c|}{45.73} &
  43.54 \\ \hline
\textbf{DenseNet201} &
  \multicolumn{1}{c|}{53.25} &
  44.61 &
  \multicolumn{1}{c|}{64.60} &
  59.60 &
  \multicolumn{1}{c|}{71.43} &
  71.42 &
  \multicolumn{1}{c|}{45.32} &
  43.15 \\ \hline
\textbf{ViT\_B\_32} &
  \multicolumn{1}{c|}{44.01} &
  32.19 &
  \multicolumn{1}{c|}{23.94} &
  21.30 &
  \multicolumn{1}{c|}{46.82} &
  46.60 &
  \multicolumn{1}{c|}{39.65} &
  39.30 \\ \hline
\textbf{ViT\_L\_32} &
  \multicolumn{1}{c|}{44.71} &
  37.31 &
  \multicolumn{1}{c|}{48.95} &
  44.43 &
  \multicolumn{1}{c|}{67.33} &
  68.03 &
  \multicolumn{1}{c|}{43.89} &
  41.49 \\ \hline
\textbf{VGG11} &
  \multicolumn{1}{c|}{56.53} &
  43.00 &
  \multicolumn{1}{c|}{62.99} &
  57.23 &
  \multicolumn{1}{c|}{51.70} &
  51.87 &
  \multicolumn{1}{c|}{43.24} &
  40.25 \\ \hline
\textbf{VGG19} &
  \multicolumn{1}{c|}{50.20} &
  39.57 &
  \multicolumn{1}{c|}{65.11} &
  60.31 &
  \multicolumn{1}{c|}{53.58} &
  53.69 &
  \multicolumn{1}{c|}{42.24} &
  38.58 \\ \hline
\textbf{EfficientNet\_B1} &
  \multicolumn{1}{c|}{54.36} &
  38.58 &
  \multicolumn{1}{c|}{61.86} &
  57.50 &
  \multicolumn{1}{c|}{63.04} &
  61.42 &
  \multicolumn{1}{c|}{42.93} &
  39.36 \\ \hline
\textbf{EfficientNet\_B2} &
  \multicolumn{1}{c|}{51.82} &
  39.68 &
  \multicolumn{1}{c|}{62.18} &
  56.65 &
  \multicolumn{1}{c|}{64.37} &
  65.10 &
  \multicolumn{1}{c|}{43.59} &
  41.70 \\ \hline
\textbf{EfficientNet\_B6} &
  \multicolumn{1}{c|}{53.46} &
  38.93 &
  \multicolumn{1}{c|}{61.99} &
  56.84 &
  \multicolumn{1}{c|}{67.58} &
  65.78 &
  \multicolumn{1}{c|}{44.85} &
  41.75 \\ \hline
\textbf{EfficientNet\_V2\_S} &
  \multicolumn{1}{c|}{54.87} &
  40.26 &
  \multicolumn{1}{c|}{63.57} &
  58.61 &
  \multicolumn{1}{c|}{67.67} &
  68.22 &
  \multicolumn{1}{c|}{45.59} &
  45.80 \\ \hline
\textbf{EfficientNet\_V2\_M} &
  \multicolumn{1}{c|}{53.99} &
  37.25 &
  \multicolumn{1}{c|}{62.56} &
  57.47 &
  \multicolumn{1}{c|}{70.47} &
  69.62 &
  \multicolumn{1}{c|}{45.79} &
  44.58 \\ \hline
\textbf{EfficientNet\_V2\_L} &
  \multicolumn{1}{c|}{53.42} &
  37.06 &
  \multicolumn{1}{c|}{63.26} &
  57.29 &
  \multicolumn{1}{c|}{73.02} &
  72.39 &
  \multicolumn{1}{c|}{44.34} &
  43.44 \\ \hline
\textbf{Swin\_T} &
  \multicolumn{1}{c|}{54.02} &
  40.33 &
  \multicolumn{1}{c|}{64.75} &
  59.68 &
  \multicolumn{1}{c|}{69.65} &
  70.24 &
  \multicolumn{1}{c|}{45.17} &
  43.32 \\ \hline
\textbf{Swin\_S} &
  \multicolumn{1}{c|}{54.32} &
  42.50 &
  \multicolumn{1}{c|}{63.47} &
  57.55 &
  \multicolumn{1}{c|}{71.85} &
  70.73 &
  \multicolumn{1}{c|}{45.02} &
  45.04 \\ \hline
\textbf{Swin\_V2\_T} &
  \multicolumn{1}{c|}{55.12} &
  44.09 &
  \multicolumn{1}{c|}{64.81} &
  58.91 &
  \multicolumn{1}{c|}{68.49} &
  68.64 &
  \multicolumn{1}{c|}{45.41} &
  43.37 \\ \hline
\textbf{Swin\_V2\_S} &
  \multicolumn{1}{c|}{55.43} &
  40.97 &
  \multicolumn{1}{c|}{64.23} &
  59.26 &
  \multicolumn{1}{c|}{70.33} &
  69.37 &
  \multicolumn{1}{c|}{43.77} &
  44.08 \\ \hline
\textbf{ConvNeXt\_Tiny} &
  \multicolumn{1}{c|}{57.78} &
  42.91 &
  \multicolumn{1}{c|}{65.08} &
  65.08 &
  \multicolumn{1}{c|}{65.85} &
  65.68 &
  \multicolumn{1}{c|}{45.90} &
  45.21 \\ \hline
\textbf{ConvNeXt\_Small} &
  \multicolumn{1}{c|}{56.53} &
  46.66 &
  \multicolumn{1}{c|}{65.10} &
  60.35 &
  \multicolumn{1}{c|}{75.00} &
  73.63 &
  \multicolumn{1}{c|}{47.82} &
  46.23 \\ \hline
\textbf{ConvNeXt\_Large} &
  \multicolumn{1}{c|}{55.77} &
  41.74 &
  \multicolumn{1}{c|}{63.52} &
  58.20 &
  \multicolumn{1}{c|}{74.69} &
  73.54 &
  \multicolumn{1}{c|}{48.23} &
  45.48 \\ \hline
\textbf{EfficientNet\_V2\_S} &
  \multicolumn{1}{c|}{54.87} &
  40.26 &
  \multicolumn{1}{c|}{63.57} &
  58.61 &
  \multicolumn{1}{c|}{67.67} &
  68.22 &
  \multicolumn{1}{c|}{45.59} &
  45.80 \\ \hline
\textbf{EfficientNet\_V2\_M} &
  \multicolumn{1}{c|}{53.99} &
  37.25 &
  \multicolumn{1}{c|}{62.56} &
  57.47 &
  \multicolumn{1}{c|}{70.47} &
  69.62 &
  \multicolumn{1}{c|}{45.79} &
  44.58 \\ \hline
\textbf{EfficientNet\_V2\_L} &
  \multicolumn{1}{c|}{53.42} &
  37.06 &
  \multicolumn{1}{c|}{63.26} &
  57.29 &
  \multicolumn{1}{c|}{73.02} &
  72.39 &
  \multicolumn{1}{c|}{44.34} &
  43.44 \\ \hline
\textbf{Swin\_T} &
  \multicolumn{1}{c|}{54.02} &
  40.33 &
  \multicolumn{1}{c|}{64.75} &
  59.68 &
  \multicolumn{1}{c|}{69.65} &
  70.24 &
  \multicolumn{1}{c|}{45.17} &
  43.32 \\ \hline
\textbf{Swin\_S} &
  \multicolumn{1}{c|}{54.32} &
  42.50 &
  \multicolumn{1}{c|}{63.47} &
  57.55 &
  \multicolumn{1}{c|}{71.85} &
  70.73 &
  \multicolumn{1}{c|}{45.02} &
  45.04 \\ \hline
\textbf{Swin\_V2\_T} &
  \multicolumn{1}{c|}{55.12} &
  44.09 &
  \multicolumn{1}{c|}{64.81} &
  58.91 &
  \multicolumn{1}{c|}{68.49} &
  68.64 &
  \multicolumn{1}{c|}{45.41} &
  43.37 \\ \hline
\textbf{Swin\_V2\_S} &
  \multicolumn{1}{c|}{55.43} &
  40.97 &
  \multicolumn{1}{c|}{64.23} &
  59.26 &
  \multicolumn{1}{c|}{70.33} &
  69.37 &
  \multicolumn{1}{c|}{43.77} &
  44.08 \\ \hline
\textbf{ConvNeXt\_Tiny} &
  \multicolumn{1}{c|}{57.78} &
  42.91 &
  \multicolumn{1}{c|}{65.08} &
  65.08 &
  \multicolumn{1}{c|}{65.85} &
  65.68 &
  \multicolumn{1}{c|}{45.90} &
  45.21 \\ \hline
\textbf{ConvNeXt\_Small} &
  \multicolumn{1}{c|}{56.53} &
  46.66 &
  \multicolumn{1}{c|}{65.10} &
  60.35 &
  \multicolumn{1}{c|}{75.00} &
  73.63 &
  \multicolumn{1}{c|}{47.82} &
  46.23 \\ \hline
\textbf{ConvNeXt\_Large} &
  \multicolumn{1}{c|}{55.77} &
  41.74 &
  \multicolumn{1}{c|}{63.52} &
  58.20 &
  \multicolumn{1}{c|}{74.69} &
  73.54 &
  \multicolumn{1}{c|}{48.23} &
  45.48 \\ \hline
\end{tabular}
\end{table*}

\begin{table*}[]
\centering
\caption{Performance comparison (in \%) between various baseline for VA estimation in AffectNet-VA and AFEW}\label{table7}
\renewcommand{\arraystretch}{1.2}
\begin{tabular}{|c|cccc|cccc|}
\hline
\multirow{3}{*}{\textbf{Model}} &
  \multicolumn{4}{c|}{\textbf{AFEW}} &
  \multicolumn{4}{c|}{\textbf{AffectNet-VA}} \\ \cline{2-9} 
 &
  \multicolumn{2}{c|}{\textbf{CCC\_V}} &
  \multicolumn{2}{c|}{\textbf{CCC\_A}} &
  \multicolumn{2}{c|}{\textbf{CCC\_V}} &
  \multicolumn{2}{c|}{\textbf{CCC\_A}} \\ \cline{2-9} 
 &
  \multicolumn{1}{c|}{\textbf{Valid}} &
  \multicolumn{1}{c|}{\textbf{Test}} &
  \multicolumn{1}{c|}{\textbf{Valid}} &
  \textbf{Test} &
  \multicolumn{1}{c|}{\textbf{Valid}} &
  \multicolumn{1}{c|}{\textbf{Test}} &
  \multicolumn{1}{c|}{\textbf{Valid}} &
  \textbf{Test} \\ \hline
\textbf{ResNet18} &
  \multicolumn{1}{c|}{49.6} &
  \multicolumn{1}{c|}{46.0} &
  \multicolumn{1}{c|}{44.5} &
  48.6 &
  \multicolumn{1}{c|}{82.2} &
  \multicolumn{1}{c|}{82.1} &
  \multicolumn{1}{c|}{60.1} &
  59.5 \\ \hline
\textbf{ResNet50} &
  \multicolumn{1}{c|}{46.1} &
  \multicolumn{1}{c|}{52.5} &
  \multicolumn{1}{c|}{40.3} &
  49.1 &
  \multicolumn{1}{c|}{82.0} &
  \multicolumn{1}{c|}{81.9} &
  \multicolumn{1}{c|}{60.7} &
  59.7 \\ \hline
\textbf{resnext50\_32x4d} &
  \multicolumn{1}{c|}{46.2} &
  \multicolumn{1}{c|}{44.5} &
  \multicolumn{1}{c|}{38.1} &
  51.5 &
  \multicolumn{1}{c|}{81.9} &
  \multicolumn{1}{c|}{81.8} &
  \multicolumn{1}{c|}{60.9} &
  60.3 \\ \hline
\textbf{DenseNet121} &
  \multicolumn{1}{c|}{48.8} &
  \multicolumn{1}{c|}{42.5} &
  \multicolumn{1}{c|}{42.5} &
  53.3 &
  \multicolumn{1}{c|}{82.4} &
  \multicolumn{1}{c|}{82.3} &
  \multicolumn{1}{c|}{61.2} &
  60.4 \\ \hline
\textbf{ViT\_B\_16} &
  \multicolumn{1}{c|}{46.2} &
  \multicolumn{1}{c|}{46.9} &
  \multicolumn{1}{c|}{50.0} &
  53.1 &
  \multicolumn{1}{c|}{70.1} &
  \multicolumn{1}{c|}{70.3} &
  \multicolumn{1}{c|}{49.3} &
  49.1 \\ \hline
\textbf{VGG16} &
  \multicolumn{1}{c|}{51.3} &
  \multicolumn{1}{c|}{47.4} &
  \multicolumn{1}{c|}{50.0} &
  55.9 &
  \multicolumn{1}{c|}{82.3} &
  \multicolumn{1}{c|}{82.1} &
  \multicolumn{1}{c|}{61.1} &
  60.4 \\ \hline
\textbf{EfficientNet\_B0} &
  \multicolumn{1}{c|}{49.4} &
  \multicolumn{1}{c|}{47.2} &
  \multicolumn{1}{c|}{36.4} &
  51.1 &
  \multicolumn{1}{c|}{82.7} &
  \multicolumn{1}{c|}{82.7} &
  \multicolumn{1}{c|}{61.6} &
  60.9 \\ \hline
\textbf{EfficientNet\_B7} &
  \multicolumn{1}{c|}{45.0} &
  \multicolumn{1}{c|}{44.8} &
  \multicolumn{1}{c|}{40.0} &
  59.0 &
  \multicolumn{1}{c|}{82.7} &
  \multicolumn{1}{c|}{82.7} &
  \multicolumn{1}{c|}{62.4} &
  61.4 \\ \hline
\textbf{Swin\_B} &
  \multicolumn{1}{c|}{52.2} &
  \multicolumn{1}{c|}{39.2} &
  \multicolumn{1}{c|}{40.3} &
  56.2 &
  \multicolumn{1}{c|}{82.3} &
  \multicolumn{1}{c|}{82.4} &
  \multicolumn{1}{c|}{61.6} &
  61.4 \\ \hline
\textbf{Swin\_V2\_B} &
  \multicolumn{1}{c|}{49.0} &
  \multicolumn{1}{c|}{42.2} &
  \multicolumn{1}{c|}{41.4} &
  56.7 &
  \multicolumn{1}{c|}{82.8} &
  \multicolumn{1}{c|}{82.8} &
  \multicolumn{1}{c|}{61.8} &
  61.6 \\ \hline
\textbf{ConvNeXt\_Base} &
  \multicolumn{1}{c|}{54.5} &
  \multicolumn{1}{c|}{42.9} &
  \multicolumn{1}{c|}{44.2} &
  53.1 &
  \multicolumn{1}{c|}{82.5} &
  \multicolumn{1}{c|}{82.6} &
  \multicolumn{1}{c|}{62.1} &
  62.1 \\ \hline
\textbf{iResNet101} &
  \multicolumn{1}{c|}{56.7} &
  \multicolumn{1}{c|}{45.3} &
  \multicolumn{1}{c|}{47.6} &
  57.5 &
  \multicolumn{1}{c|}{82.5} &
  \multicolumn{1}{c|}{82.7} &
  \multicolumn{1}{c|}{61.4} &
  60.8 \\ \hline
\end{tabular}
\end{table*}

\begin{table*}[]
\centering
\caption{Performance comparison (in \%) between various baseline and state-of-the-art works for VA estimation in AffectNet-VA}\label{table7}
\setlength{\tabcolsep}{3mm}
\renewcommand{\arraystretch}{1.2}
\begin{tabular}{|c|cccc|}
\hline
\multirow{3}{*}{\textbf{Model}} & \multicolumn{4}{c|}{\textbf{AffectNet-VA}}                                                     \\ \cline{2-5} 
                                & \multicolumn{2}{c|}{\textbf{CCC\_V}}                    & \multicolumn{2}{c|}{\textbf{CCC\_A}} \\ \cline{2-5} 
 & \multicolumn{1}{c|}{\textbf{Valid}} & \multicolumn{1}{c|}{\textbf{Test}} & \multicolumn{1}{c|}{\textbf{Valid}} & \textbf{Test} \\ \hline
\textbf{ResNet34}               & \multicolumn{1}{c|}{82.22} & \multicolumn{1}{c|}{82.20} & \multicolumn{1}{c|}{60.46}  & 60.11  \\ \hline
\textbf{ResNet101}              & \multicolumn{1}{c|}{82.12} & \multicolumn{1}{c|}{81.87} & \multicolumn{1}{c|}{61.08}  & 59.98  \\ \hline
\textbf{ResNet152}              & \multicolumn{1}{c|}{82.32} & \multicolumn{1}{c|}{82.28} & \multicolumn{1}{c|}{61.62}  & 60.70  \\ \hline
\textbf{resnext101\_32x8d}      & \multicolumn{1}{c|}{82.57} & \multicolumn{1}{c|}{82.51} & \multicolumn{1}{c|}{61.59}  & 60.60  \\ \hline
\textbf{resnext101\_64x4d}      & \multicolumn{1}{c|}{82.48} & \multicolumn{1}{c|}{82.43} & \multicolumn{1}{c|}{60.38}  & 59.94  \\ \hline
\textbf{DenseNet161}            & \multicolumn{1}{c|}{83.03} & \multicolumn{1}{c|}{82.88} & \multicolumn{1}{c|}{61.25}  & 60.61  \\ \hline
\textbf{DenseNet201}            & \multicolumn{1}{c|}{82.44} & \multicolumn{1}{c|}{82.23} & \multicolumn{1}{c|}{62.36}  & 61.72  \\ \hline
\textbf{ViT\_B\_32}             & \multicolumn{1}{c|}{80.68} & \multicolumn{1}{c|}{80.62} & \multicolumn{1}{c|}{58.85}  & 57.91  \\ \hline
\textbf{ViT\_L\_32}             & \multicolumn{1}{c|}{80.68} & \multicolumn{1}{c|}{80.62} & \multicolumn{1}{c|}{58.85}  & 57.91  \\ \hline
\textbf{VGG11}                  & \multicolumn{1}{c|}{81.40} & \multicolumn{1}{c|}{81.19} & \multicolumn{1}{c|}{60.19}  & 59.77  \\ \hline
\textbf{VGG19}                  & \multicolumn{1}{c|}{82.25} & \multicolumn{1}{c|}{82.00} & \multicolumn{1}{c|}{61.08}  & 60.53  \\ \hline
\textbf{EfficientNet\_B1}       & \multicolumn{1}{c|}{82.65} & \multicolumn{1}{c|}{82.71} & \multicolumn{1}{c|}{61.59}  & 61.08  \\ \hline
\textbf{EfficientNet\_B2}       & \multicolumn{1}{c|}{82.94} & \multicolumn{1}{c|}{83.00} & \multicolumn{1}{c|}{62.45}  & 61.44  \\ \hline
\textbf{EfficientNet\_B6}       & \multicolumn{1}{c|}{82.70} & \multicolumn{1}{c|}{82.84} & \multicolumn{1}{c|}{62.24}  & 61.69  \\ \hline
\textbf{EfficientNet\_V2\_S}    & \multicolumn{1}{c|}{82.75} & \multicolumn{1}{c|}{82.76} & \multicolumn{1}{c|}{62.46}  & 61.92  \\ \hline
\textbf{EfficientNet\_V2\_M}    & \multicolumn{1}{c|}{82.91} & \multicolumn{1}{c|}{82.94} & \multicolumn{1}{c|}{61.66}  & 61.30  \\ \hline
\textbf{EfficientNet\_V2\_L}    & \multicolumn{1}{c|}{82.98} & \multicolumn{1}{c|}{82.95} & \multicolumn{1}{c|}{62.31}  & 61.78  \\ \hline
\textbf{Swin\_T}                & \multicolumn{1}{c|}{82.77} & \multicolumn{1}{c|}{82.52} & \multicolumn{1}{c|}{62.52}  & 61.59  \\ \hline
\textbf{Swin\_S}                & \multicolumn{1}{c|}{83.03} & \multicolumn{1}{c|}{82.99} & \multicolumn{1}{c|}{62.13}  & 61.11  \\ \hline
\textbf{Swin\_V2\_T}            & \multicolumn{1}{c|}{82.83} & \multicolumn{1}{c|}{82.70} & \multicolumn{1}{c|}{61.89}  & 61.23  \\ \hline
\textbf{Swin\_V2\_S}            & \multicolumn{1}{c|}{82.99} & \multicolumn{1}{c|}{82.96} & \multicolumn{1}{c|}{62.43}  & 61.61  \\ \hline
\textbf{ConvNeXt\_Tiny}         & \multicolumn{1}{c|}{82.69} & \multicolumn{1}{c|}{82.60} & \multicolumn{1}{c|}{60.92}  & 61.17  \\ \hline
\textbf{ConvNeXt\_Small}        & \multicolumn{1}{c|}{82.40} & \multicolumn{1}{c|}{82.26} & \multicolumn{1}{c|}{60.32}  & 59.83  \\ \hline
\textbf{ConvNeXt\_Large}        & \multicolumn{1}{c|}{82.40} & \multicolumn{1}{c|}{82.38} & \multicolumn{1}{c|}{60.63}  & 60.39  \\ \hline
\end{tabular}
\end{table*}
